\RequirePackage{fix-cm}
\documentclass[twocolumn]{svjour3}          
\smartqed  

\usepackage{graphicx}
\usepackage{hyperref}   
\usepackage{dsfont}
\usepackage{makecell}
\usepackage{url}
\usepackage{wrapfig}
 \usepackage[table]{xcolor}
\usepackage[symbol]{footmisc}

\newcommand\restr[2]{{
  \left.\kern-\nulldelimiterspace 
  #1 
  \vphantom{\big|} 
  \right|_{#2} 
  }}
\usepackage[utf8]{inputenc} 
\usepackage[T1]{fontenc}    
\usepackage{url}            
\usepackage{booktabs}       
\usepackage{nicefrac}       
\usepackage{microtype}      
\usepackage{xcolor}         
\usepackage{graphicx}
\usepackage{xurl} 
\usepackage{mathtools, bbm}
\usepackage{amssymb}
\usepackage{algorithm2e}
\usepackage{enumerate}
\usepackage{svg}
\usepackage{booktabs}
\usepackage{textcomp}
\usepackage{pifont}
\usepackage{multirow}
\usepackage{wrapfig}
\usepackage{float}
\newcommand{\cmark}{\ding{51}}%
\newcommand{\xmark}{\ding{55}}%
\newcommand{\minisection}[1]{\vspace{0.1in} \noindent {\bf #1}}

\definecolor{darkgreen}{RGB}{0,100,0}

\newenvironment{dataavailability}{%
   \par\addvspace{17pt}\small\noindent\textbf{Data Availability}~%
}{\par\addvspace{6pt}}

\begin{document}

\title{EFC++: Elastic Feature Consolidation with Prototype Re-balancing for Cold Start Exemplar-free Incremental Learning}

\author{Simone Magistri$^{*}$  \and Tomaso Trinci$^{*}$ \and  Albin Soutif{-}Cormerais \and  \\ Joost van de Weijer \and Andrew D. Bagdanov
}

\authorrunning{Simone Magistri, Tomaso Trinci et al.} 
\titlerunning{EFC++: Elastic Feature Consolidation with Prototype Re-balancing}

\institute{Simone Magistri. ORCID: \href{https://orcid.org/0000-0002-0520-8463}{0000-0002-0520-8463} \at 
Media Integration and Communication Center (MICC),  University of Florence, Italy\\
\email{simone.magistri@unifi.it} 
\and 
Tomaso Trinci. ORCID: \href{https://orcid.org/0000-0002-4052-1930}{0000-0002-4052-1930} \at 
 Global Optimization Laboratory, University of Florence, Italy  
\and 
Albin Soutif{-}Cormerais. ORCID: \href{https://orcid.org/0009-0008-1564-2029}{0009-0008-1564-2029} \at 
LAMP Team, Computer Vision Center, Barcelona, Spain 
\and 
Joost van de Weijer. ORCID: \href{https://orcid.org/0000-0002-9656-9706}{0000-0002-9656-9706}    \at 
LAMP Team, Computer Vision Center, Universitat Aut\`{o}noma de Barcelona, Spain \and 
Andrew D. Bagdanov. ORCID: \href{https://orcid.org/0000-0001-6408-7043}{0000-0001-6408-7043} \at 
Media Integration and Communication Center (MICC), University of Florence, Italy  \and
* Equal contribution 
}

\date{}

 
\def\makeheadbox{}  

\maketitle

\begin{abstract}
 
Exemplar-free Class Incremental Learning (EFCIL) aims to learn from a sequence of tasks without having access to previous task data. In this paper, we consider the challenging Cold Start scenario in which insufficient data is available in the first task to learn a high-quality backbone. This is especially challenging for EFCIL since it requires high plasticity, resulting in feature drift which is difficult to compensate for in the exemplar-free setting. To address this problem, we propose an effective approach to consolidate feature representations by regularizing drift in directions highly relevant to previous tasks while employing prototypes to reduce task-recency bias. Our approach, which we call Elastic Feature Consolidation++ (EFC++) exploits a tractable second-order approximation of feature drift based on a proposed Empirical Feature Matrix (EFM). The EFM induces a pseudo-metric in feature space which we use to regularize feature drift in important directions and to update Gaussian prototypes. In addition, we introduce a post-training prototype re-balancing phase that updates classifiers to compensate for feature drift. This strategy allows to improve over our previous EFC method by mitigating the misalignment between stored prototypes and the evolving feature space. Extensive experimental results on Tiny-ImageNet, ImageNet-Subset, ImageNet-1K, and DomainNet  show that EFC++ achieves a strong stability--plasticity trade-off in Cold Start and outperforms recent exemplar-free baselines. Code is available at \url{https://github.com/simomagi/elastic_feature_consolidation}.
 
\end{abstract}

\section{Introduction}
 Deep Neural Networks (DNNs) like Convolutional Neural Networks~\cite{resnet} and Vision Transformers~\cite{dosovitskiy2021an}  achieve strong performance  across many applications. In real-world applications like autonomous driving~\cite{Shaheen2022}, multimedia forensics~\cite{MAGISTRI202482}, healthcare~\cite{Lee2020} or remote sensing~\cite{9771396}, data are often not stationary, necessitating the adaptation of DNNs to new data and tasks. Updating models is challenging due to the phenomenon known as \textit{catastrophic forgetting}~\cite{mccloskey1989catastrophic,French1999}, i.e., the tendency of DNNs to lose previously acquired knowledge when learning new tasks. Class Incremental Learning (CIL) addresses this problem by enabling models to learn new classes without retraining from scratch or storing all previous data~\cite{verwimp2024continual,facil}. The key challenge in CIL is to find the right equilibrium between stability and plasticity during network training. Striking this balance is often referred to as the \textit{stability-plasticity} dilemma~\cite{10444954,10.3389/fpsyg.2013.00504}.

Relying solely on data from the current task to mitigate catastrophic forgetting is challenging because the final deep representation and classifier tend to be highly biased towards the most recent task (a phenomenon known as \textit{task-recency bias}). Additionally, the final classifier is never trained to distinguish among all the encountered classes (a phenomenon called \textit{inter-task confusion}~\cite{facil}). To address this, many recent CIL approaches, known as Exemplar-based Class Incremental Learning (EBCIL) methods, store samples from previous tasks (e.g., 20 per class) as exemplars and replay them while learning new tasks~\cite{adaptive_feature_consolidation,der,memo}. While effective, these methods require additional storage and computational resources, which can scale with the number of tasks~\cite{how-green,lanzillotta2026heads}. Furthermore, storing samples can pose significant privacy issues, as it may not always be feasible to retain such data due to privacy concerns~\cite{privacy-inc,10.1145/3785663}. For these reasons, in this paper we focus on the more general and challenging incremental learning setting without exemplars, a paradigm known as Exemplar-free Class Incremental Learning (EFCIL). 

Elastic Weight Consolidation (EWC) is a pioneering work in EFCIL~\cite{ewc}. EWC addresses catastrophic forgetting via weight regularization. EWC is an elegant approach based on a Laplace approximation of the previous task posterior. When training on a new task, an approximate Fisher Information Matrix is used to regularize parameter drift to prevent the optimization from harming parameters important for previous tasks, while simultaneously increasing plasticity to learn new tasks. Since computing the full Fisher Information Matrix is intractable (its size is quadratic in the number of network parameters), EWC uses a diagonal approximation of the matrix. This approximation is an aggressive simplification compared to the full matrix and results in a significant performance drop~\cite{rotate_net,kronecker_fisher}.

Given the difficulties in regularizing network weights, recent EFCIL methods rely on the more tractable functional regularization paradigm which mitigates network activation drift during incremental learning. Feature distillation, consisting in regularizing feature space drift via $\ell_2$ regularization, is a central component in recent state-of-the-art EFCIL approaches~\cite{generative_features,pass,ssre,evanescent,sdc}. Feature distillation is often combined with class prototypes, either learned~\cite{evanescent} or based on class means~\cite{pass}, to perform pseudo-rehearsal of features from previous tasks in order to reduce task-recency bias and inter-task confusion phenomenon. Prototypes -- differently from exemplars -- offer an efficient and privacy-preserving way to mitigate forgetting. 

While feature distillation effectively mitigates forgetting through increased stability, it does so at the cost of reduced plasticity for learning new tasks~\cite{podnet}. A similar emphasis on stability is also observed in other recent EFCIL methods that freeze the backbone after the first task and focus solely on incrementally learning the classifier~\cite{fetril,goswami2023fecam}. This emphasis on stability largely stems from the standard evaluation setting for EFCIL methods~\cite{ssre,pass,evanescent,fetril,goswami2023fecam}. In this setting, called the \textit{Warm Start} scenario, the first task includes a large fraction of the total classes, typically 40\% or 50\%, making stability more critical than plasticity. Since a high-quality feature extractor is learned in the first task, it improves generalization in subsequent tasks. However, when the initial task is smaller, an incremental learning method must increase plasticity to compensate for the lack of a strong initial feature extractor, requiring substantial updates to the backbone in order to effectively learn new tasks.

\begin{figure}
\centering
\includegraphics[width=0.9\columnwidth]{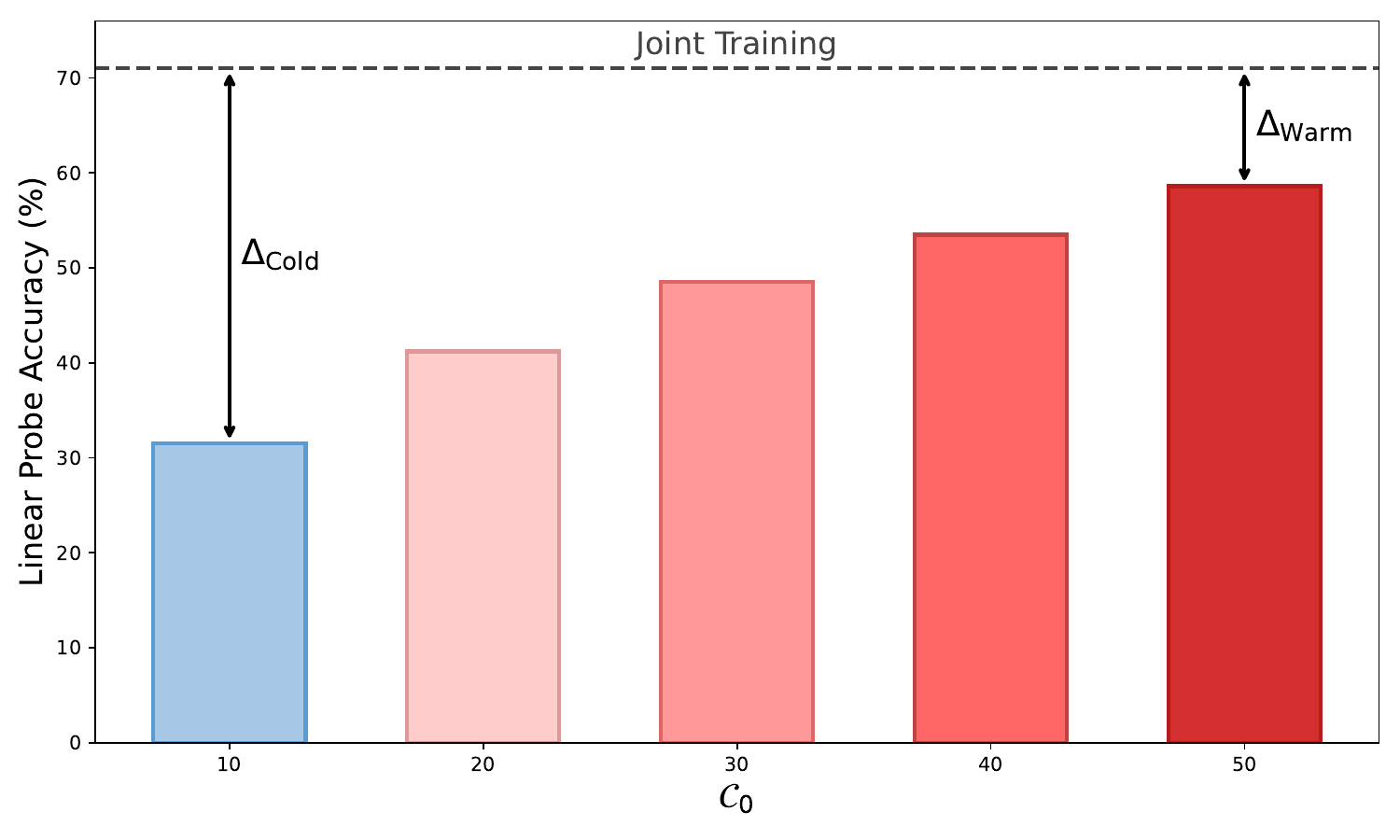}
\caption{Plasticity potential in Cold and Warm Start. 
We train a ResNet-18 on $\mathcal{C}_0$ classes of CIFAR-100 for $\mathcal{C}_0 = 10, 20, \dots, 50$, and evaluate feature quality via linear probing~\cite{davari2022probing} on all 100 classes. The plasticity potential $\Delta$, 
defined as the maximum performance gain which can be obtained with a plastic versus a frozen backbone, is quantified as the performance gap between Joint Training and a model frozen on a subset of classes. Note that  $\Delta_{\text{Cold}}$ (the \emph{Cold Start} scenario) is significantly larger than $\Delta_{\text{Warm}}$ (the \emph{Warm Start} case). This is due to the inability to learn a strong feature extractor on only $\mathcal{C}_0$ classes and hence greater plasticity is required to incrementally learn new classes. Freezing the backbone in such settings limits adaptability and ultimately constrains performance on subsequent tasks.}
\label{fig:probe}
\end{figure}

In Figure~\ref{fig:probe} we give an empirical analysis via linear probing to demonstrate the importance of plasticity in EFCIL for different initial task sizes. These results show that for smaller initial task sizes (e.g. $\mathcal{C}_0=10$), the plasticity potential $\Delta$ -- defined as the maximum performance gain which can be obtained with a plastic versus a frozen backbone -- is highest. This reflects the greater need for feature adaptation in later incremental learning stages, as a weaker initial backbone demands more flexibility to incorporate new knowledge.

In this paper we examine EFCIL in the more challenging \emph{Cold Start} scenario in which the first task is insufficiently large to learn a high-quality backbone and methods must be plastic and adapt their backbone to new tasks. Although this scenario has been explored in the EBCIL literature~\cite{memo,der}, it is rarely addressed in EFCIL due to several challenges. In Cold Start scenarios, representations are more prone to change than in Warm Start, making it difficult to control drift without compromising plasticity -- especially when there are no exemplars to rely on. Additionally, using a strict regularizer, like feature distillation or freezing the backbone, prevents the network from adapting to new data. Therefore, an alternative to feature distillation is necessary and an exemplar-free mechanism is needed to adapt previous-task classifiers to the evolving backbone. 

To address the challenges of Cold Start, we propose a novel EFCIL approach, called \emph{Elastic Feature Consolidation with Prototype Re-balancing} (EFC++), that builds upon our previous  method EFC~\cite{magistri2024elastic}. We introduce the Empirical Feature Matrix (EFM), a tractable second-order pseudo metric that regularizes changes along directions most relevant to past tasks while promoting the plasticity needed in Cold Start. Unlike the Fisher Information Matrix, the EFM can be easily computed and stored, since its size depends only on the feature space dimensionality. As in EFC, EFC++ also relies on Gaussian prototypes to mitigate inter-task confusion. However, we show that in high-plasticity Cold Start regimes, the joint optimization of the feature extractor and prototype-rehearsal based classifier used in EFC makes stored prototypes progressively misaligned with the evolving feature space. EFC++ addresses this limitation by decoupling feature learning from classifier adaptation: features are learned with EFM regularization, while prototypes are updated after feature learning and used to re-balance a global classifier over all seen classes. This design mitigates prototype-feature misalignment and enables a more effective use of EFM regularization.
In addition, compared to our previous work~\cite{magistri2024elastic}, we perform a more detailed diagnostic study of the method, including Warm versus Cold Start drift behavior, prototype re-balancing ablations, spectral and perturbation analyses of the EFM, hyperparameter sensitivity, efficiency considerations, and remaining failure modes, including the central role of class mean estimation over covariance estimation in EFCIL prototype rehearsal.

\minisection{Contributions.}
Our main contributions are:
\begin{itemize}
    \item \textbf{Cold Start EFCIL.}
    We study exemplar-free class-incremental learning under the challenging \textit{Cold Start} setting, in which the first task is insufficient to learn a strong backbone. We show that the plasticity required in this regime induces structured feature drift, disproportionately affects earlier tasks, and progressively degrades prototype reliability.

    \item \textbf{Empirical Feature Matrix.}
    We present the Empirical Feature Matrix (EFM), a tractable feature-space analogue of second-order importance estimation. The EFM induces a pseudo-metric that selectively regularizes drift along directions important for previous tasks, while preserving plasticity in less relevant directions.

    \item \textbf{Elastic Feature Consolidation.}
    We propose Elastic Feature Consolidation (EFC) and its extension \textit{Elastic Feature Consolidation with Prototype Re-\allowbreak balancing} (EFC++). EFC combines EFM-based regularization with asymmetric prototype rehearsal and EFM-guided prototype drift compensation. EFC++ further decouples feature learning from classifier adaptation through prototype re-balancing, improving prototype alignment in Cold Start scenarios.

    \item \textbf{Comprehensive Evaluation.}
    We evaluate EFC and EFC++ on CIFAR-100, Tiny ImageNet, ImageNet Subset, ImageNet-1K, and DomainNet in \textit{Cold Start} and \textit{Warm Start} regimes. Results show  competitive performance on CIFAR-100, and state-of-the-art performance on more challenging benchmarks compared to recent EFCIL methods.
\end{itemize}

The paper is organized as follows. Section~\ref{sec:feature-drift-regularization} introduces the Empirical Feature Matrix (EFM). Section~\ref{sec:efc++} presents Elastic Feature Consolidation with Prototype Re-balancing (EFC++). Sections~\ref{sec:experimental_setting}--\ref{sec:experimental-results} present the experimental setup and comparisons with state-of-the-art EFCIL methods. Section~\ref{sec:ablation-analysis} reports ablations and additional analyses. Finally, Section~\ref{sec:conclusion_and_limitation} concludes with our main contributions and future directions.

\begin{figure*}
    \centering
\includegraphics[width=\textwidth]{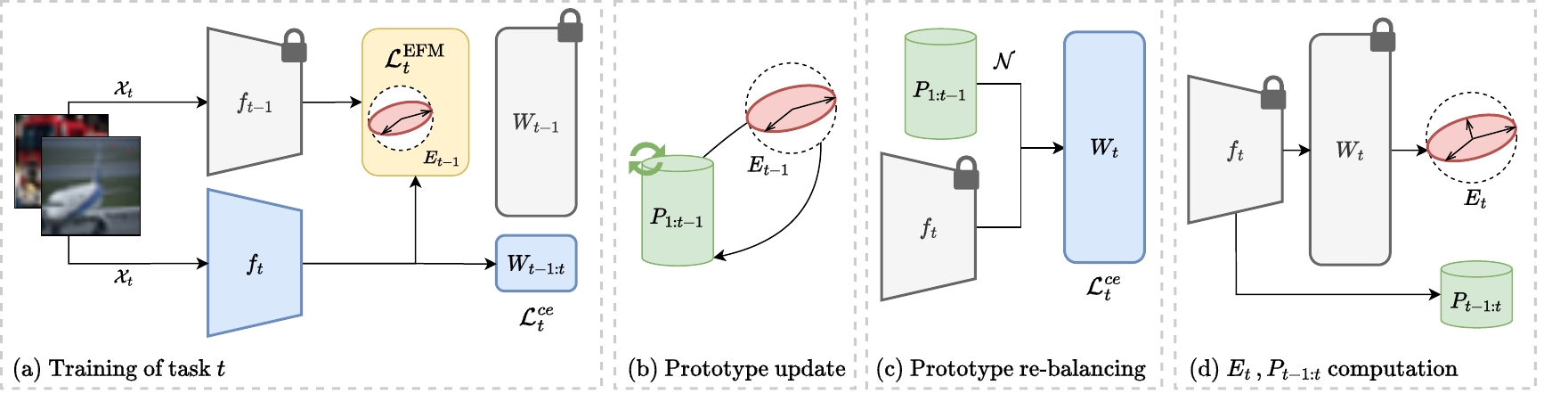}
    \caption{Elastic Feature Consolidation with Prototype Re-balancing (EFC++). \textbf{(a)} EFC++ leverages the Empirical Feature Matrix (EFM) to mitigate drift in feature representations by identifying important directions for previous tasks to reduce forgetting while enhancing plasticity for learning new tasks (Section~\ref{sec:empirical_feat}). In this phase, the feature extractor $f_t$ and the current task classifier with weights $W_{t-1:t}$ are trained with EFM regularization and cross-entropy loss (Section~\ref{sec:feature-learning-no-rehearsal}) \textbf{(b)} After training, EFC++ uses the EFM to update the prototypes of previous task classes based on the drift induced by the most recent task (Section~\ref{sec:drift-compensation}). \textbf{(c)} EFC++ uses Gaussian prototypes, together with current task features, for training previous and current task classifiers with weights $W_t = [W_{t-1},W_{t-1:t}]$  via a prototype re-balancing phase (Section~\ref{sec:proto-balance}).
    \textbf{(d)} Before training on the next task, the new EFM and the prototypes of the current task classes are computed.}
    \label{fig:EfC}
\end{figure*}

\section{Related Work}
\label{sec:related_work}
The primary goal of incremental learning is to continuously acquire new knowledge from a non-stationary data stream while mitigating the issue of catastrophic forgetting. In this study, we focus on \textit{offline}  \textit{class incremental} learning problems. The class-incremental scenario is distinguished from the task-incremental one in that no task information is available at test time and thus is more challenging~\cite{vandeVen2022,9349197}. Offline
incremental learning allows multiple passes on task datasets during training, while online incremental learning considers continuous data streams in which each sample is accessed just once during training~\cite{NIPS2017_f8752278,NEURIPS2019_15825aee,NEURIPS2022_5ebbbac6,soutifcormerais2023comprehensive,Aljundi_2019_CVPR}. Recently, more general lifelong learning formulations have also been proposed beyond the class-incremental setting considered in this work~\cite{gido-tpami}, including continual learning for vision-brain understanding across subjects~\cite{Nguyen2026COBRA}.

A significant portion of offline class incremental learning research focuses on Exemplar-Based Class Incremental Learning (EBCIL) approaches. These methods integrate regularization techniques with replay mechanisms, utilizing a memory of previous class samples that are replayed during the learning of new tasks. iCaRL~\cite{icarl} employs Nearest Mean Classification along with exemplars and uses Knowledge Distillation~\cite{knowledge_dist} during training to mitigate forgetting. Techniques like BiC~\cite{Wu_2019_CVPR} and IL2M~\cite{Belouadah_2019_ICCV} correct network outputs using exemplars to  mitigate inter-task confusion and task-recency bias.  DER++ \cite{der++} and its recent variant X-DER~\cite{9891836} replay both logits and ground-truth labels via self-distillation on exemplars. LUCIR~\cite{ucir}, PODNet~\cite{podnet}, and AFC~\cite{adaptive_feature_consolidation} combine cosine normalization and distillation losses with exemplars to reduce forgetting. Another category of exemplar-based approaches focuses on expanding sub-network structures across incremental steps. AANET~\cite{aanet} adds two residual blocks to mask layers, balancing stability and plasticity. DER~\cite{der} trains a single backbone for each incremental task, while MEMO~\cite{memo} shares generalized residual blocks and extends specialized blocks for new tasks to improve performance and efficiency. 

Recent exemplar-based works further analyze the role of replay scheduling, buffer size, and rehearsal strategies from both empirical and theoretical perspectives~\cite{Kang_2025_CVPR,lanzillotta2026heads,deng2025unlocking}. This research has also expanded to more challenging areas like the video domain. For instance Jiang et al.~\cite{Jiang2026TeacherAgent}, develop a Teacher Agent approach that minimizes rehearsal overhead by substituting the teacher network with a lightweight agent that generates soft labels directly. The main drawbacks of exemplar-based approaches are the lack of privacy preservation and high computational and storage costs, especially when using a growing memory buffer or when dynamically expanding the network.

Exemplar-free Class Incremental Learning (EFCIL) mitigates catastrophic forgetting without relying upon exemplars from previous tasks. This setting has also been explored in application-specific domains such as traffic sign classification, where storing past samples may be undesirable in deployed systems~\cite{exemplar-free-trafficlight}. Early works such as EWC~\cite{ewc,huszar}, Rwalk~\cite{rwalk}, MAS~\cite{mas}, SI~\cite{pmlr-v70-zenke17a}, focused on calculating importance scores for all neural network weights. These importance scores are then used to regularize the network training, ensuring that the current learned weights do not update parameters crucial to the performance of previous task. Benzing et al.~\cite{pmlr-v151-benzing22a} show that the importance score computed by the aforementioned methods approximates the Fisher Information Matrix. Recently, Liu and Chang~\cite{Liu_2026_CVPR} showed that modifying the logits during the computation of the Fisher Information Matrix can improve EWC weight regularization by producing more reliable parameter importance estimates.

Another research direction, exemplified by methods like Learning without Forgetting (LwF)~\cite{lwf} and Less-forgetful Learning (LfL)~\cite{lfl}, utilizes functional regularization techniques such as knowledge and feature distillation respectively. These methods constrain network activations to mirror those of a previously trained network, a strategy that has been shown to surpass weight regularization in performance~\cite{facil}. Most recent functional regularization approaches use feature distillation~\cite{il2a,sdc,pass,ssre,evanescent,generative_features}, or even freeze the feature extractor after the first task~\cite{fetril,goswami2023fecam,deesil,HE2026114901}, greatly reducing the plasticity of the network. For this reason, they predominantly consider Warm Start scenarios in which the first task is much larger than the others, so that reducing plasticity after the first task has less impact on the performance.  

On top of this decrease in plasticity, merely using functional regularization is not enough to learn new boundaries between classes from previous tasks and classes from new ones. To address this issue without using exemplars, a line of research employs generative models~\cite{generative_features,NIPS2017_0efbe980,Xiang_2019_ICCV,smith2021always,ayub2021eec,gao2022r,WANG2025107053} to create synthetic samples from previous tasks and replay them while learning the new tasks. However, training generative models concurrently with new tasks is inefficient, and these models are prone to forgetting~\cite{pass}. Consequently, recent approaches combine regularization with \textit{prototype rehearsal}~\cite{pass,il2a}.  Prototypes are feature space statistics (typically class means) used to reinforce the boundaries of previous-task classes without the need to preserve exemplars. PASS~\cite{pass}  and IL2A~\cite{il2a} propose the integration of Gaussian prototypes and self-supervised learning to learn transferable features for future tasks. Evanescent~\cite{evanescent} learns and updates prototype representations during incremental learning. SSRE~\cite{ssre} replays  over-sampled prototypes during the current task training and employs a strategy to reorganize the network structure to transfer invariant knowledge across tasks. FeTrIL~\cite{fetril} fixes the feature extractor after the initial large task and generates pseudo-samples, combining prototypes with current task features, to train a linear model that discriminates all observed classes. Similarly, FeCAM~\cite{goswami2023fecam} freezes the backbone after the first task, but uses class prototypes and covariances to classify samples with the Mahalanobis distance.

In our previous work, Elastic Feature Consolidation (EFC)~\cite{magistri2024elastic}, we revisited the \textit{Cold Start} scenario for EFCIL, showing the need for greater plasticity throughout incremental learning. Since then, recent works have addressed this setting through drift compensation~\cite{ldc,he2025semantic}, old-class recovery from current task data~\cite{resurrecting}, or class statistics adaptation under feature drift~\cite{task-recency}. In this extension of our previous work we further analyze the challenges of Cold Start and the limitations of EFC, and propose EFC++, a training strategy that better couples representation learning and classifier adaptation in regimes requiring high plasticity.


\section{Feature Drift Regularization via the Empirical Feature Matrix}
\label{sec:feature-drift-regularization}
In this section, we introduce the general EFCIL framework and then derive a pseudo-metric in feature space, induced by a positive semi-definite matrix we call the \textit{Empirical Feature Matrix} (EFM), used as a regularizer to control feature drift during class-incremental learning.

\subsection{Exemplar-free Class-Incremental Learning (EFCIL)}
During class-incremental learning, a model $\mathcal{M}_t$ is sequentially trained on $K$ tasks, each characterized by a disjoint set of classes  $\{\mathcal{C}_t\}_{t=1}^K$, where each $\mathcal{C}_t$ is the set of labels associated with the task $t$. We denote the incremental dataset as $\mathcal{D} = \{\mathcal{X}_t, \mathcal{Y}_t\}_{t=1}^K$, where $\mathcal{X}_t$ and $\mathcal{Y}_t$ are, respectively, the set of samples and the set of labels for task $t$. The incremental model $\mathcal{M}_t$ at task $t$ consists of a feature extraction backbone $f(\cdot; \theta_t)$ shared across all tasks and whose parameters $\theta_t$ are updated during training, and a classifier $W_t \in \mathbb{R}^{n \times  \sum_{j=1}^t \vert  \mathcal{C}_j \vert }$ which grows with each new task. Let $\sigma$ be the softmax activation function, the model output at task $t$ is the composition of feature extractor and classifier:
\begin{equation}
    \mathcal{M}_t(x; \theta_t, W_t) \equiv p(y | x; \theta_t, W_t) = \sigma (W_t^{\top} f(x; \theta_t)).
\end{equation} 
For notation purposes, we define \(\mathbf{X} \sim \mathcal{X}_t\) as a batch of current task data with labels \(\mathbf{Y} \sim \mathcal{Y}_t\), sampled from the overall dataset, and \(f(\mathbf{X}; \theta_t)\) as the feature extractor output on the mini-batch. Thus, the model output at task \(t\) for the mini-batch \(\mathbf{X}\) is:
\begin{equation}
    \mathcal{M}_t(\mathbf{X}; \theta_t, W_t) \equiv p(\mathbf{Y} | \mathbf{X}; \theta_t, W_t) = \sigma (W_t^{\top} f(\mathbf{X}; \theta_t)).
\end{equation} 
Directly training $\mathcal{M}_t$ (i.e. fine-tuning) with a cross-entropy loss at each task $t$ results in backbone drift because at task $t$ we see no examples from previous-task classes. This progressively invalidates the previous-task classifiers and results in forgetting. This issue is closely related to feature forgetting in continually learned representations~\cite{hess2024knowledge}. In addition to the cross-entropy loss, many state-of-the-art EFCIL approaches use a regularization loss to reduce backbone drift, and a prototype loss to adapt previous-task classifiers to new features.
 
\subsection{Weight Regularization and Feature Distillation}
\label{ewc_fd}
 Kirkpatrick et al.~\cite{ewc} showed that using $\ell_2$ regularization to constrain weight drift reduces forgetting but does not leave enough plasticity for learning new tasks. Hence, they proposed Elastic Weight Consolidation, which relaxes the $\ell_2$ constraint with a quadratic constraint based on a diagonal approximation of the Empirical Fisher Information Matrix (E-FIM):
 \begin{equation}
   F_{t} \!=\underset{x\sim \mathcal{X}_{t}}{ \mathbb{E}}  \biggr[\underset{y\sim p(y|x;\theta^{*}_{t})}{\mathbb{E}}\biggl[ \biggl( \frac{ \partial \log p(y)} {\partial \theta^{*}_{t}} \biggl)  \biggl( \frac{\partial \log p(y)  } {\partial \theta^{*}_{t}} \biggl)^\top \biggl] \biggr],  
  \label{eq:fisher}
 \end{equation}
where $\theta_t^*$ is the model trained after task $t$. The E-FIM induces a pseudo-metric in parameter space~\cite{amari} encouraging parameters to stay in a low-error region for previous-task models: 
\begin{equation}
\begin{gathered}
\mathcal{L}_{t}^{\text{E-FIM}} \vcentcolon =  \lambda_{\text{E-FIM}} (\theta_t - \theta^{*}_{t-1})^\top  F_{t-1} (\theta_t - \theta^{*}_{t-1}).
\end{gathered}
\end{equation}
The main limitation of approaches of this type is the need for approximations of the E-FIM (e.g. a diagonal assumption). This makes the computation of the E-FIM tractable, but in practice is unrealistic since it does not consider off-diagonal entries that represent influences of interactions between weights on the log-likelihood. To overcome these limitations, more recent EFCIL approaches~\cite{pass,ssre} rely on feature distillation (FD), which scales better in the number of parameters:  
\begin{equation}
\label{eq:fd}
\mathcal{L}_{t}^{\text{FD}} \vcentcolon =  \sum_{x \in \mathbf{X}} || f(x; \theta_t) - f(x; \theta_{t-1}) ||_2,
\end{equation}
where $\mathbf{X}\sim\mathcal{X}_t$.

The isotropic regularizer using the $\ell_2$ distance is too harsh a constraint for learning new tasks~\cite{podnet}. We propose to use a pseudo-metric in \textit{feature space}, induced by a matrix which we call the \textit{Empirical Feature Matrix} (EFM), which constrains directions in \textit{feature space} most important for previous tasks, while allowing more plasticity in other directions when learning new tasks. As shown in Figure~\ref{fig:probe}, increasing plasticity is particularly relevant in Cold Start incremental learning since, in the initial task, a high-quality feature representation cannot be learned and methods must be plastic to adapt their backbone to new tasks.

\subsection{The Empirical Feature Matrix}
\label{sec:empirical_feat}

\begin{figure*}[t]

    \centering
    \begin{minipage}{0.3\textwidth}
        \centering
        \includegraphics[width=\linewidth]{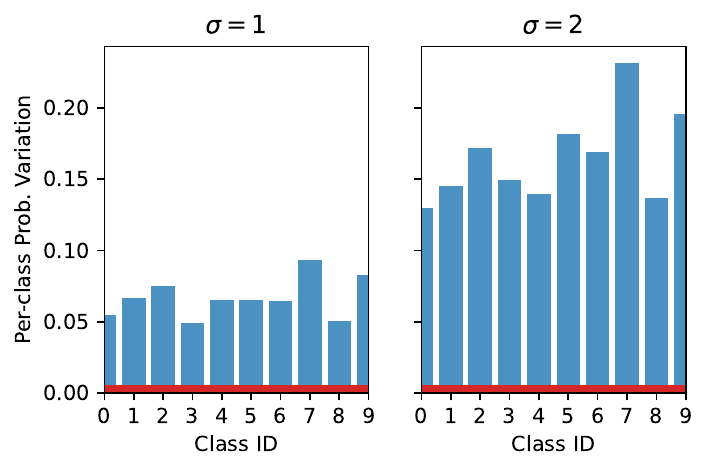}
    \end{minipage}%
    \hspace{3mm}
    \begin{minipage}{0.3\textwidth}
        \centering
        \includegraphics[width=\linewidth]{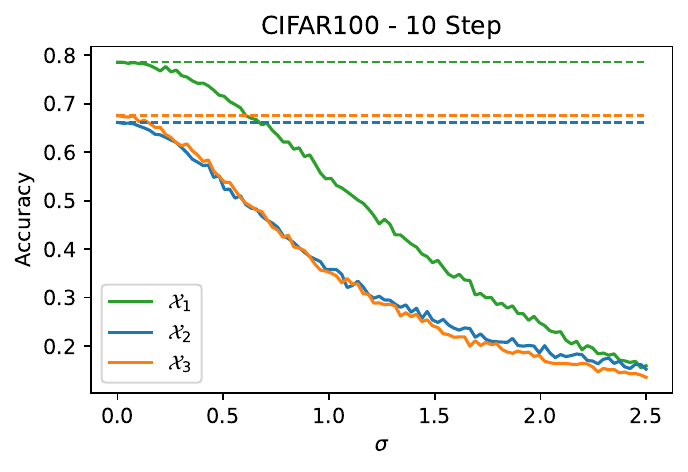}
    \end{minipage}%
    \hspace{3mm}
    \begin{minipage}{0.3\textwidth}
        \centering
        \includegraphics[width=\linewidth]{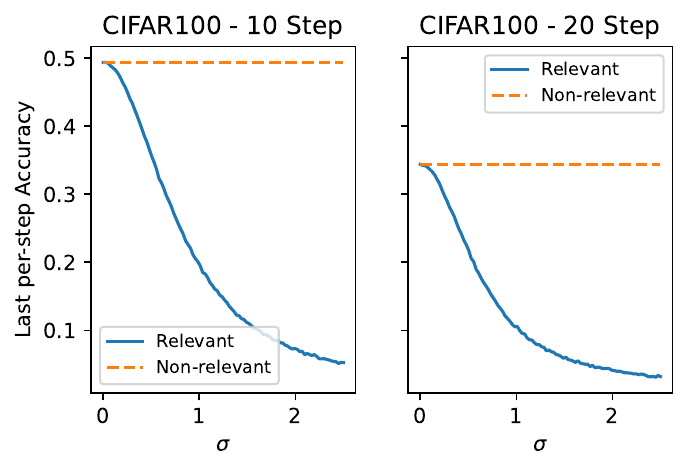}
    \end{minipage}
    \caption{The regularizing effects of $E_t$ on the Cold Start CIFAR-100 - 10 and 20 step scenarios (see Section \ref{sec:experimental_setting} for details on dataset settings). \textbf{Left}: Perturbing features in the principal directions of $E_1$ results in significant changes in classifier outputs (in blue\color{black}), while perturbations in non-principal directions leave the outputs unchanged (in red\color{black}). \textbf{Middle}: If we continue incremental learning up through task 3 and perturb features from all three tasks in the principal (solid lines) and non-principal (dashed lines) directions of $E_3$, we see that $E_3$ captures \emph{all} important directions in feature space up through task 3. \textbf{Right}: At the end of training, we observe the same behavior: the last per-step accuracy (see Eq.~\ref{eq:metrics}), representing the average accuracy over all tasks after the last training session, decreases only when perturbed in directions of $E_{10}$ or $E_{20}$ relevant for previous tasks in the 10-step and 20-step scenarios, respectively.}
    \label{fig:experimental-evidence}
\end{figure*}

Our goal is to regularize feature drift, but we do not want to do so isotropically as in feature distillation. Instead, we draw inspiration from how the E-FIM identifies important directions in parameter space and take a similar approach in feature space (see Figure \ref{fig:EfC}a).

\minisection{The local and empirical feature matrices.}
Let $f(x;\theta_t)$ be the feature vector extracted from input $x \in \mathcal{X}_t$ from task $t$ and $p(y)=p(y|f(x;\theta_t); W_t)$ the discrete probability distribution over all the classes seen so far. We define the \textit{local feature matrix} as:
\begin{equation}
\label{eq:localfm}
   E_{f(x;\theta_t)} = \underset{y\sim p(y)}{\mathbb{E}}\biggl[ \biggl( \frac{ \partial \log p(y) } {\partial f(x;\theta_t)} \biggl)  \biggl( \frac{\partial \log p(y)  } {\partial f(x;\theta_t)} \biggl) ^\top \biggl].
\end{equation}
The Empirical Feature Matrix (EFM) associated with task $t$ is obtained by taking the expected value of Eq.~\ref{eq:localfm} over the entire dataset at task $t$:
\begin{equation}
\label{efm}
   E_{t} = \underset{x\sim \mathcal{X}_{t}}{\mathbb{E}}  [E_{f(x;\theta_t)}].
 \end{equation}
Both $E_{f(x;\theta_t)}$ and $E_t$ are symmetric, as they are weighted sums of symmetric matrices, and positive semi-definite. Naively computing  $E_{f(x;\theta_t)}$ requires a complete forward pass and a backward pass up to the feature embedding, however we exploit an analytic formulation that avoids computing gradients:
\begin{equation}
\label{eq:closed_form_efm}
E_{f(x;\theta_t)} = \underset{y\sim p(y)}{\mathbb{E}}[W_t(I_m-P)_y (W_t(I_m-P)_y)^{\top}],
\end{equation}
where $m = \sum_{j=1}^t \vert  \mathcal{C}_j \vert$, $I_m$ the identity matrix of dimension $m \times m$, and $P$ is a matrix built from the row of softmax outputs associated with $f(x;\theta_t)$ replicated $m$ times (see Appendix B for a derivation).

To gain additional insight into what this matrix measures, we highlight some (information) geometry aspects of $E_t$. The Empirical Fisher Information Matrix provides information about the information geometry of parameter space. In particular, it can be obtained as the second derivative of the KL-divergence for small weight perturbations~\cite{naturalgradient,datamatrix}. Similarly, the Empirical Feature Matrix provides information about the information geometry of \textit{feature} space. Since $E_{f_t(x)}$ is positive semi-definite, we can interpret it as a pseudo-metric in feature space and use it to understand which perturbations $\delta$ of features most affect the predicted probability:
\begin{equation}
\begin{aligned}
\label{kl_features}
    & \text{KL}(p(y \mid f(x;\theta_t) + \delta; W_t) \,\, || \,\, p(y \mid f(x,\theta_t); W_t)) \\& = \delta^\top E_{f(x;\theta_t)} \delta + \mathcal{O}(||\delta||^{3}).
\end{aligned}
\end{equation}

\minisection{The Empirical Feature Matrix loss.} We now have all the ingredients for our regularization loss. Let  $\delta(x) =f(x;\theta_t) - f(x; \theta_{t-1})$ be the drift between the features of sample $x$ extracted from the current model $\mathcal{M}_t$ and the model $\mathcal{M}_{t-1}$ trained on the previous task, then
\begin{equation}
\label{eq:training_loss}
\mathcal{L}_{t}^{\text{EFM}} \vcentcolon =  \mathop{\mathbb{E}}\limits_{x \in \mathbf{X}}\! \left[\delta(x)^\top (\lambda_{\text{EFM}} E_{t\text{-}1} +\eta I)\delta(x) \right],
\end{equation}
where $\mathbf{X} \sim \mathcal{X}_t$.
In Eq.~\ref{eq:training_loss} we use a damping term $\eta \in \mathbb{R}$ to constrain features to stay in a region where the quadratic second-order approximation of the KL-divergence of the log-likelihood in Eq.~\ref{kl_features} remains valid~\cite{Martens2012TrainingDA}. However, to ensure that our regularizer does not degenerate into feature distillation, we must ensure that $\lambda_{\text{EFM}} \nu_i > \eta$, where $\nu_i$ are the non-zero eigenvalues of the EFM. Additional analysis of these constraints and the spectrum of the EFM are given in Section \ref{sec:eigenval-analysis}. 

It should be noted that the regularization in Eq.~\ref{eq:training_loss} does not explicitly suppress perturbations along the principal directions of $E_{t-1}$, but rather discourages movement along them. The term  $\delta(x)^{\top}  E_{t-1} \delta(x) = \allowbreak \delta(x)^{\top} U_{t-1} \Lambda_{t-1} U_{t-1}^T$ $\delta(x)$, highlights this, where $U_{t-1}$ is the eigenvector matrix of $E_{t-1}$ and $\Lambda_{t-1}$ is the corresponding diagonal matrix of eigenvalues sorted from larger to smaller. The penalty is larger when $\delta(x)$ aligns with the eigenvectors of $E_{t-1}$ associated with larger eigenvalues in $\Lambda_{t-1}$. Conversely, the loss is lower when $\delta(x)$ aligns with eigenvectors corresponding to smaller eigenvalues. As a result, updates are naturally biased toward directions of lower significance since they incur a smaller penalty. The damping term $\eta \delta(x)^{\top} I\delta(x)$ stabilizes updates and limits drift in the directions associated with lower eigenvalues. In the next section we give empirical evidence of the effect of this elastic regularization.

\minisection{The Regularizing effects of the EFM.}
By definition, eigenvectors of the EFM associated with strictly positive eigenvalues correspond to feature directions where perturbation most significantly impact the predictions. To empirically validate this, we perturb features along these \textit{principal} directions and measure the variation in probability outputs. 

Let $f_t=f(x; \theta_t) \in \mathbb{R}^n$ be the feature vector extracted from $x$, and $E_t$ denote the EFM after training on task $t$. We define a perturbation vector $\varepsilon_{1:k} \in \mathbb{R}^n$, where the first $k$ entries are sampled from a Gaussian $\mathcal{N}(0, \sigma)$, with $k$ being the number of strictly positive eigenvalues of the spectral decomposition of $E_t$. The remaining $n\text{-}k$ entries are set to zero. Then, letting $U_t$ be the matrix whose columns are the eigenvectors of $E_t$, we compute the perturbed features vector $\tilde{f_t}$ as follows:
\begin{equation}
    \tilde{f_t} = f + U_t^{\top} \varepsilon_{1:k}.
\end{equation}
In Figure~\ref{fig:experimental-evidence} (left), perturbing the feature space along these principal directions after training the first task, leads to a substantial variation in class probabilities. Conversely, applying the complementary perturbation $\varepsilon_{k:n}$, i.e., setting zero on the first $k$ directions and applying Gaussian noise on the remaining $n\text{-}k$ directions, has no impact on the output. In Figure~\ref{fig:experimental-evidence} (middle) we see that, after training three tasks, perturbing in the principal directions of $E_3$ degrades performance across all tasks, while perturbations in non-principal directions continue to have no effect on the accuracy. This shows that $E_3$ captures the important feature directions for all previous tasks and that regularizing drift in these directions should mitigate forgetting. 

Finally, the rightmost plot in Figure~\ref{fig:experimental-evidence} illustrates the per-step accuracy at the end of the training, reflecting the overall continual learning performance (see Section~\ref{sec:experimental_setting} for a formal definition). As observed earlier, perturbing features along the principal directions of $E_{10}$ (or $E_{20}$) rapidly decreases performance, while perturbations in non-principal directions have no impact on the overall accuracy. This confirms that this behavior remains consistent across all tasks.

\section{Elastic Feature Consolidation with Prototype Re-balancing}
\label{sec:efc++}
Regularizing  the feature space via the EFM in combination with a classification loss on current task classes is not sufficient by itself in exemplar-free class incremental learning. Although EFM limits feature drift along directions important for previous tasks, the final classifier must still discriminate among all the classes observed so far. This is challenging because the classifier $W_{t-1}$ is  extended to include the new $|\mathcal{C}_t|$ classes, yielding $W_t=[W_{t-1}, W_{t-1:t}]$. Trained only on current task samples, it cannot  separate old and new classes, leading to \textit{inter-task confusion} and \textit{task-recency bias}~\cite{facil,zhuang2022acil,Wu_2019_CVPR}.
 
In this section, we introduce the notation for Gaussian prototype rehearsal, which we use for classifier adaptation between old and new classes. We revisit the PR-ACE loss in EFC and discuss why coupling prototype rehearsal with feature extractor training can be problematic in the high-plasticity Cold Start regime. Finally, we introduce Elastic Feature Consolidation with Prototype Re-balancing (EFC++), which decouples feature learning from inter-task classifier adaptation through a post-training prototype re-balancing phase.

\subsection{Gaussian Prototypes for Classifier Adaptation}
Let $\mathcal{C}_t$ be the set of classes introduced at task $t$, and let
$\mathcal{C}_{1:t-1} \vcentcolon= \bigcup_{s=1}^{t-1} \mathcal{C}_s$ denote the set of classes observed before task $t$.
For each previous class $c \in \mathcal{C}_{1:t-1}$, we denote by $p^c_{t-1}$ the prototype stored before learning task $t$. When a class $c \in \mathcal{C}_s$ is first introduced at task $s$, with $s \in \{1,\ldots, t-1\}$ its prototype is initialized as the empirical mean of its deep features~\cite{pass}, while its corresponding covariance $\Sigma^c$ is computed from the same features~\cite{generative_features}. The resulting class-level statistics are retained across incremental steps in the prototype buffer $P_{1:t-1}$.

During classifier adaptation at task $t$, we rehearse previous classes by sampling Gaussian-perturbed prototypes $\tilde{p}^{\,c}_{t-1} \sim \mathcal{N}(p^c_{t-1}, \Sigma^c)$. These Gaussian prototypes provide synthetic samples at feature level for previous classes and allow the classifier weights $W_t$ to be trained over old and current classes \textit{without} storing real images.

\subsection{EFC and the Limitations of PR-ACE in Cold Start}

Elastic Feature Consolidation (EFC)~\cite{magistri2024elastic} combines the empirical feature matrix loss $\mathcal{L}_t^{\text{EFM}}$ (Eq.~\ref{eq:training_loss}), inhibiting drift in important directions for previous tasks, with the \textit{Asymmetric Prototype Replay} (PR-ACE) loss:
\begin{equation}
      \mathcal{L}^{\text{EFC}}_t  \vcentcolon = \mathcal{L}_t^{\text{EFM}} + \mathcal{L}_t^{\text{PR-ACE}}, 
\end{equation}
where
\begin{equation}
\label{eqn:pr-ace}
\mathcal{L}_t^{\text{PR-ACE}} \! \vcentcolon = \! \mathcal{L}^{\text{ce}}_t(f(\mathbf{X},\theta_t); W_{t\text{-}1:t}) \!+\! \mathcal{L}^{\text{ce}}_t(\tilde{\mathbf{P}} \cup f(\hat{\mathbf{X}}, \theta_t); W_{t}).
\end{equation}
Here $\mathbf{X}, \hat{\mathbf{X}} \sim \mathcal{X}_t$ are two distinct batches of current task data, while $\tilde{\mathbf{P}}$ is a batch of augmented prototypes, where each prototype $\tilde{p}_{t-1}^c \in \tilde{\mathbf{P}}$ is sampled from the Gaussian $\mathcal{N}(p_{t-1}^c, \Sigma^c)$. The term $\mathcal{L}^{\text{ce}}_t(f(\mathbf{X},\theta_t); W_{t-1:t})$ is a cross-entropy loss restricted to the current task classes, while the term  with prototypes $\mathcal{L}^{\text{ce}}_t(\tilde{\mathbf{P}} \cup f(\hat{\mathbf{X}}, \theta_t); W_{t})$ affects old and current task classifier weights $W_t=[W_{t-1}, W_{t-1:t}]$. PR-ACE samples prototypes and images uniformly over all seen classes.

After each task, EFC updates the stored prototypes with the EFM to partially compensate for backbone drift and computes the prototypes, covariances, and EFM used in subsequent tasks; see Appendix~C for the full EFC algorithm. We detail the EFM-based prototype update in Section~\ref{sec:drift-compensation}, where it also becomes a key component of EFC++.

\minisection{Limitations of PR-ACE in Cold Start.} \begin{figure}[t]
    \centering
    \includegraphics[width=\columnwidth]{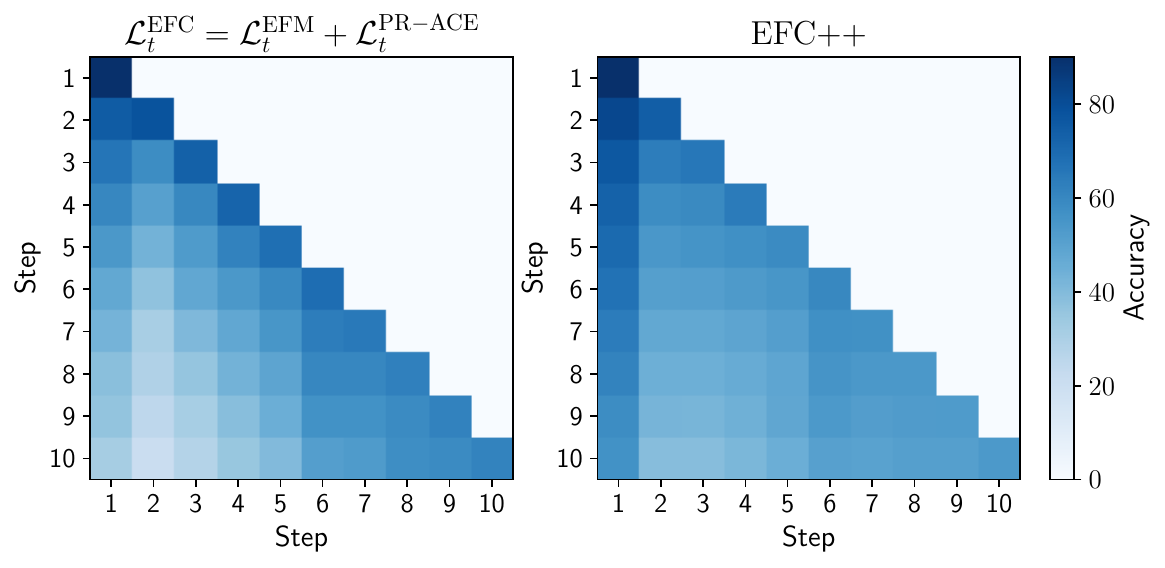} 
    \caption{Accuracy after each incremental step on the Cold Start CIFAR-100 10-step scenario. \textbf{Left}: In EFC, which combines EFM regularization with the asymmetric PR-ACE loss to balance current task data with prototypes during training, older tasks are forgotten more quickly than more recent ones.  \textbf{Right}: EFC++, which applies EFM regularization during backbone training and a post-training prototype re-balancing phase, achieves a better plasticity-stability trade-off.} 
    \label{fig:efc-sym}
\end{figure}
\label{sec:limitations-prace}
We identify limitations in using PR-ACE for training the final classifier. Since PR-ACE is applied during feature extractor training, the stored prototypes for previous classes gradually lose their representativeness of the true class means over the course of current task epochs. This occurs because the feature space itself is evolving over time as a result of optimization, leading to a shift in the alignment between stored prototypes and their actual class means. Consequently, the classifier is trained with prototypes that increasingly deviate from the actual class means.

This misalignment is further compounded by the cross-entropy term  $\mathcal{L}^{\text{ce}}_t(\tilde{\mathbf{P}} \cup f(\hat{\mathbf{X}}, \theta_t); W_{t})$  on the right-hand side of Equation~\ref{eqn:pr-ace}, which induces an \textit{additional drift in the feature representation} along the relevant directions for previous tasks identified by the EFM. This drift arises from the backward pass on $\hat{\mathbf{X}}$, which updates the weights $\theta_t$ and alters the feature representations. Because PR-ACE uniformly samples prototypes $\tilde{\mathbf{P}}$ and new data $\hat{\mathbf{X}}$ based on class labels, mini-batches contain a larger proportion of current task data in earlier tasks compared to later tasks. Consequently, classes introduced in earlier incremental steps undergo stronger feature drift during their own training stage (see Figure \ref{fig:drift-efc-efc++} for drift analysis).

For illustration, consider the Cold Start scenario where each task introduces an equal number of classes. In the second task, PR-ACE forms batches containing 50\% prototypes and 50\% image data; in the third task, this shifts to two-thirds prototypes and one-third current data. Because the cross-entropy loss is averaged over both prototypes and new data, the gradient updates on the feature extractor become less severe for later tasks. As more tasks are added, the drift in the feature representation along the relevant directions of the EFM diminishes. Consequently, prototypes from earlier tasks remain farther from the actual class means than those from more recent tasks, leading to forgetting. This effect is illustrated in Figure~\ref{fig:efc-sym} (left), where in EFC the performance on the first five tasks declines faster than on the last five. 

Managing the drift caused by PR-ACE is particularly crucial in the Cold Start scenario, where, during initial training, a strong backbone cannot be learned and thus representations are more susceptible to change compared to the Warm Start regime (see Section~\ref{sec:drift-ws-cs}). As we will show in the next section, EFC++ mitigates this issue by removing prototype replay from feature extractor training and postponing inter-task classifier adaptation to a separate post-training phase.

\subsection{Feature Learning without Prototype Rehearsal}
\label{sec:feature-learning-no-rehearsal}
We decouple the regularization phase via the Empirical Feature Matrix and the phase of learning the final inter-task classifier. In the first stage (Figure~\ref{fig:EfC}a), we train \textit{the feature extractor} and \textit{the current task classifier} with the following training loss: 
\begin{equation}
\label{eq:train-EFC++}
     \mathcal{L}_t^{\text{train}} \vcentcolon = \mathcal{L}_t^{\text{EFM}}  +  \mathcal{L}^{\text{ce}}_t(f(\mathbf{X},\theta_t); W_{t-1:t}).
\end{equation}
where $\mathbf{X}\sim\mathcal{X}_t$  and \(W_{t-1:t}\) represents the weights of the classifier associated with the latest classes. The cross-entropy loss is evaluated only on the current task classes, and is used to learn the features of the current task. \(\mathcal{L}_t^{\text{EFM}}\) is the regularization loss based on Empirical Feature Matrix (see Eq.~\ref{eq:training_loss}) and is used to mitigate feature drift in the relevant directions for the previous task while allowing plasticity in the other directions.

\subsection{Prototype Drift Compensation via the EFM}
\label{sec:drift-compensation}
 After feature learning, the backbone has changed from $\theta_{t-1}$ to $\theta_t$. Therefore, the prototypes stored before learning task $t$ may no longer be aligned with the current feature space. Yu et al.~\cite{sdc} showed that the drift in \textit{embedding networks} can be estimated using current task data. They suggest estimating the drift of class means from previous-tasks by looking at feature drift in closely related classes from the current task. They leverage this drift to update the class means in nearest-mean classification.  

Let $p^c_{t-1}$ be the prototype of the class $c$ at task $t-1$. Our aim is to update $p^c_{t-1}$ after the training of task $t$ to account for backbone drift. These updated prototypes are  used for  prototype rehearsal for training the model classifier. As proposed by SDC~\cite{sdc}, we update the prototypes after the backbone is trained for each task as follows: 
\begin{equation}
\label{eq:sdc_formula}
      p^c_{t\text{-}1} \leftarrow p^c_{t\text{-}1} + \frac{\underset{x_i \sim \mathcal{X}_t}{\sum}  w_i \delta_i}{\underset{x_i \sim \mathcal{X}_t}{\sum} w_i},\, \, \,
      \delta_i=f(x_i;\theta_t) - f(x_i,\theta_{t\text{-}1}),  
\end{equation}
where $\delta_i$ is the feature drift of samples $x_i$ between task $t-1$ and $t$.

We define the weights $w_i$ using the EFM (Figure~\ref{fig:EfC}b). In Section~\ref{sec:empirical_feat} we showed that the EFM induces a pseudo-metric in feature space, providing a second-order approximation of the KL-divergence due to feature drift. We extend the idea of feature drift estimation to softmax-based classifiers by weighting the overall drift of our class prototypes. Writing $p^c = p^c_{\text{t}-1}$ and $f_{t\text{-}1}(x_i) = f(x_i;\theta_{t-1})$ to simplify notation, our weights are defined as:
\begin{equation}
\label{our_weight}
\begin{aligned}
w_i \! ~&= \! \exp \left( -\dfrac{(f_{t\text{-}1}(x_i)  - p^c) E_{t\text{-}1} (f_{t\text{-}1}(x_i) - p^c)^\top}{2 \sigma^2}\right)  \\ & \approx \!
 \exp \left( -\dfrac{ \text{KL}(p(y|f_{t\text{-}1}(x_i); W_{t\text{-}1}) \, || \, p(y|p^c; W_{t\text{-}1}))}{2 \sigma^2}\right)
\end{aligned}
\end{equation}
where $E_{t\text{-}1}$ is the EFM after training task $t-1$. Eq.~\ref{our_weight} assigns higher weights to the samples more closely aligned to prototypes in terms of probability distribution. Specifically, higher weights are assigned to samples whose softmax prediction matches that of the prototypes, indicating a strong similarity for the classifier.

\subsection{Inter-task Classifier Adaptation via Prototype Re-balancing}
\label{sec:proto-balance}
During the post-training stage, which we call \textit{prototype re-balancing} phase (Figure~\ref{fig:EfC}c), we train all classifiers up to the current task using the prototypes updated using Eq.~\ref{eq:sdc_formula} and freezing the feature extractor:
\begin{equation}
\label{eq:post_training-efc++}
    \mathcal{L}_t^{\text{post-train}} \vcentcolon = \mathcal{L}^{\text{ce}}_t(\tilde{\mathbf{P}} \cup f(\mathbf{X}, \theta^*_t); W_{t}),
\end{equation}
where \(\theta_t^*\) represents the frozen weights of the feature extractor and $W_t=[W_{t-1}, W_{t-1:t}]$  represents the learnable weights of the final classifier for the previous and current task classes. In this phase, we construct mini-batches by mixing prototypes, $\tilde{p}_{t-1}^c \in \tilde{\mathbf{P}}$, sampled from a Gaussian distribution $\mathcal{N}(p_{t-1}^c, \Sigma^c)$,  with current task representations obtained from the frozen feature extractor $f(\mathbf{X},\theta_t^*)$. The covariance matrix $\Sigma^c$ represents the feature covariance of class $c$, computed in the task where class $c$ was introduced. Prototypes and current task data are uniformly sampled from all encountered classes to ensure balanced representation. Since the goal is to enable the final classifier to distinguish among classes from different tasks, the cross-entropy loss is evaluated over all encountered classes.

\subsection{Elastic Feature Consolidation++} In summary, the training process begins with feature extractor training, where we optimize Eq.~\ref{eq:train-EFC++} to learn the new task while simultaneously regularizing relevant directions from previous tasks (Figure~\ref{fig:EfC}a). Once the feature extractor is trained, we update the prototypes using the EFM (see Eqs.~\ref{eq:sdc_formula},~\ref{our_weight} in Section~\ref{sec:drift-compensation}), ensuring that representations from the previous task remain aligned with the current task drift (Figure~\ref{fig:EfC}b). 

After updating the prototypes, we perform the \textit{prototype re-balancing} phase to train the classifier (Eq.~\ref{eq:post_training-efc++}). Since the prototypes have already been aligned with the updated feature extractor, this guarantees that the classifier is trained with consistent representations (Figure~\ref{fig:EfC}c). Finally, new prototypes and their corresponding class covariances are computed. The Empirical Feature Matrix $E_t$ is then calculated using the current task data and is used as regularizer for the next incremental task (Figure~\ref{fig:EfC}d).  

We call this procedure, detailed in Algorithm~\ref{alg:train}, Elastic Feature Consolidation++ (EFC++) to highlight its extension of our previous approach, Elastic Feature Consolidation (EFC), introduced in~\cite{magistri2024elastic}.

\RestyleAlgo{ruled}
 \DontPrintSemicolon
 \begin{algorithm}[t]
  \SetNoFillComment
 \caption{EFC++: Elastic Feature Consolidation with Prototype Re-balancing}
 \label{alg:train}
        \KwData{ $\mathcal{M}_1$, $E_1$, $P_1$}
 \For{$t=2, \dots T$}{
   Initialize $\theta_t \leftarrow \theta_{t-1}$ and extend the classifier 
$W_t=[W_{t-1},W_{t-1:t}]$ \;
    \tcc{Backbone Training}
\For{\text{each optimization step}}{
        sample $\mathbf{X} \sim \mathcal{X}_t$ \;

        compute  $\mathcal{L}^{\text{EFM}}_t$ \,\, (Eq.~\ref{eq:training_loss}) \;

        $\mathcal{L}_t^{\text{train}} \!\!\! = \! \mathcal{L}_t^{\text{EFM}}  \! \! + \!   \mathcal{L}^{\text{ce}}_t(f(\mathbf{X},\theta_t); W_{t\text{-}1:t})$ (Eq.~\ref{eq:train-EFC++}) \;
         Update  $\theta_t \gets \theta_t - \alpha_1\frac{\partial \mathcal{L}_t^{\text{train}}}{\partial \theta_t}$ \;

        Update  $W_{t-1:t} \gets W_{t-1:t} - \alpha_2\frac{\partial \mathcal{L}_t^{\text{train}}}{\partial W_{t-1:t}}$ \;
         
     }
     Set $\theta_t^* \leftarrow \theta_t$ \tcc{\!\!freeze the feature extractor\!\!\!} \;
     \vspace{-0.6em}
\tcc{Prototype Update}
Update $P_{1:t-1}$ using Eqs.~\ref{eq:sdc_formula},~\ref{our_weight} \;     
    \vspace{0.5em}
\tcc{Prototype Re-balancing}
    \For{\text{each optimization step}}{
        sample $\mathbf{X} \sim \mathcal{X}_t$, $\tilde{\mathbf{P}} \sim P_{1:t-1}$ \;

        $\mathcal{L}_t^{\text{post-train}} = \mathcal{L}^{\text{ce}}_t(\tilde{\mathbf{P}} \cup f(\mathbf{X}, \theta^*_t); W_{t})$ \, (Eq.~\ref{eq:post_training-efc++}) 
        
        Update  $W_t \gets W_t -  \alpha_3\frac{\partial \mathcal{L}_t^{\text{post-train}}}{\partial W_t}$
     }
      Compute $P_t$ from current-task features \;
       $ P_{1:t}   \gets P_{1:t\text{-}1} \cup P_t$ \;
        \vspace{0.5em}
       
     $E_t \gets \text{EFM}(\mathcal{X}_t,   \mathcal{M}_t )$ (Eq. \ref{eq:closed_form_efm})      \tcc{\!\!Compute EFM \!\!\!} \;
 }
 \end{algorithm}

\minisection{Addressing PR-ACE limitations.}
EFC++ addresses the limitations of PR-ACE by decoupling the feature extractor training (Eq.~\ref{eq:train-EFC++}) from inter-task classifier adaptation, which is instead performed after feature learning through prototype re-balancing (Eq.~\ref{eq:post_training-efc++}). This approach ensures that no additional drift occurs in the feature extractor while learning the current task, unlike in PR-ACE. Moreover, because EFC++ updates the prototypes before the post-training phase, the final classifier is trained using prototypes that more accurately reflect the real old class representations than those used by PR-ACE in EFC. As a result, \textit{EFC++ more effectively controls drift in directions relevant to previous task classifiers without compromising plasticity} (see Figure \ref{fig:efc-sym}, right). 

To validate this claim, we consider a model continuously trained using either EFC or EFC++ up to task $t-1$. We then measure the drift of the class means before and after training task $t$. Specifically, for each class $c \in \mathcal{C}_t$, we denote the feature representation of class $c$ prior to task $t$ by $\mu^c_{t-1}$ and after training task $t$ by $\mu^c_{t}$. These representations are obtained by averaging the feature representations of the data for class $c$. We define the feature drift of the class mean as the difference between these feature representations, specifically,
\begin{equation}
\label{eq:local_drift}
    \Delta\mu^c = \mu^c_{t} - \mu^c_{t-1}.
\end{equation}
The squared pseudo-norm of $\Delta\mu^c$ via $E_t$, denoted as
\begin{equation}
\label{eq:norm_drift}
    \vert \vert \Delta\mu^c \vert \vert^2_{E_t} = (\Delta\mu^c)^\top E_t \Delta\mu^c,
\end{equation}
measures the magnitude of the drift of the class mean along the relevant directions identified after the training of task $t$. We then define the drift of task $t$ as the average of the drifts of the class means, expressed as
\begin{equation}
\label{eq:total_norm_drift}
    \Delta = \dfrac{1}{\vert \mathcal{C}_t \vert} \sum_{c \in \mathcal{C}_t} \vert \vert \Delta\mu^c \vert \vert^2_{E_t}.
\end{equation}

\begin{figure}[t]
   \includegraphics[width=\columnwidth]{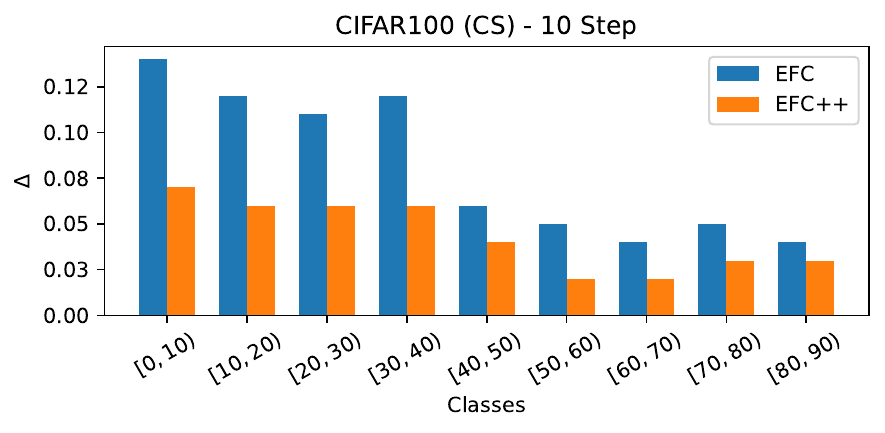}
    \caption{Average drift of the class means in the relevant directions of the EFM before and after training the task in which they are involved, in both EFC and EFC++ on CIFAR-100 (CS) 10-step. EFC++ consistently exhibits less drift than EFC, especially in the initial tasks, where the drift of the classes is more pronounced (double) for EFC. This confirms that EFC++ better controls the drift along relevant directions.
}
    \label{fig:drift-efc-efc++}
\end{figure}

Figure~\ref{fig:drift-efc-efc++} illustrates that, with EFC, the drift in the relevant directions due to PR-ACE is more pronounced in the earlier tasks compared to EFC++. In EFC++, decoupling the feature extractor training phase from the inter-task classifier learning phase allows for better control of the drift in the relevant directions. The reduction in drift along these directions does not compromise learning on the new classes. This is evidenced in Figure \ref{fig:efc-sym}, which shows that the values on the diagonal of the confusion matrix for both EFC and EFC++ are comparable.

\section{Experimental Settings}
\label{sec:experimental_setting}
In this section, we describe the experimental setup used to evaluate Elastic Feature Consolidation++ (EFC++) against the state-of-the-art in Exemplar-free Class Incremental Learning (EFCIL).

\minisection{Datasets.} We evaluate EFC++ on five distinct datasets, with their statistics detailed in Appendix D.1. We categorize CIFAR-100~\cite{CIFAR100}, Tiny-ImageNet~\cite{TinyImageNet}, and ImageNet-Subset~\cite{ImageNet} as \textit{small-scale} datasets, containing 100, 200, and 100 classes, respectively. For a more comprehensive \textit{large-scale} scenario, we use ImageNet-1K~\cite{ImageNet}, which includes 1000 diverse object classes with varying numbers of samples per class. Additionally, we utilize DN4IL, a well-crafted and balanced subset of the standard DomainNet dataset~\cite{gowda2023dual}. DN4IL consists of 600 classes, with 100 classes per domain, specifically \textit{clipart}, \textit{infograph}, \textit{painting}, \textit{quickdraw}, \textit{real}, and \textit{sketch}. This dataset is instrumental in comparing our approach to the state-of-the-art under \textit{large domain shifts}.

\begin{table}
\centering
\caption{Class Distribution in Warm Start and Cold Start: $|\mathcal{C}_0|$ is the number of classes in the pre-training phase, while $|\mathcal{C}_{i>0}|$ represents classes introduced per incremental step.
}
\setlength{\tabcolsep}{3.5pt}
\resizebox{0.99\linewidth}{!}{%
\begin{tabular}{lcccc|cccc}
\toprule
\multirow{3}{*}{\textbf{Dataset}} & \multicolumn{4}{c|}{\textbf{Warm Start (WS)}} & \multicolumn{4}{c}{\textbf{Cold Start (CS)}} \\
 
                         & \multicolumn{2}{c}{\textbf{10 Step}} & \multicolumn{2}{c|}{\textbf{20 Step}} & \multicolumn{2}{c}{\textbf{10 Step}} & \multicolumn{2}{c}{\textbf{20 Step}} \\ 
 
                         & $\vert \mathcal{C}_0 \vert$ & $\vert \mathcal{C}_{i>0} \vert$ & $\vert \mathcal{C}_0 \vert$ & $\vert \mathcal{C}_{i>0} \vert$ & $\vert \mathcal{C}_0 \vert$ & $\vert \mathcal{C}_{i>0} \vert$ & $\vert \mathcal{C}_0 \vert$ & $\vert \mathcal{C}_{i>0} \vert$ \\
\cmidrule(lr){1-1} \cmidrule(lr){2-3}  \cmidrule(lr){4-5} \cmidrule(lr){6-7} \cmidrule(lr){8-9}  
CIFAR-100                & 50  & 5   & 40  & 3   & 0   & 10  & 0   & 5   \\ 
 
Tiny-ImageNet            & 100 & 10  & 100 & 5  & 0  & 20  & 0  & 10  \\ 
 
ImageNet-Subset        & 50  & 5   & 40  & 3   & 0  & 10  & 0   & 5   \\          
 
ImageNet-1K              & 500 & 50  & 400 & 30  & 0 & 100 & 0  & 50  \\ \bottomrule
\end{tabular}
}
\label{tab:ws-cs-statistics}
\end{table}
\minisection{Warm and Cold Start. } We consider two scenarios for evaluating all approaches. The first, which we refer to as the \textbf{Warm Start (WS)} scenario which is commonly considered in the EFCIL literature~\cite{pass,ssre,fetril,evanescent}, involves pre-training the model on $50\%$ or $40\%$ of all available classes, denoted as $\mathcal{C}_0$. The remaining classes are then equally distributed among the subsequent incremental learning steps. This setting, which begins with a larger task favors continual learning techniques that prioritize \textit{stability} in retained knowledge over the \textit{plasticity} required to learn new classes. 

 The second scenario, which we call \textbf{Cold Start (CS)}, follows the setting in EFC~\cite{magistri2024elastic}. In this scenario, the model is randomly initialized before incremental learning, and all available classes are equally split among the incremental learning steps, resulting in no initial pre-training phase. Consequently, the number of classes per step is larger compared to Warm Start. See Table~\ref{tab:ws-cs-statistics} for a detailed description of how the classes are distributed in Warm Start and Cold Start across all datasets.

We are primarily interested in the Cold Start scenario since it demands high backbone plasticity and poses a significant challenge for EFCIL. Mitigating forgetting becomes difficult without using exemplars when larger backbone shifts occur (in Section~\ref{sec:drift-ws-cs} we provide an empirical analysis of this phenomenon).

 \minisection{Metrics. }We use  the \textbf{per-step incremental accuracy} $A_{\text{step}}^K$ and \textbf{average incremental accuracy} $A_{\text{inc}}^K$:
\begin{equation}
\label{eq:metrics}
    A_{\text{step}}^K = \frac{\sum_{i=s}^K \vert \mathcal{C}_i \vert  a^K_i}{\sum_{i=s}^K \vert \mathcal{C}_i \vert}
   , \,\, A_{\text{inc}}^K = \dfrac{1}{K-s+1} \sum_{i=s}^K  A_{\text{step}}^i,
\end{equation}
where $a^K_i$ represents the accuracy on task $i$ after training task $K$, and $s=1$ in the Cold Start scenario and $s=0$ in the Warm Start scenario. It should be noted that in the Warm Start (WS) scenario, the performance on the initial classes, indicated as step $s=0$, is commonly included in the evaluation metric~\cite{pass,evanescent,ssre,fetril,goswami2023fecam}, and we follow this same approach. Therefore, achieving higher stability on the large number of initial classes significantly impacts the final evaluation. In contrast, in the Cold Start (CS) scenario the classes are equally distributed among the incremental learning tasks and the evaluation metrics starting from $s=1$ give equal importance to each step.

To gain a more comprehensive understanding of impact of different regularization techniques employed by the tested methods, we use \textbf{average forgetting} $F^{K}$ to measure stability during training and \textbf{cumulative average accuracy on the last task} $\text{PL}^{K}$ to measure plasticity (i.e. the average of the diagonal of the matrix in Figure \ref{fig:efc-sym})~\cite{lange2023continual}. More specifically, the forgetting of a task is the difference between its maximum accuracy obtained in the past and its current accuracy:
\begin{equation}
f_j^K = \max_{i \in \{s,...,k-1\}} (a_j^i-a_j^K ), \quad j < K.
\end{equation}
Thus, $F^{K}$ and $\text{PL}^{K}$ at the $K$-th task are defined as:
 \begin{equation}
 \label{forg-plasticity}
     F^{K} = \dfrac{1}{K-s} \sum_{j=s}^{K-1}f_j^K ,  \,\,\,\, \text{PL}^{K} = \frac{1}{K-s+1}\sum_{i=s}^K  a^i_i.
 \end{equation}

\minisection{Hyperparameter settings. }We use standard ResNet-18~\cite{resnet} trained from scratch for all experiments. We train the first task of each state-of-the-art method using the same optimizer, number of epochs, and data augmentations. In particular, we train the first task using self-rotation as in \cite{pass,evanescent} to align the performance of all techniques.

For the incremental steps of EFC++ we used Adam with a weight decay of 2e-4 and a fixed learning rate of 1e-4 for Tiny-ImageNet and CIFAR-100, while for ImageNet-Subset, ImageNet-1K, DN4IL we use a learning rate of 1e-5 for the backbone and 1e-4 for the heads. We fixed the total number of epochs to 100 and use a batch size of 64, except for ImageNet-1K where the batch size is fixed to 256. We set $\lambda_{\text{EFM}}=10$ and $\eta=0.1$ in Eq.~\ref{eq:training_loss} for all the experiments. In Section~\ref{sec:eigenval-analysis}, we discuss the theoretical basis for selecting these hyperparameters and analyze their impact on performance. For EFC++, during post-training we use SGD with a fixed learning rate 1e-3 to train the final classifier on both prototype and current task features for 50 epochs.

We ran all approaches using five random seeds and shuffling the classes in order to reduce the bias induced by the choice of class ordering~\cite{icarl,facil}. In Appendix D.2 we provide the optimization settings for each state-of-the art method we evaluated.

\section{Experimental Results}
\label{sec:experimental-results}
 In this section we compare EFC++ with state-of-the-art Exemplar-free Class-Incremental Learning (EFCIL) methods across small-scale Class-Incremental (Class-IL) benchmarks (CIFAR-100, Tiny-ImageNet, and ImageNet Subset), large-scale Class-IL ImageNet-1K, and the domain- and class-incremental DN4IL benchmark.  

We organize the discussion around two  main questions: whether EFC++ remains competitive in the standard Warm Start setting despite the increased plasticity induced by the EFM, and whether its advantage increases in the more challenging Cold Start setting. We mainly discuss the last per-step accuracy $A_{\text{step}}^K$,  while the average incremental accuracy $A_{\text{inc}}^K$ follows the same trends.
 
\subsection{Small-scale Class-IL Experiments}
\label{sec:comp_sota}

\begin{table*}[t]
\centering
\caption{Small-scale Class-IL experiments. Fusion results are those reported in \cite{evanescent} as no code is provided to reproduce them. The result for ImageNet-Subset Cold Start for ABD and R-DFCIL are those reported in~\cite{gao2022r}.}

\setlength{\tabcolsep}{6pt}
\resizebox{\textwidth}{!}{%
\begin{tabular}{cl c cc c|c cc cc }
 
\toprule
 & & \multicolumn{4}{c}{\textbf{Warm Start (WS)}} &  \multicolumn{4}{c}{\textbf{Cold Start (CS)}} \\
 
\multicolumn{2}{l}{} & \multicolumn{2}{c}{  $\boldsymbol{A_{\mathbf{step}}^K}$ } & \multicolumn{2}{c}{$\boldsymbol{A_{\mathbf{inc}}^K}$} & \multicolumn{2}{c}{$\boldsymbol{A_{\mathbf{step}}^K}$ } & \multicolumn{2}{c}{$\boldsymbol{A_{\mathbf{inc}}^K}$} \\

  & \textbf{Method}    & \textbf{10 Step}  & \textbf{20 Step}  & \textbf{10 Step}  & \textbf{20 Step}  & \textbf{10 Step}  & \textbf{20 Step}   & \textbf{10 Step}  & \textbf{20 Step}   \\
\cmidrule(lr){2-2}
\cmidrule(lr){3-4}
\cmidrule(lr){5-6}
\cmidrule(lr){7-8}
\cmidrule(lr){9-10}
\multirow{11}{*}{\rotatebox[origin=c]{90}{CIFAR-100}} 
      & EWC   & $21.08 \pm 1.09$ & $13.53 \pm 1.11$ & $41.00 \pm 1.11$ & $31.79 \pm 2.77$ & $31.17 \pm 2.94$  &$17.37 \pm 2.43$  & $49.14 \pm 1.28$ & $31.02 \pm 1.15$   \\
      & LwF   & $20.73 \pm 1.47$ & $11.78 \pm 0.60$ & $41.95 \pm 1.30$ & $28.93 \pm 1.62$ & $32.80 \pm 3.08$& $17.44 \pm 0.73$ & $53.91 \pm 1.67$  & $38.39\pm 1.05$ &    \\
      & PASS  & $53.42 \pm 0.48$ & $47.51 \pm 0.37$ & $63.42 \pm 0.69$ & $59.55 \pm 0.97$ & $30.45 \pm 1.01$& $17.44 \pm 0.69$ & $47.86\pm 1.93$ & $32.86 \pm 1.03$ & \\
      
      & Fusion & $56.86$ & $51.75$ & $65.10$   & $61.60$  & $-$ & $-$ & $-$   & $-$ &     \\
      
      & FeTrIL  & $56.79 \pm 0.40$ & $52.61 \pm 0.81$  & $65.03 \pm 0.66$   & $62.50 \pm 1.03$  &$34.94 \pm 0.46$ & $23.28 \pm 1.24$  & $51.20 \pm 1.13$   & $38.48 \pm 1.07$  &  \\
      & SSRE     & $57.48 \pm 0.55$ & $52.98 \pm 0.63$  & $65.78 \pm 0.59$   & $63.11 \pm 0.84$  &  $30.40	\pm 0.74$  & $17.52 \pm 0.80$  & $47.26 \pm 1.91 $  &$32.45	\pm 1.07$  &  \\
      & FeCAM  & $\underline{62.13}\pm 0.51$ & $\textbf{58.74} \pm 0.60$ &    $\underline{68.87}\pm0.40$ &$\textbf{67.42}\pm0.88$   &  $37.63\pm0.52$ &$23.82\pm0.87$&$52.53\pm1.19$& $38.51\pm0.82$  \\
      & ABD\footnotemark[2] & $44.91 \pm 1.00$ & $33.22 \pm 1.11$ & $56.81 \pm 1.45$ & $47.55 \pm 0.19$ & $43.11 \pm 0.60 $ & $28.93 \pm 0.21$ & $\underline{59.14} \pm 1.72$ & $47.12 \pm 0.85$ \\
      & R-DFCIL\footnotemark[2] & $52.24 \pm 0.32$ & $40.97 \pm 0.12$ & $63.09 \pm 0.21$ & $55.65 \pm 0.10$ & $42.14 \pm 0.54$ & $30.35 \pm 0.05$ & $57.77 \pm 0.50$ & $\underline{47.41} \pm 0.27$ \\
      & EFC    & $ 60.87 \pm 0.39 $& $55.78 \pm 0.42$  & $68.23 \pm 0.68$   & $ 65.90 \pm 0.97$ &  $\underline{43.62} \pm 0.70$  & $\underline{32.15} \pm 1.33$  & $58.58 \pm 0.91$   &$47.36 \pm 1.37$  &  \\
\cmidrule(lr){2-2}
\cmidrule(lr){3-4}
\cmidrule(lr){5-6}
\cmidrule(lr){7-8}
\cmidrule(lr){9-10}
        & EFC++  & $\textbf{62.15} \pm 0.48 $ & $\underline{57.55} \pm 0.66 $ & $\textbf{69.26} \pm 0.63 $ & $\underline{67.11} \pm 0.91 $ & $\textbf{47.52} \pm 0.68 $ & $\textbf{33.71} \pm 1.41 $ & $\textbf{61.57} \pm 0.66 $ & $\textbf{49.20} \pm 1.10 $ \\
\cmidrule[0.6pt](l){1-10}

\multirow{11}{*}{\rotatebox[origin=c]{90}{Tiny-ImageNet}}    
    & EWC       & $\phantom{0}6.73 \pm 0.44$ & $\phantom{0}5.96 \pm 1.17$  & $18.48 \pm 0.60$ & $13.74 \pm 0.52$ & $\phantom{0}8.00 \pm 0.27$ & $\phantom{0}5.16 \pm  0.54$ & $24.01 \pm 0.51$ & $15.70 \pm 0.35$  \\
    & LwF       & $24.00 \pm 1.44$ & $\phantom{0}7.58 \pm 0.37$ & $43.15 \pm 1.02$ & $22.89 \pm 0.64$ & $ 26.09 \pm 1.29$ & $15.02\pm 0.67$ & $45.14 \pm 0.88$ &  $32.94 \pm 0.54$   \\
    & PASS      & $41.67 \pm 0.64$ & $35.01 \pm 0.39$ & $51.18 \pm 0.31$  &$46.65 \pm 0.47$ & $ 24.11 \pm 0.48$ & $18.73\pm1.43$ & $39.25\pm 0.90$ &               $32.01 \pm 1.68$
                  \\
    & Fusion &$46.92$ & $44.61$ & $-$ & $-$ &  $-$  & $-$ & $-$   & $-$ \\
    & FeTrIL     & $45.71 \pm 0.39$ & $44.63 \pm 0.49$ & $53.95 \pm 0.42$ & $52.96 \pm 0.45$ & $30.97 \pm 0.90$ & $25.70 \pm 0.61$  & $45.60 \pm 1.67$   & $39.54 \pm 1.19$  &  \\
    & SSRE       & $44.66 \pm 0.45$ & $44.68 \pm 0.36$ & $53.27 \pm 0.43$ & $52.94 \pm 0.42$ & $22.93 \pm 0.95 $  & $17.34 \pm 1.06$   &$38.82 \pm 1.99$  & $30.62 \pm 1.96$  \\
    & FeCAM  & $\underline{50.77}\pm0.62$ & $\textbf{50.73}\pm0.46$ &   $\underline{57.67}\pm0.48$ &	$\underline{57.57}\pm0.37$ & $30.79\pm0.69$&$25.36\pm0.74$&$45.00\pm1.35$&$38.59\pm1.29$ \\
    & ABD\footnotemark[2] & $24.07 \pm 1.08$ & $11.17 \pm 1.30$ & $44.92 \pm 1.41$ & $34.70 \pm 1.38$ & $23.24 \pm 0.57$ & $12.21 \pm 0.65$ & $42.80 \pm 1.07$ & $33.37 \pm 1.20$ \\
    & R-DFCIL\footnotemark[2] & $43.06 \pm 0.12$ & $36.55 \pm 0.06$ & $53.39 \pm 0.20$ & $49.89 \pm 0.25$ & $\underline{35.49} \pm 0.25$ & $28.23 \pm 0.17$ & $\underline{49.14} \pm 0.35$ & $\underline{42.30} \pm 0.17$ \\
    & EFC   &  50.40 $\pm 0.25$  & $ 48.68\pm 0.65$  & $ 57.52 \pm 0.43$   & $ 56.52 \pm 0.53$  &  $ 34.10 \pm 0.77 $  & $\underline{28.69} \pm 0.40$  & $47.95 \pm 0.61$   & $42.07 \pm 0.96$  &  \\
\cmidrule(lr){2-2}
\cmidrule(lr){3-4}
\cmidrule(lr){5-6}
\cmidrule(lr){7-8}
\cmidrule(lr){9-10}
    & EFC++  & \textbf{51.67} $\pm 0.31$ & \underline{50.41} $\pm 0.51$ & \textbf{58.54} $\pm 0.51$ & \textbf{57.79} $\pm 0.54$ & \textbf{37.48} $\pm 0.52$ & \textbf{32.56} $\pm 0.44$ & \textbf{50.18} $\pm 1.11$ & \textbf{45.58} $\pm 1.32$ \\

\cmidrule[0.6pt](l){1-10}

\multirow{11}{*}{\rotatebox[origin=c]{90}{ImageNet-Subset}} 
      & EWC      & $16.19 \pm 2.48$ & $10.66 \pm 1.74$ & $23.58 \pm 2.01$ & $18.05 \pm 1.10$ & $24.59 \pm 4.13 $ & $12.78 \pm 1.95 $ &$39.40\pm 3.05 $  & $26.95 \pm 1.02$ &   \\
      & LwF       & $21.89 \pm 0.52$ & $13.24 \pm 1.61$ & $37.15 \pm 2.47$ & $25.96 \pm 0.95$ & $37.71 \pm 2.53   $& $18.64 \pm 1.67   $ & $56.41 \pm 1.03 $  & $40.23\pm 0.43   $ &    \\
      & PASS     & $52.04 \pm 1.06$ & $44.03 \pm 2.19$ & $65.14 \pm 0.36$  &$58.88 \pm 2.15$ & $26.40 \pm 1.33$& $ 14.38\pm 1.22$ & $ 45.74\pm 0.18$ & $31.65 \pm 0.42$ &
          \\
      & Fusion  &$60.20$ & $51.60$ & $70.00$   & $63.70$  &  $-$  & $-$ & $-$   & $-$  &   \\
      & FeTrIL   & $63.56 \pm 0.59$ & $57.62 \pm 1.13$  & $71.87 \pm 1.46$   & $68.01 \pm 1.60$  &  $36.17 \pm  1.18$ & $26.63 \pm  1.45$  & $ 52.63 \pm 0.56$   & $ 42.43\pm 2.05$  &  \\
      & SSRE     & $61.84 \pm 0.93$ & $55.19 \pm 0.97$  & $70.68 \pm 1.37$   & $66.73 \pm 1.61$  &  $25.42 \pm 1.17$  & $ 16.25 \pm 1.05  $  & $43.76 \pm 1.07$   &$31.15 \pm 1.53$  &  \\
      & FeCAM & $67.68\pm0.45$& $\underline{62.74}\pm 1.04$ &   $74.59\pm0.94$& $\underline{71.72}\pm1.45$  & $35.54\pm1.20$&$26.03\pm1.33$&$51.90\pm1.29$&$40.87\pm2.00$ \\
      & ABD & $38.68 \pm 0.76$ & $21.47 \pm 2.08$ & $55.72 \pm 1.06$ & $40.92 \pm 2.30$ & $35.96 $ & $22.40$ & $57.76$ & $44.89$ \\
      & R-DFCIL & $49.81 \pm 0.23$ & $38.19 \pm 0.13$ & $64.50 \pm 0.31$ & $56.8 \pm 0.69$ & $42.28$ & $30.28$ & $59.10$ & $47.33$ \\
        & EFC    & $\underline{68.85} \pm 0.58$  & $ 62.17 \pm 0.69$  & $\underline{75.40} \pm 0.92$   & $ 71.63 \pm 1.13$  &  $\underline{47.38} \pm 1.43$  & $\underline{35.75} \pm 1.74$  & $\underline{59.94} \pm   1.38$   &$\underline{49.92} \pm 2.05$  &  \\
      
\cmidrule(lr){2-2}
\cmidrule(lr){3-4}
\cmidrule(lr){5-6}
\cmidrule(lr){7-8}
\cmidrule(lr){9-10}
      & EFC++  & \textbf{69.28} $\pm 0.46$ & \textbf{62.75} $\pm 0.46$ & \textbf{75.71} $\pm 1.00$ & \textbf{72.05} $\pm 1.33$ & \textbf{53.90} $\pm 1.18$ & \textbf{40.86} $\pm 1.65$ & \textbf{65.90} $\pm 1.62$ & \textbf{55.42} $\pm 2.44$ \\

\bottomrule
\end{tabular}
}
\label{tab:sota}
\end{table*}
\footnotetext[2]{ABD and R-DFCIL are run on CIFAR-100 and Tiny-ImageNet with ResNet-18 for consistency, although the original papers used ResNet-32 except on ImageNet-Subset.}

Table~\ref{tab:sota} compares EFC++ with baseline and recent state-of-the-art EFCIL approaches, including EWC~\cite{ewc}, LwF~\cite{lwf}, PASS~\cite{pass}, Fusion~\cite{evanescent}, FeTrIL~\cite{fetril}, SSRE~\cite{ssre}, FeCAM~\cite{goswami2023fecam}, ABD~\cite{smith2021always}, R-DFCIL~\cite{gao2022r}, and EFC~\cite{magistri2024elastic}. 
The comparison covers both Warm Start (WS) and Cold Start (CS), using protocols with 10 and 20 incremental learning steps. All methods are evaluated under the same first-task training and data augmentation protocol. 
This may lead to differences from originally reported SSRE and FeCAM results, especially in WS where the large first task strongly affects the final metrics  (Eq. \ref{eq:metrics})\cite{pycl,endtoend}. 
Further implementation details and per-step plots are reported in Appendices D.1, J.1, and J.2.

\minisection{Warm Start.} EFC++ consistently improves over methods based on feature distillation and prototype rehearsal, such as PASS, Fusion, and SSRE, as well as over FeTrIL, which freezes the feature extractor, with gains typically ranging from 2 to 6 percentage points. Methods based on generative inversion, such as ABD and R-DFCIL, perform worse than both feature distillation methods and methods that freeze the backbone, suggesting that inversion is less effective when the large first task dominates the incremental protocol. Compared to EFC, EFC++ achieves consistent but moderate improvements. For CIFAR-100 and Tiny-ImageNet, the gains over EFC are approximately 1.3\% and 1.7\% in the 10-step and 20-step settings, respectively, while they are smaller on ImageNet-Subset. This indicates that prototype re-balancing is still beneficial in WS, although its impact is reduced compared to Cold Start because the stronger initial backbone limits feature drift and, consequently, the misalignment between prototypes and features induced by PR-ACE in EFC. This trend is further analyzed in Section~\ref{sec:drift-ws-cs}.

\begin{figure*}[t]
    \centering
    \includegraphics[width=\textwidth]{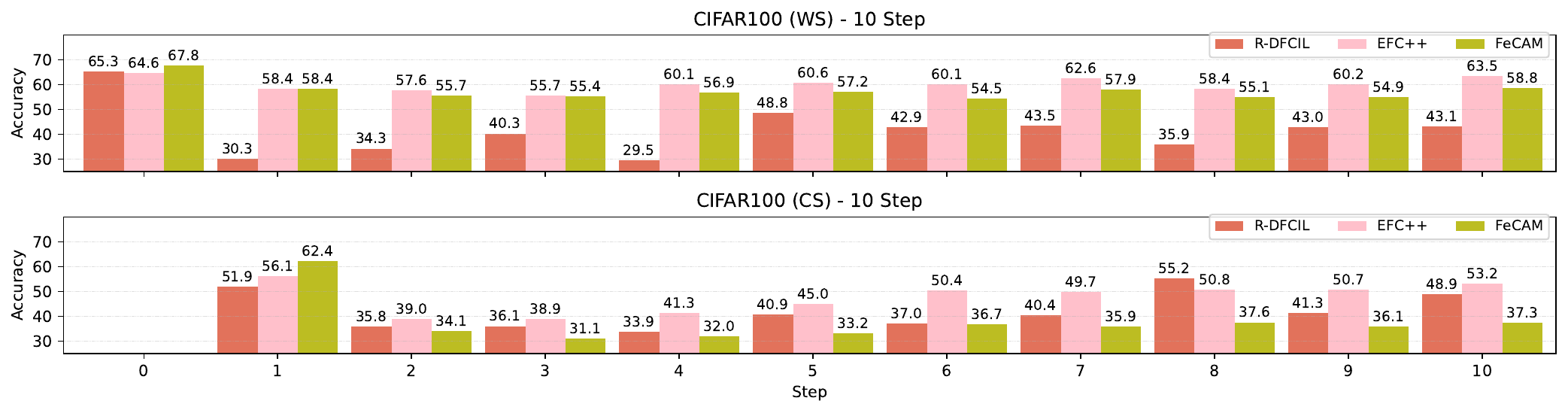}
    \caption{Accuracy on each task after the final training step on CIFAR-100 Warm Start and Cold Start. EFC++ achieves the best stability-plasticity trade-off compared to the nearest competitor in Warm Start (FeCAM) and Cold Start (R-DFCIL).  Task 0 represents the large first task in WS, which is absent in CS (see Table~\ref{tab:ws-cs-statistics})}
    \label{fig:cifar_bars}
\end{figure*}

Finally, EFC++ remains competitive with FeCAM, the strongest competitor in WS. The two methods are comparable on CIFAR-100, while on Tiny-ImageNet and ImageNet-Subset, EFC++ improves performance by about 1.2\% in the 10-step setting and remains comparable in the 20-step setting. This is noteworthy because FeCAM freezes the backbone after the large first task, which preserves first task accuracy and strongly influences the per-step accuracy in WS. In contrast, EFC++ keeps updating the feature extractor through EFM regularization, improving plasticity for later tasks while maintaining competitive stability on the initial task. The WS panel of Figure~\ref{fig:cifar_bars} confirms this behavior: FeCAM better preserves the large initial task, while EFC++ obtains stronger accuracy on the other tasks after the last training step.

\minisection{Cold Start. }In Cold Start (Table~\ref{tab:sota}, right), methods based on feature distillation, such as SSRE and PASS, and methods that freeze the feature extractor, such as FeTrIL and FeCAM, show a substantial drop in accuracy compared to EFC++. This confirms that, when no strong initial representation is available, excessive stability limits the ability to learn later tasks. Most of these methods also perform worse than LwF, despite using prototypes for classifier adaptation, suggesting that prototypes alone are not sufficient when the backbone lacks plasticity.

Interestingly, R-DFCIL reverses its behavior with respect to Warm Start: despite its weak WS performance, it becomes the strongest non-EFC baseline in CS. This suggests that its knowledge distillation loss increases plasticity, improving the learning of new tasks and therefore the final per-step accuracy, where all tasks contribute equally to the metric in CS (see Eq.~\ref{eq:metrics}). Nevertheless, EFC++ outperforms R-DFCIL across all benchmarks in both incremental protocols: by 5\% and 3\% on CIFAR-100, 2\% and 4\% on Tiny-ImageNet, and 11\% and 10\% on ImageNet-Subset for the 10-step and 20-step settings, respectively. The larger gap on ImageNet-Subset suggests that the generative mechanism used by R-DFCIL to rehearse old class samples becomes more challenging as image resolution increases ($224 \times 224$ for ImageNet-Subset). The CS panel of Figure~\ref{fig:cifar_bars} confirms this trend: after the final incremental step, EFC++ achieves higher accuracy than R-DFCIL across all tasks, while FeCAM achieves better performance only on the first task.

Compared with EFC, EFC++ achieves consistent and substantial gains of approximately 3--4 percentage points across all CS settings. This improvement is due to the decoupling of backbone training from classifier head balancing. 
In EFC, prototypes are used while the feature space is still evolving, so they can become outdated during training. EFC++ avoids this issue by using the updated prototypes only after backbone training, which is especially important in Cold Start, where feature drift is significant (see Section~\ref{sec:drift-ws-cs} for the drift comparison in WS and CS).

\subsection{Large-scale Domain- and Class-IL Experiments}

\begin{table}[t]
\centering
\caption{Large scale  Class- and Domain-IL  Experiments on ImageNet-1K (CS) and DN4IL~\cite{gowda2023dual} (CS) using $A_{\text{step}}^K$.}
\label{tab:merged-classdomain-il}
\setlength{\tabcolsep}{5pt}
\resizebox{\linewidth}{!}{%
\begin{tabular}{l c c c}
\toprule
& \multicolumn{2}{c}{\textbf{ImageNet-1K (CS)}} 
& \textbf{DN4IL (CS)} 
\\

\textbf{Method} 
& \textbf{10 Step} 
& \textbf{20 Step} 
& \textbf{6 Step} 
\\
\cmidrule(lr){1-1}
\cmidrule(lr){2-3}
\cmidrule(lr){4-4}
FeTrIL  
& $34.28 \pm 0.20$ 
& $26.64 \pm 0.25$ 
& $30.95 \pm 0.17$
\\
FeCAM   
& $36.16 \pm 0.10$ 
& $27.24 \pm 0.19$
& $36.42 \pm 0.12$ 
\\
R-DFCIL 
& $29.36 \pm 0.17$ 
& $22.30 \pm 0.19$ 
& $22.31 \pm 0.21$
\\
EFC     
& $\underline{42.62} \pm 0.08$ 
& $\underline{36.32} \pm 0.35$ 
& $\underline{38.27} \pm 0.16$
\\
\cmidrule(lr){1-1}
\cmidrule(lr){2-3}
\cmidrule(lr){4-4}
EFC++   
& $\textbf{44.72} \pm 0.11$ 
& $\textbf{37.46} \pm 0.03$ 
& $\textbf{39.32} \pm 0.15$
\\
\bottomrule
\end{tabular}
}
\end{table}

We next evaluate EFC++ on ImageNet-1K to assess its scalability to a large-scale class-incremental setting, and on DN4IL (a subset of DomainNet) to test its performance in a more challenging scenario involving both class-incremental and domain-incremental shifts. Here, we report the final per-step accuracy in CS; the average incremental accuracy in CS and WS experiments are provided in Appendix~F, and follow the same trends observed on ImageNet-Subset.

\minisection{Experiments on ImageNet-1K}. EFC++ achieves the best per-step accuracy in both the 10-step and 20-step CS settings, improving over EFC by 2.1 and 1.2 percentage points, respectively. Freezing the feature extractor limits the plasticity of FeTrIL and FeCAM, leading to drops of more than 6 and 8 percentage points, respectively, compared to EFC and EFC++. Although R-DFCIL performs well in small-scale experiments, it underperforms FeTrIL and FeCAM in this setting, likely because its generative inversion network must synthesize high-resolution images for 1,000 distinct classes by the end of the training, ten times more than ImageNet-Subset.

\minisection{Experiments on DN4IL (DomainNet)}. As shown in Table~\ref{tab:merged-classdomain-il},  EFC++ improves upon EFC by approximately 1.1\% in final per-step accuracy and outperforms the remaining state-of-the-art approaches. Notably, EFC++ surpasses the best competitor, FeCAM, by roughly 3 points. R-DFCIL struggles here for reasons similar to those observed on ImageNet-1K. Additional comparisons are provided in Appendix F. Figure~\ref{fig:domain-net-lasttask} reports the per-task accuracy after the last training phase (domain in this setting). EFC++ remains competitive on the first domain compared to methods freezing the backbone, while showing better plasticity on later domains. The main exception is the \textit{painting} domain, where FeCAM performs better and FeTrIL is comparable, possibly because this domain is more strongly correlated with the first one. Compared with EFC, EFC++ achieves a better stability--plasticity trade-off, as EFC loses nearly 10 points on the first domain.

\begin{figure}[t]
    \centering
    \includegraphics[width=\columnwidth]{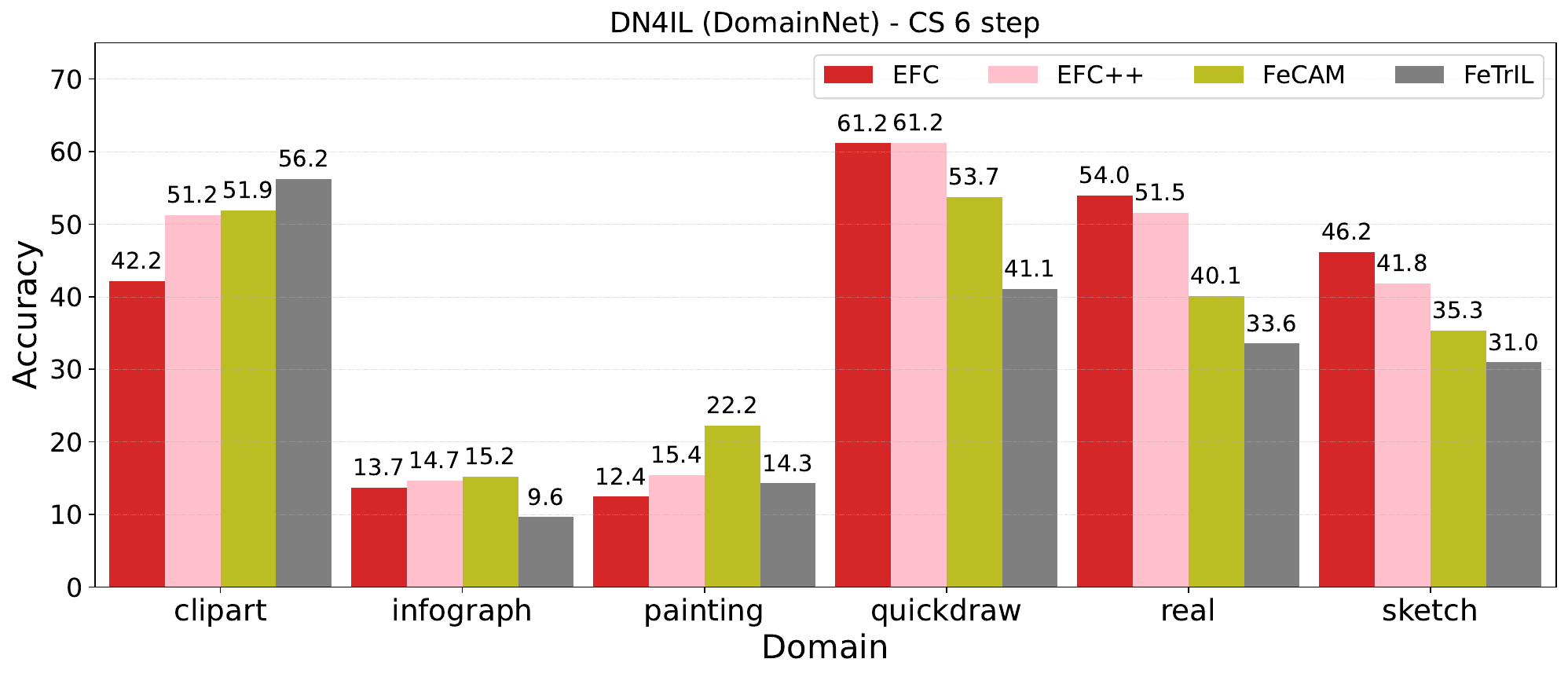}
    \caption{Accuracy on each domain after the final training step on DN4IL. Methods that freeze the backbone, such as FeCAM and FeTrIL, achieve higher accuracy on the first domain, while EFC and EFC++ excel at learning new domains.}
    \label{fig:domain-net-lasttask}
\end{figure}

\subsection{Comparison using the PyCIL framework with recent EFCIL methods}
\label{sec:pycil-comparison}

\begin{table}[t]
\centering
\small
\caption{Final per-step accuracy $A_{\text{step}}^K$ on CIFAR-100, Tiny-ImageNet, and ImageNet-100 in Class-IL Cold Start. Results are obtained using the DPCR codebase~\cite{he2025semantic}, based on the PyCIL framework~\cite{zhou2023pycil}.
}
\label{tab:af_comparison}
\resizebox{0.98\linewidth}{!}{%
\begin{tabular}{llcc}
\toprule
\textbf{Dataset} & \textbf{Method} & \textbf{10 Step} & \textbf{20 Step} \\
\midrule
\multirow{5}{*}{CIFAR-100}
& DS-AL (\hyperlink{cite.dsal}{2024})   & 36.83 & 28.90 \\
& ADC (\hyperlink{cite.resurrecting}{2024})        & 46.80 & 34.69 \\
& LDC (\hyperlink{cite.ldc}{2025})       & 46.60 & 36.76 \\
&  DPCR (\hyperlink{cite.he2025semantic}{2025})  &  \textbf{50.24} & \textbf{38.98} \\
\cmidrule(lr){2-4}
& EFC++                                  & \underline{49.47} & \underline{36.85} \\
\midrule

\multirow{5}{*}{Tiny-ImageNet}
& DS-AL (\hyperlink{cite.dsal}{2024})   & 27.01 & 21.86 \\
& ADC (\hyperlink{cite.resurrecting}{2024})      & 32.90 & 20.69 \\
& LDC (\hyperlink{cite.ldc}{2025})      & 33.74 & 24.49 \\
&  DPCR (\hyperlink{cite.he2025semantic}{2025})& \underline{35.20} &  \underline{26.54} \\
\cmidrule(lr){2-4}
& EFC++                                  & \textbf{38.66} & \textbf{32.65} \\
\midrule
\multirow{5}{*}{ImageNet-100}
& DS-AL (\hyperlink{cite.dsal}{2024})  & 45.55 & 34.10 \\
& ADC (\hyperlink{cite.resurrecting}{2024})      & 46.69 & 32.21 \\
& LDC (\hyperlink{cite.ldc}{2025})      & 49.98 &  34.87 \\
&  DPCR (\hyperlink{cite.he2025semantic}{2025})&  \underline{52.16} &  \underline{38.35} \\
\cmidrule(lr){2-4}
& EFC++                                  & \textbf{55.86} & \textbf{38.60} \\
\bottomrule
\end{tabular}%
}
\end{table}

To compare with very recent exemplar-free Class-IL methods, we additionally evaluate EFC++ within the PyCIL framework~\cite{zhou2023pycil,zhou2024class,zhou2024continual}. We use the DPCR codebase~\cite{he2025semantic} to ensure a fair comparison with recent methods, adopting the same data augmentations, first-task optimization protocol, class order, and random seed. We note that these results are \textit{not} directly comparable with those in Table~\ref{tab:sota}, since PyCIL uses different data augmentations and evaluates ImageNet-100, whose class subset differs from the ImageNet-Subset benchmark from PODNet~\cite{podnet} used in our main experiments.

Table~\ref{tab:af_comparison} compares EFC++ with DS-AL, ADC, LDC, and the recent DPCR method. On CIFAR-100, DPCR obtains the best results, while EFC++ remains competitive and ranks second in both the 10-step and 20-step settings. On Tiny-ImageNet, EFC++ improves over DPCR by 3.46 and 6.11 points in the 10-step and 20-step settings, respectively. On ImageNet-100, EFC++ obtains the best results, improving over DPCR by 3.70 and 0.25 points. These results show that EFC++ remains competitive with the latest methods on CIFAR-100 and achieves stronger performance on Tiny-ImageNet and ImageNet-100 under the same PyCIL protocol. 

A possible explanation for the comparatively smaller gains on CIFAR-100 is that its lower image resolution and relatively limited scale may make learning transferable feature representations from scratch in the Cold Start setting more difficult. Despite the strong data augmentations used in PyCIL, we observed signs of overfitting during training, consistent with the findings of Mittal et al.~\cite{Mittal_2021_CVPR}, who showed that overfitting during first-task training degrades representation quality and class-incremental learning performance on CIFAR-100. Consequently, EFM-based feature consolidation has less opportunity to exploit robust feature representations, leading to smaller improvements than on larger, more diverse datasets.

\section{Ablations and Additional analysis}
\label{sec:ablation-analysis}
In this section, we systematically ablate the key components of EFC++ to assess their individual contributions. We then analyze differences in feature representation drift between Warm Start and Cold Start scenarios, followed by a spectral analysis of the Empirical Feature Matrix and the impact of hyperparameter selection for EFM regularization. Finally, we discuss computational and storage requirements and examine the main limitations of EFC++, focusing on prototype rehearsal and covariance drift.

\subsection{Ablation Studies}
\label{sec:ablation}

\begin{figure}[t]
    \centering
        \includegraphics[width=\linewidth]{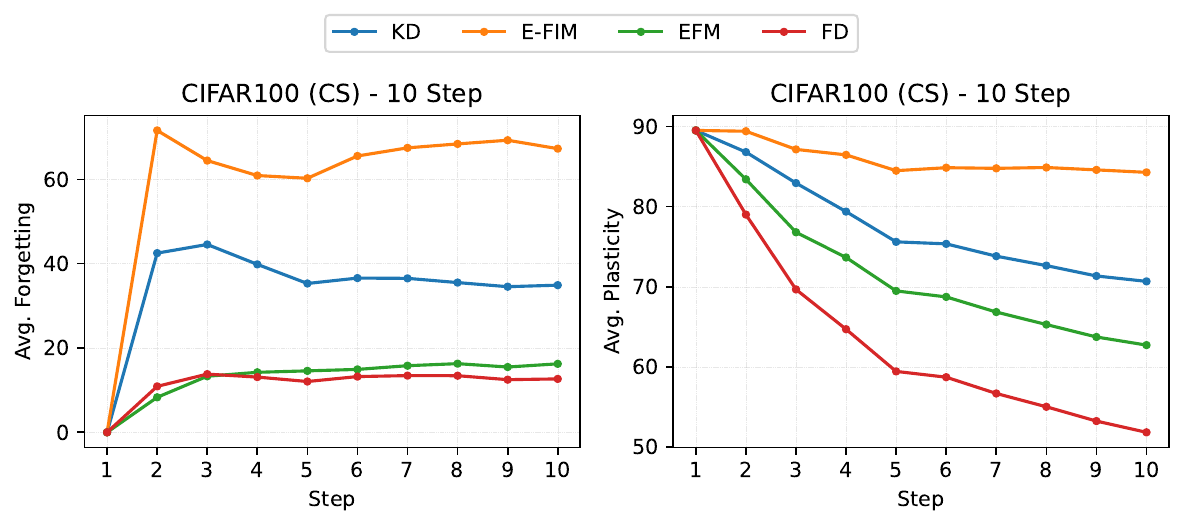}
        \caption{Comparison of regularization methods on CIFAR-100 (CS) 10-step. EFM balances stability and plasticity, reducing forgetting (left) and improving plasticity (right) over FD. KD and EWC exhibit strong initial plasticity but suffer from severe forgetting after the first task.}
        \label{fig:reg_alb_plot}
\end{figure}

\minisection{Ablation on EFM regularizer.}
We evaluate our proposed regularizer, EFM, against three established regularizer: Fisher Information Matrix (E-FIM, Eq.~\ref{eq:fisher}), Feature Distillation (FD, Eq.~\ref{eq:fd}), and Knowledge Distillation (KD)~\cite{lwf}. To carry out this comparison, we directly replaced EFM regularization in our training framework without modifying any other components. Specifically, we maintained the same prototype updates and prototype re-balancing procedures. In order to perform these experiments, we carefully validate the hyperparameters associated with each regularizer (see Appendix H for more details on the validation). For these experiments, in addition to the per-step accuracy, we report the average forgetting and the average plasticity (see Eq.~\ref{forg-plasticity}), to measure the stability and the plasticity induced by each regularizer.  As shown in Table~\ref{tab:reg_abl_result}, EFM achieves the best trade-off between stability and plasticity, leading to the highest final accuracy in both the 10-step and 20-step settings on CIFAR-100. Figure~\ref{fig:reg_alb_plot} further highlights EFM effectiveness: its average forgetting (left) remains comparable to FD during the incremental steps, while its plasticity (right) is significantly improved. In contrast, while KD and E-FIM  exhibit strong plasticity (right), they suffer from severe forgetting (left) after the first task, ultimately degrading their final performance as shown in Table~\ref{tab:reg_abl_result}.

\begin{table}[t]
    \centering
    \caption{Ablation study on regularizers. We report the final average forgetting ($F^K$), average plasticity ($\text{PL}^K$), and per-step accuracy ($A_{\text{step}}^K$) for CIFAR-100 at 10 and 20 steps.}
    \setlength{\tabcolsep}{5pt}
\resizebox{\linewidth}{!}{%
        \begin{tabular}{clcccccc}
        \toprule
        & & \multicolumn{3}{c}{\textbf{CS - 10 Step}} & \multicolumn{3}{c}{\textbf{CS - 20 Step}} \\

        & \textbf{Regularizer} &  $\mathbf{F^K}(\downarrow)$ & $\mathbf{\text{\textbf{PL}}^{K}}(\uparrow)$ & $\mathbf{A_{\text{step}}^K}(\uparrow)$ &  $\mathbf{F^K}(\downarrow)$ & $\mathbf{\text{\textbf{PL}}^{K}}(\uparrow)$ & $\mathbf{A_{\text{step}}^K}(\uparrow)$ \\

      \cmidrule(lr){2-2}
      \cmidrule(lr){3-5}
      \cmidrule(lr){6-8}

        \multirow{4}{*}{\rotatebox[origin=c]{90}{CIFAR-100}} 

      & E-FIM  & 67.28 & \textbf{84.27} & 23.72  & 71.98 & \textbf{83.22} & 14.83\\
      & KD  & 34.90 & \underline{70.67} & 39.26  & 41.11 & \underline{65.41} & 26.39 \\
      & FD  & \textbf{12.67} & 51.83 & \underline{40.44}  & \textbf{\phantom{0}8.78} & 35.85 & \underline{29.02} \\
      \cmidrule(lr){2-2}
      \cmidrule(lr){3-5}
      \cmidrule(lr){6-8}

      & EFM  & \underline{16.24} & 62.72 & \textbf{47.52}  & \underline{18.52} & 50.79 & \textbf{33.71}\\
\bottomrule
\end{tabular}} 
\label{tab:reg_abl_result}
\end{table}

\begin{table}[t]
\centering
\caption{Ablation study on prototype update. The performance of EFC++ improves with prototype updates in both Warm and Cold Start.}
\setlength{\tabcolsep}{4pt}
\resizebox{\linewidth}{!}{%
\begin{tabular}{l l c c c c c}

\toprule
 & \multirow{2}{*}{\textbf{Dataset}} & \multirow{2}{*}{\textbf{Step}} & \multicolumn{2}{c}{\textbf{Update Proto (WS)}} & \multicolumn{2}{c}{\textbf{Update Proto (CS)}} \\
                                   & & & \xmark & \cmark &  \xmark & \cmark   \\
\cmidrule(lr){2-3}
\cmidrule(lr){4-5}
\cmidrule(lr){6-7}

   & \multirow{2}{*}{CIFAR-100}        & 10 & $60.26 \pm 0.72$ & $\textbf{62.15} \pm 0.48$ & $46.30 \pm 1.45$ & $\textbf{47.52} \pm 0.68$ \\
                                   &                                    & 20 & $56.10 \pm 0.34$ & $\textbf{57.55} \pm 0.66$ & $31.71 \pm 2.26$ & $\textbf{33.71} \pm 1.41$ \\
\cmidrule(lr){2-3}
\cmidrule(lr){4-5}
\cmidrule(lr){6-7}
                                   & \multirow{2}{*}{Tiny-ImageNet}     & 10 & $50.57 \pm 0.40$ & $\textbf{51.67} \pm 0.31$ & $36.17 \pm 0.41$ & $\textbf{37.48} \pm 0.52$ \\
                                   &                                    & 20 & $46.86 \pm 1.19$ & $\textbf{50.41} \pm 0.51$ & $29.04 \pm 0.65$ & $\textbf{32.56} \pm 0.44$ \\
\cmidrule(lr){2-3}
\cmidrule(lr){4-5}
\cmidrule(lr){6-7}
                                   & \multirow{2}{*}{ImageNet-Subset}   & 10 & $69.09 \pm 0.30$ & $\textbf{69.28} \pm 0.46$ & $52.95 \pm 1.00$ & $\textbf{53.90} \pm 1.18$ \\
                                   &                                    & 20 & $\textbf{62.77} \pm 0.63$ & $62.75 \pm 0.46$ & $39.10 \pm 1.67$ & $\textbf{40.86} \pm 1.65$ \\
\bottomrule
\end{tabular}}
\label{tab:supp_ablation}
\end{table}

\begin{figure*}
    \centering
    \includegraphics[width=0.7\linewidth]{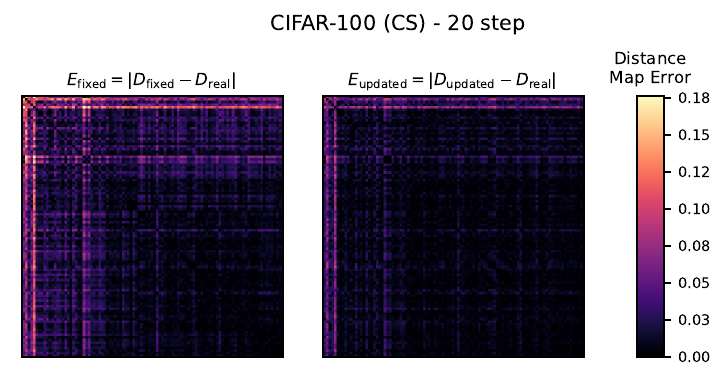}
\caption{Inter-class distance-map analysis on CIFAR-100 Cold Start with 20 steps.
We compare the absolute error between the cosine-distance maps computed from fixed prototypes ($D_{\text{fixed}}$) or EFM-updated prototypes ($D_{\text{updated}}$) and the map computed from real final class prototypes ($D_{\text{real}}$). EFM-updated prototypes better match the final inter-class geometry and reduce the average off-diagonal error, showing that the proposed update better preserves the feature-space structure across tasks.}
    \label{fig:inter_class_distance}
\end{figure*}

\minisection{Ablation on prototype update.}
In Table~\ref{tab:supp_ablation} we evaluate the effect of prototype updates using Eq.~\ref{eq:sdc_formula} and Eq.~\ref{our_weight}. Prototype updates significantly improve performance in both Warm Start and Cold Start across most benchmarks, except for ImageNet-Subset WS, where no significant improvement is observed.  

\subsection{Analysis of Feature Drift in Warm and Cold Start}
\label{sec:drift-ws-cs}
We further analyze how initial representation quality affects feature drift during incremental learning. Methods that freeze the backbone perform well in Warm Start but struggle in Cold Start, suggesting that strong initial features reduce the need for later plasticity. Consistently, EFC++ improves over EFC more clearly in Cold Start, where the induced drift of PR-ACE has a larger impact on final performance. This suggests that \textit{representations are less prone to change in Warm Start thanks to the stronger initialization of the feature extractor}.

\begin{figure}
    \centering    
    \includegraphics[width=\columnwidth]{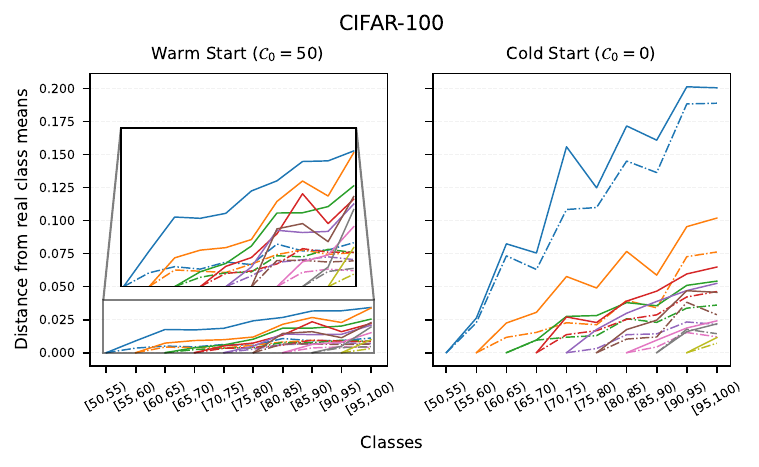}
    \caption{Distance between real class means and prototypes when kept fixed (solid) or updated via the Empirical Feature Matrix (dashed) in both Warm and Cold Start scenarios on CIFAR-100 in EFC++. In Cold Start, the average distance of the prototypes from real class means change more than in Warm Start. Our prototype update rule effectively mitigates this drift.}    
    \label{fig:drift_evaluation}
\end{figure}

To empirically evaluate this, we propose applying EFC++ in two settings. In a Warm Start setting, the model is pre-trained on half of the available classes (i.e., $\vert \mathcal{C}_0 \vert = 50$ for CIFAR-100) and then incrementally trained for 10 steps, each with 5 classes per step ($\vert \mathcal{C}_{i > 0} \vert = 5$). In a Cold Start setting, the model starts with a random initialization (i.e., $\vert \mathcal{C}_0 \vert = 0$) and is then incrementally trained for 10 steps by following the same order and the same number of classes per step as in the Warm Start setting.
Figure~\ref{fig:drift_evaluation} presents the results of this analysis. For each set of classes (on the $x$-axis), we plot two lines representing the average distance between the actual class means ($\mu$) and the estimated prototypes ($p$) throughout the experiment, with and without updates (dashed and solid lines, respectively) using the Empirical Feature Matrix as pseudo-metric. Comparing the two scenarios, we see that the average distance between the real class means and fixed prototypes is greater in the Cold Start setting, indicating that the representations are more prone to change compared to the Warm Start setting. Additionally, our prototype update rule (see Equations~\eqref{eq:sdc_formula} and~\eqref{our_weight}) effectively mitigates prototype drift in both scenarios. However, it is less effective in the more challenging Cold Start scenario due to the greater representation drift.

\minisection{Visual analysis of inter-class feature geometry.}
To further assess whether the EFM-based prototype update preserves the structure of the feature space, we analyze inter-class distance maps on CIFAR-100 in the Cold Start 20-step setting. Let $D_{\text{fixed}}$, $D_{\text{updated}}$, and $D_{\text{real}}$ denote the pairwise cosine-distance matrices computed from fixed prototypes stored when each class is first encountered, EFM-updated prototypes after the final training step, and real final class prototypes, respectively. The real final prototypes are obtained by forwarding the corresponding class data through the final backbone. We compare each prototype-based distance map with the real final one by computing
$E_{\text{fixed}} = |D_{\text{fixed}} - D_{\text{real}}|$ and
$E_{\text{updated}} = |D_{\text{updated}} - D_{\text{real}}|$.

Figure~\ref{fig:inter_class_distance} illustrates $E_{\text{fixed}}$ and $E_{\text{updated}}$. Fixed prototypes exhibit larger errors, indicating that the original inter-class geometry becomes misaligned as the feature extractor evolves. In contrast, EFM-updated prototypes produce a distance map closer to the real final one and reduce the average off-diagonal error. This confirms that the proposed update better preserves the relative geometry among classes across tasks.

\subsection{EFM Spectral Analysis and Hyperparameters}
\label{sec:eigenval-analysis}
Since EFC++ uses $E_t$ to selectively regularize drift in feature space, it is natural to question \emph{how} selective its regularization is. We can gain insight into this by analyzing how the rank of $E_t$ evolves with increasing incremental tasks.

Consider training a model incrementally on CIFAR-100 using EFC++ up to the tenth and final task, computing along the way the spectrum of $E_t$ for each $t \in [1, \dots , 10]$. We chose the ten task sequence for simplicity, but the same empirical conclusions hold regardless of the number of tasks. Recalling that $E_t \in \mathbb{R}^{n \times n}$ is symmetric, there then exist $U_t$ and $\Lambda_t \in \mathbb{R}^{n \times n}$ such that $E_t = U_t^{\top} \Lambda_t U_t$, 
where $U_t$ is an orthogonal matrix and $\Lambda_t=\text{diag}(\lambda_1, .., \lambda_n)$ is a diagonal matrix whose entries contain the eigenvalues of $E_t$, sorted in descending order. Since $E_t$ is positive semi-definite,  $\lambda_i \ge 0$ for each $i \in [1,\dots,n]$.

\begin{figure}
\begin{center}
\begin{tabular}{c}
\includegraphics[width=0.95\columnwidth]{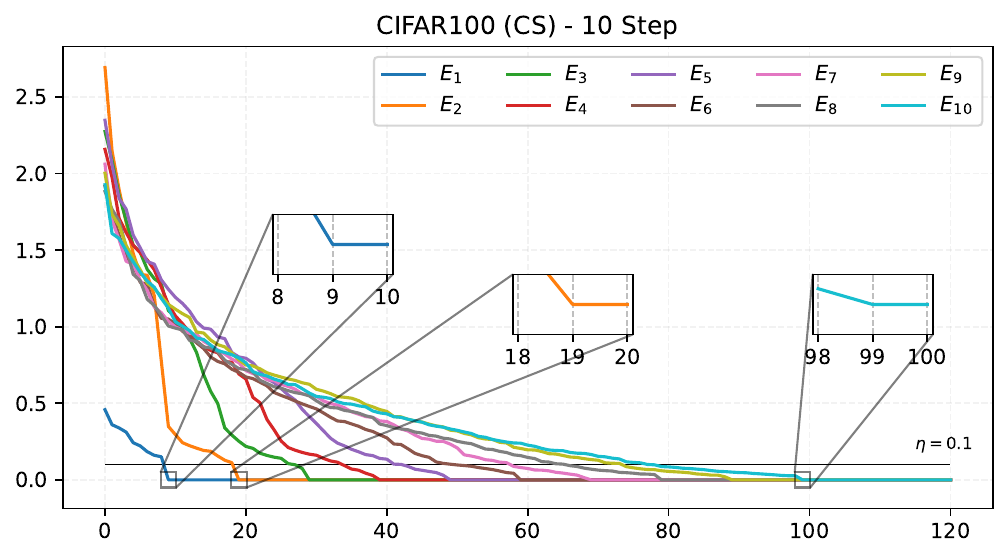} 
\end{tabular}
\end{center}
\caption{The spectrum of the Empirical Feature Matrix across incremental learning steps. For a better visualization of the spectrum in the analysis we considered a Cold Start 10-step scenario on CIFAR-100. The $x$-axis is truncated at the $120$th eigenvalue.}
\label{fig:rank_trace}
\end{figure}

In Figure~\ref{fig:rank_trace}, we plot the spectrum of $E_t$ at each $t \in [1,\dots, 10]$. This plot shows that the number of classes on which the matrix is estimated corresponds to an elbow in the curve beyond which the spectrum of the matrix vanishes. \textit{This implies that the rank of the matrices is exactly equal to the number of observed classes minus one.} 
 
\begin{figure*}[t]
    \centering
\includegraphics[width=\textwidth]{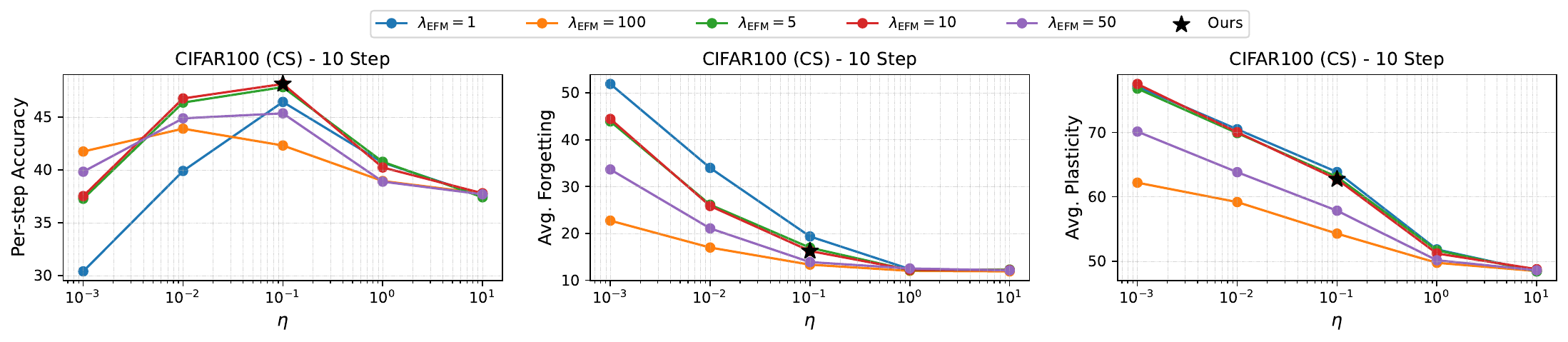} 
\caption{A grid search testing 25 combinations of $\lambda_{\text{EFM}}$, with the $x$-axis in log scale for better visualization. \textbf{Left}: The selected combination, represented by the star symbol, achieves the peak per-step accuracy. \textbf{Middle}: Reducing the value of $\eta$ weakens regularization and increases forgetting (see Eq. \ref{forg-plasticity}, left). \textbf{Right}: For high values of $\eta$, plasticity is significantly compromised (see Eq. \ref{forg-plasticity}, right).}
\label{fig:abl_eta_lambda}
\end{figure*}

Moreover, the horizontal line compares the magnitude of the eigenvalues across tasks with the damping term. Recalling that the results reported for EFC++ were obtained using parameter values of $\lambda_{\text{EFM}} = 10$ and $\eta = 0.1$, we observe that the plasticity constraints $\lambda_{\text{EFM}} \nu_i > \eta$ (described in Section~\ref{sec:empirical_feat}) are satisfied for nearly every $\nu_i > 0$, clearly demonstrating that \textit{our regularizer effectively constrains the features in a non-isotropic manner, distinguishing it from feature distillation}.

To provide empirical proof of the importance of calibrating such hyperparameters (i.e., avoiding too low a value of $\eta$ or degenerating into feature distillation), we also conducted a grid search testing 25 different combinations of $\lambda_{\text{EFM}}$ and $\eta$. Figure~\ref{fig:abl_eta_lambda} (left) shows that the selected combination achieves the peak value in terms of final per-step accuracy. In particular, it reveals that excessively low values of $\eta$ lead to weaker regularization and increased forgetting (middle), whereas excessively high values significantly reduce plasticity (right). See Appendix G for a more comprehensive analysis on other scenarios.

\subsection{Training Time and Memory Requirements}
\begin{figure}
\begin{center}
\begin{tabular}{c}
\includegraphics[width=0.95\columnwidth]{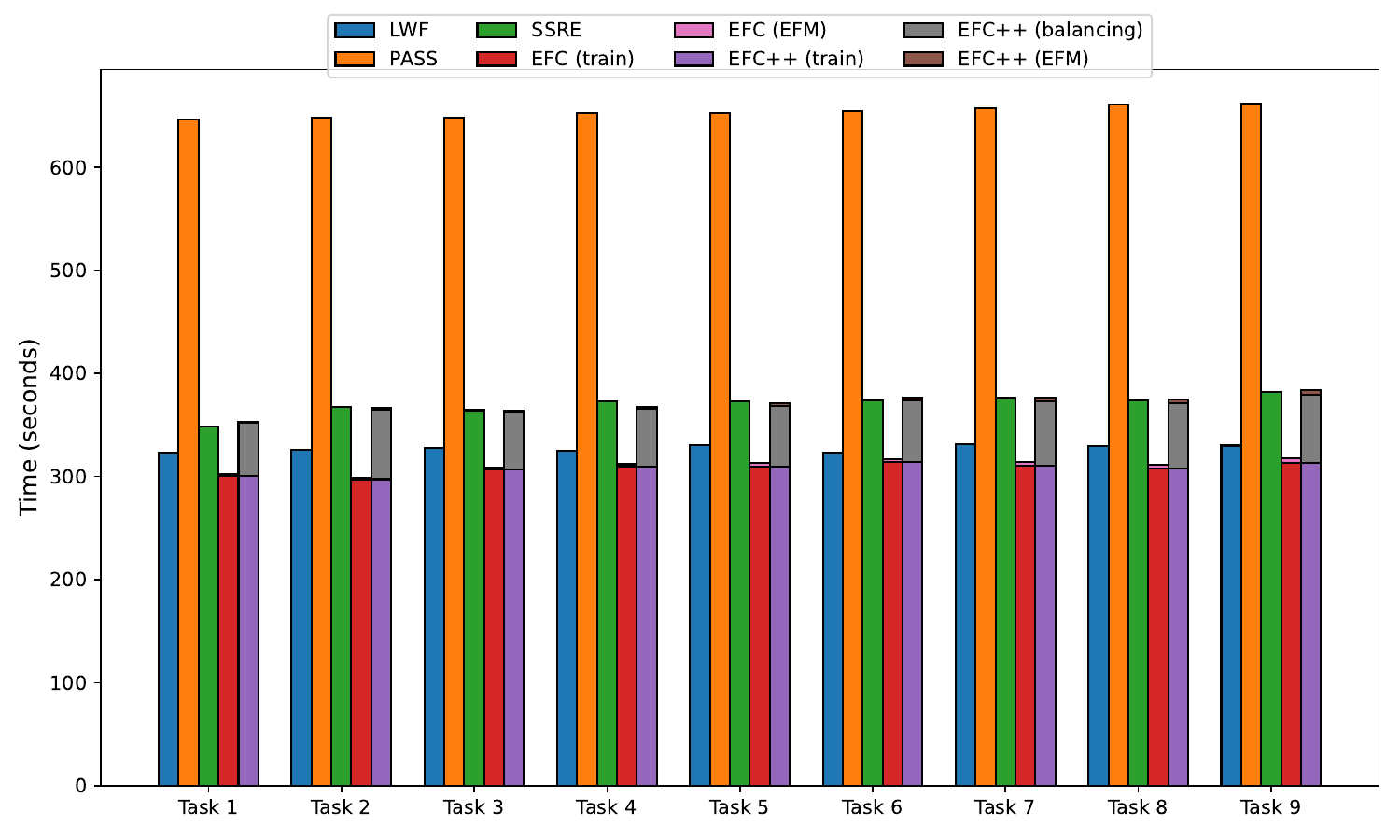} 
\end{tabular}
\end{center}
\caption{Training times. For EFC++ we report the time required for backbone training, EFM and prototype computation, and the prototype re-balancing phase. Overall, EFC++ exhibits a slight increase in computational time compared to EFC and LwF, shows comparable timings to SSRE, and remains more efficient than PASS, R-DFCIL, and ABD. Timings for R-DFCIL and ABD are omitted as the deep inversion model training time alone already significantly exceeds the time required for the other approaches (about 15 minutes on CIFAR-100).}
\label{fig:time}
\end{figure}

\minisection{Training Time.} FeTrIL and FeCAM are the fastest approaches as they freeze the backbone. Next is EWC, which does not apply any distillation and only E-FIM reguarization. LwF, R-DFCIL, ABD, EFC, and EFC++ achieve comparable training times for the feature extractor, with two forward passes required for computing distillation losses in LwF, ABD and R-DFCIL, and two forward passes for computing EFM regularization in EFC and EFC++. SSRE slightly increases the training time due to architectural modifications introduced by the method. PASS is the most computationally expensive approach, as it applies self-rotation to each task, effectively doubling the training time.

In post-training, both EFC++, R-DFCIL and ABD introduce an additional computational overhead. However, the computation of EFM and prototypes in EFC++ is negligible compared to the training time, as it only requires a single forward pass over the backbone using the current task's training dataset. Additionally, EFM can be computed without any backward pass (see Eq. \ref{eq:closed_form_efm}). The prototype re-balancing phase takes approximately one-fifth of the time required for training the backbone, since the only classifier is trained and the feature extractor is frozen. In this phase, a single forward pass over the current training dataset is required to compute the task features because the feature extractor is frozen. The training time slightly increases across tasks due to the growing number of sampled prototypes.

R-DFCIL and ABD further extend training time by requiring the training of a deep generative inversion network after each task, which takes approximately 15 minutes -- about three times longer than training the main model on CIFAR-100. This computation further increases when the resolution of the images to be generated increases (e.g. ImageNet). 

Figure \ref{fig:time} shows the complete training times, with ABD and R-DFCIL omitted due to their significantly larger scale. Overall, when considering both its training and prototype re-balancing phases, EFC++ exhibits slightly higher training times compared to LwF and EFC, comparable timing to SSRE, and significantly lower training time than PASS. Training times were measured using an NVIDIA RTX 3090 GPU.

\minisection{Memory requirements.}  
EWC and LwF do not use prototypes, so no additional storage is required. FeTrIL, PASS, and SSRE need to store only class means. R-DFCIL increases storage requirements by including a generative deep inversion model with about 33M parameters (roughly 125.89 MB at 32-bit precision). EFC, EFC++, and FeCaM must store both class means and covariances. Comparing their memory requirements to R-DFCIL’s, storing 100 class covariance matrices of size 512×512 (as in CIFAR-100 with a ResNet-18 backbone) requires about 100 MB -- slightly less than R-DFCIL. However, as the number of classes increases, the storage requirements for EFC, EFC++, and FeCaM grow with respect to R-DFCIL. In our previous work~\cite{magistri2024elastic}, we showed that these costs can be mitigated through covariance approximation, thus eliminating the need to store a covariance matrix per class—or by applying a low-rank approximation. This significantly reduces storage without significant performance degradation.

\subsection{Analysis of the limitations of EFC++}

\begin{table}[t]
\centering
\small
\caption{Impact of Mean and Covariance Drift on EFC++ (CIFAR-100 Cold Start).}
\label{tab:performance_summary}
\resizebox{0.9\linewidth}{!}{%
\begin{tabular}{lllcc}
\toprule
\textbf{Step} & \textbf{Mean} & \textbf{Covariance} & \textbf{Accuracy} & \textbf{Delta} \\
\midrule
\multirow{5}{*}{10} 
    & Fixed     & Fixed     & $46.30 \pm 1.45$ & — \\
   
    & \cellcolor{cyan!20}EFM   & \cellcolor{cyan!20}Fixed & \cellcolor{cyan!20}$47.52 \pm 0.68$ & \cellcolor{cyan!20} \quad \quad\,\,\,\phantom{0}+1.22 (\textbf{Ours}) \\
     & EFM   & Real      & $47.37 \pm 0.19$ & +1.07 \\
    & Real      & Fixed     & $49.22 \pm 0.17$ & +2.92 \\

    & Real      & Real      & $49.33 \pm 0.20$ & +3.03 \\

\cmidrule{1-5}
\multirow{5}{*}{20} 
    & Fixed     & Fixed     & $31.71 \pm 2.26$ & — \\  
    & \cellcolor{cyan!20}EFM   & \cellcolor{cyan!20}Fixed     & \cellcolor{cyan!20}$33.71 \pm 1.41$ &\cellcolor{cyan!20} \quad \quad\,\,\,\phantom{0}+2.00 (\textbf{Ours}) \\
       & EFM   & Real      & $33.86 \pm 1.49$ & +2.15 \\
    & Real      & Fixed     & $37.04 \pm 1.67$ & +5.33 \\
 
    & Real      & Real      & $38.59\pm 1.64$ & +6.88 \\
\cmidrule{1-5}
    & \multicolumn{2}{l}{Joint Training} &  71.00 &  \\
\bottomrule
\end{tabular}}
\end{table}

\begin{table}[t]
\centering
\small
\caption{Impact of Mean and Covariance Drift on EFC++ (ImageNet-1K Cold Start).}
\label{tab:performance_summary_imagenet}
\resizebox{0.9\linewidth}{!}{%
\begin{tabular}{lllcc}
\toprule
\textbf{Step} & \textbf{Mean} & \textbf{Covariance} & \textbf{Accuracy} & \textbf{Delta} \\
\midrule
\multirow{5}{*}{10} 
    & Fixed     & Fixed     & $43.65 \pm 0.04$  & — \\
   
    & \cellcolor{cyan!20}EFM   & \cellcolor{cyan!20}Fixed & \cellcolor{cyan!20}$44.72 \pm 0.11$  & \cellcolor{cyan!20}   \quad \quad\,\,\, +1.07 (\textbf{Ours}) \\
     & EFM   & Real      & $46.50 \pm 0.27$  & $+2.85$   \\
    & Real      & Fixed     & $49.95 \pm 0.09$ &  $+6.30$  \\

    & Real      & Real      & $49.97 \pm 0.09$  & $+6.32$  \\

\cmidrule{1-5}
\multirow{5}{*}{20} 
    & Fixed     & Fixed     & $34.00 \pm 0.03$ & — \\  
    & \cellcolor{cyan!20}EFM   & \cellcolor{cyan!20}Fixed     & \cellcolor{cyan!20}37.46 ± 0.03 &\cellcolor{cyan!20} \quad \quad\,\,\,  $+3.46$ (\textbf{Ours}) \\
       & EFM   & Real      &  $38.45 \pm 0.01$ &  $+4.45$\\
    & Real      & Fixed     &   $44.92 \pm 0.02$ & $+10.92$ \\
 
    & Real      & Real      & $45.55 \pm 0.06$  & $+11.55$  \\
\cmidrule{1-5}
    & \multicolumn{2}{l}{Joint Training} & 	
69.76  &  \\
\bottomrule
\end{tabular}}
\end{table}
 
Since EFC++ does not update class covariances across incremental learning steps, we aim to understand how \textit{covariance drift} affects the final performance. Tables \ref{tab:performance_summary} and \ref{tab:performance_summary_imagenet} compare different update strategies of class statistics: our approach (class means estimated via EFM and fixed covariances, highlighted in cyan) against an idealized setting using real mean and real covariances, computed from old task data fed in the current task backbone. While this strategy is infeasible in incremental learning -- since past data is unavailable -- it serves as an upper bound for performance and helps quantify the limitations of our prototype rehearsal strategy.

On CIFAR-100, using real class covariances provides limited benefits when the means are estimated with EFM. On ImageNet-1K, this effect is more visible, especially in the 10-step setting. Nevertheless, the dominant source of improvement is explained by the class means. Replacing our estimated class means with real ones, while keeping the covariances fixed, leads to the largest improvement: 2.92\% and 5.33\% on CIFAR-100, and 6.30\% and 10.92\% on ImageNet-1K, with respect to fixed statistics in the 10- and 20-step settings, respectively. This confirms the challenges highlighted in Section \ref{sec:drift-ws-cs} that our approach encounters when estimating the real class drift in Cold Start. Using real class covariances can further improve performance, particularly on ImageNet-1K, but this gain is much smaller than the impact of using real class means. In Appendix~I, we show that similar conclusions hold for Tiny-ImageNet and ImageNet-Subset.

From this analysis, we conclude that future work on Gaussian prototypes should prioritize more effective strategies for updating class means, beyond our current EFM approach. In this direction, hybrid strategies combining the EFM pseudo metric with recent prototype update strategies such as those in LDC~\cite{ldc}, represents a promising direction for future work. As a final note, we report the performance under Joint Training, where the gap remains substantial. Closing this gap will likely require either a more advanced strategy than Gaussian prototypes or improved regularization techniques beyond those we proposed.

\section{Conclusions}
\label{sec:conclusion_and_limitation}

Exemplar-free Class Incremental Learning requires a careful balance between model stability and plasticity. In this paper, we tackled the challenging Cold Start scenario in which inhibiting plasticity hinders the learning of novel tasks and caused a significant drop in incremental learning accuracy. We proposed Elastic Feature Consolidation++ (EFC++), improving over EFC and achieving competitive or superior performance with respect to recent state-of-the-art methods in Exemplar-free Class Incremental Learning (EFCIL).

We derived an Empirical Feature Matrix (EFM) that induces a pseudo-metric used to control feature drift and improves plasticity with respect to the usage of feature distillation. We showed that this matrix has interesting spectral properties depending on the number of classes in the incremental learning process. This matrix is also used to update prototypes, enabling them to adapt to the feature space drift, which is particularly pronounced in the Cold Start scenario. These prototypes are leveraged in a prototype re-balancing phase to learn the inter-task classifier.

Overall, EFC++ achieves superior stability-plasticity trade-off compared to EFC across small-scale, large-scale, and domain-incremental learning benchmarks. This is achieved by applying the prototype re-balancing phase and a regularization on the feature space. Future works could develop and analyze more plastic regularizers, similar to EFM, applied to feature maps in the inner activations of the network. The interaction between increased plasticity across different layers is still unclear.

Driven in part by the development of foundation models, recent works have begun investigating incremental learning scenarios in which the initial backbone is pre-trained on massive datasets~\cite{Wang_2025_ICCV,li2025addressing,Wang2026EKPC}. Recent evidence shows that the benefit of pre-training depends on how well the target incremental domain is represented in the pre-training data, and that target-domain training or adaptation can remain important under significant domain shift~\cite{Feillet_2025_WACV}. In the future, we aim to investigate how strategies that enhance plasticity, like EFC++, perform in such pre-trained scenarios. We are particularly interested in understanding whether enhancing plasticity can improve performance when there is a large domain shift from the pre-training task, which is very common in real applications.

\begin{acknowledgements}
 This work was supported by ERDF/EU and Grant AIA2025-163919-C52 funded by \\ MICIU/AEI/ 10.13039/501100011033, by the Italian national project MiSE AVIOS, by the European Union project \\ NextGenerationEU/PRTR, and by the Spanish Development Fund FEDER.
\end{acknowledgements}

\begin{dataavailability}
All data employed in this paper are publicly available. In our public repository 
(\url{https://github.com/simomagi/elastic_feature_consolidation}), we provide 
detailed instructions to download and use the data for the experiments.
\end{dataavailability}

\bibliographystyle{spmpsci}      
\bibliography{ref}   

@InProceedings{Mittal_2021_CVPR,
    author    = {Mittal, Sudhanshu and Galesso, Silvio and Brox, Thomas},
    title     = {Essentials for Class Incremental Learning},
    booktitle = {Proceedings of the IEEE/CVF Conference on Computer Vision and Pattern Recognition (CVPR) Workshops},
    month     = {June},
    year      = {2021},
    pages     = {3513-3522}
}

@InProceedings{privacy-inc,
author="Caselli, Lorenzo
and Magistri, Simone
and Bianconcini, Tommaso
and Benericetti, Andrea
and de Andrade, Douglas Coimbra
and Bagdanov, Andrew D.",
editor="Del Bue, Alessio
and Canton, Cristian
and Pont-Tuset, Jordi
and Tommasi, Tatiana",
title="Incremental and Decremental Continual Learning for Privacy-Preserving Video Recognition",
booktitle="Computer Vision -- ECCV 2024 Workshops",
year="2025",
publisher="Springer Nature Switzerland",
address="Cham",
pages="333--350",
isbn="978-3-031-92648-8"
}

@InProceedings{how-green,
author="Trinci, Tomaso
and Magistri, Simone
and Verdecchia, Roberto
and Bagdanov, Andrew D.",
editor="Del Bue, Alessio
and Canton, Cristian
and Pont-Tuset, Jordi
and Tommasi, Tatiana",
title="How Green Is Continual Learning, Really? Analyzing the Energy Consumption in Continual Training of Vision Foundation Models",
booktitle="Computer Vision -- ECCV 2024 Workshops",
year="2025",
publisher="Springer Nature Switzerland",
address="Cham",
pages="300--317",
isbn="978-3-031-92089-9"
}

@InProceedings{Feillet_2025_WACV,
    author    = {Feillet, Eva and Popescu, Adrian and Hudelot, C\'eline},
    title     = {A Reality Check on Pre-training for Exemplar-free Class-Incremental Learning},
    booktitle = {Proceedings of the Winter Conference on Applications of Computer Vision (WACV)},
    month     = {February},
    year      = {2025},
    pages     = {7614-7625}
}

@inproceedings{task-recency,
 author = {Rype\'{s}\'{c}, Grzegorz and Cygert, Sebastian and Trzci\'{n}ski, Tomasz and Twardowski, Bart\l omiej},
 booktitle = {Advances in Neural Information Processing Systems},
 doi = {10.52202/079017-2021},
 editor = {A. Globerson and L. Mackey and D. Belgrave and A. Fan and U. Paquet and J. Tomczak and C. Zhang},
 pages = {63268--63289},
 publisher = {Curran Associates, Inc.},
 title = {Task-recency bias strikes back: Adapting covariances in Exemplar-Free Class Incremental Learning},
 url = {https://proceedings.neurips.cc/paper_files/paper/2024/file/73ba81c7b25134a559c8a9c39ec1a4c3-Paper-Conference.pdf},
 volume = {37},
 year = {2024}
}

@InProceedings{ldc,
author="Gomez-Villa, Alex
and Goswami, Dipam
and Wang, Kai
and Bagdanov, Andrew D.
and Twardowski, Bartlomiej
and van de Weijer, Joost",
editor="Leonardis, Ale{\v{s}}
and Ricci, Elisa
and Roth, Stefan
and Russakovsky, Olga
and Sattler, Torsten
and Varol, G{\"u}l",
title="Exemplar-Free Continual Representation Learning via Learnable Drift Compensation",
booktitle="Computer Vision -- ECCV 2024",
year="2025",
publisher="Springer Nature Switzerland",
address="Cham",
pages="473--490",
isbn="978-3-031-72667-5"
}

@inproceedings{
he2025semantic,
title={Semantic Shift Estimation via Dual-Projection and Classifier Reconstruction for Exemplar-Free Class-Incremental Learning},
author={Run He and Di Fang and Yicheng Xu and Yawen Cui and Ming Li and Cen Chen and Ziqian Zeng and Huiping Zhuang},
booktitle={Forty-second International Conference on Machine Learning},
year={2025}
}

@INPROCEEDINGS{resurrecting,
author = {Goswami, Dipam and Soutif-Cormerais, Albin and Liu, Yuyang and Kamath, Sandesh and Twardowski, Bartlomiej and Van De Weijer, Joost },
booktitle = { 2024 IEEE/CVF Conference on Computer Vision and Pattern Recognition (CVPR) },
title = {{ Resurrecting Old Classes with New Data for Exemplar-Free Continual Learning }},
year = {2024},
volume = {},
ISSN = {},
pages = {28525-28534},
doi = {10.1109/CVPR52733.2024.02695},
url = {https://doi.ieeecomputersociety.org/10.1109/CVPR52733.2024.02695},
publisher = {IEEE Computer Society},
address = {Los Alamitos, CA, USA},
month =Jun}

@InProceedings{Liu_2026_CVPR,
    author    = {Liu, Xuan and Chang, Xiaobin},
    title     = {Elastic Weight Consolidation Done Right for Continual Learning},
    booktitle = {Proceedings of the Computer Vision and Pattern Recognition Conference (CVPR)},
    year      = {2026}
}

@ARTICLE{gido-tpami,
  author={Vogelstein, Joshua T. and Dey, Jayanta and Helm, Hayden S. and LeVine, Will and Mehta, Ronak D. and Tomita, Tyler M. and Xu, Haoyin and Geisa, Ali and Wang, Qingyang and van de Ven, Gido M. and Gao, Chenyu and Yang, Weiwei and Tower, Bryan and Larson, Jonathan and White, Christopher M. and Priebe, Carey E.},
  journal={IEEE Transactions on Pattern Analysis and Machine Intelligence}, 
  title={Simple Lifelong Learning Machines}, 
  year={2025},
  volume={47},
  number={11},
  pages={10033-10046},
  doi={10.1109/TPAMI.2025.3595364}}

@article{
hess2024knowledge,
title={Knowledge Accumulation in Continually Learned Representations and the Issue of Feature Forgetting},
author={Timm Hess and Eli Verwimp and Gido M van de Ven and Tinne Tuytelaars},
journal={Transactions on Machine Learning Research},
issn={2835-8856},
year={2024},
note={}
}

@inproceedings{
lanzillotta2026heads,
title={Heads collapse, features stay: Why Replay needs big buffers},
author={Giulia Lanzillotta and Damiano Meier and Thomas Hofmann},
booktitle={The Fourteenth International Conference on Learning Representations},
year={2026}
}

@InProceedings{Kang_2025_CVPR,
    author    = {Kang, Hankyul and Seifer, Gregor and Lee, Donghyun and Ryu, Jongbin},
    title     = {Do Your Best and Get Enough Rest for Continual Learning},
    booktitle = {Proceedings of the IEEE/CVF Conference on Computer Vision and Pattern Recognition (CVPR)},
    month     = {June},
    year      = {2025},
    pages     = {10077-10086}
}

@inproceedings{
deng2025unlocking,
title={Unlocking the Power of Rehearsal in Continual Learning: A Theoretical Perspective},
author={Junze Deng and Qinhang Wu and Peizhong Ju and Sen Lin and Yingbin Liang and Ness Shroff},
booktitle={Forty-second International Conference on Machine Learning},
year={2025}
}

@article{Jiang2026TeacherAgent,
author = {Jiang, Shengqin and Fang, Yaoyu and Zhang, Haokui and Liu, Qingshan and Qi, Yuankai and Yang, Yang and Wang, Peng}, 
title = {Teacher Agent: A Knowledge Distillation-Free Framework for Rehearsal-Based Video Incremental Learning},
journal = {International Journal of Computer Vision}, 
year = {2026},
volume = {134}, 
number = {4},
pages = {190},
doi = {10.1007/s11263-026-02779-2}, 
url = {}, issn = {1573-1405} }

@article{zhou2023pycil,
    author = {Da-Wei Zhou and Fu-Yun Wang and Han-Jia Ye and De-Chuan Zhan},
    title = {PyCIL: a Python toolbox for class-incremental learning},
    journal = {SCIENCE CHINA Information Sciences},
    year = {2023},
    volume = {66},
    number = {9},
    pages = {197101},
    doi = {https://doi.org/10.1007/s11432-022-3600-y}
  }

@article{zhou2024class,
    author = {Zhou, Da-Wei and Wang, Qi-Wei and Qi, Zhi-Hong and Ye, Han-Jia and Zhan, De-Chuan and Liu, Ziwei},
    title = {Class-Incremental Learning: A Survey},
    journal={IEEE Transactions on Pattern Analysis and Machine Intelligence},
    volume={46},
    number={12},
    pages={9851--9873},
    year = {2024}
}

@inproceedings{zhou2024continual,
    title={Continual learning with pre-trained models: A survey},
    author={Zhou, Da-Wei and Sun, Hai-Long and Ning, Jingyi and Ye, Han-Jia and Zhan, De-Chuan},
    booktitle={IJCAI},
    pages={8363-8371},
    year={2024}
}

@InProceedings{Wang_2025_ICCV,
    author    = {Wang, Yan and Zhou, Da-Wei and Ye, Han-Jia},
    title     = {Integrating Task-Specific and Universal Adapters for Pre-Trained Model-based Class-Incremental Learning},
    booktitle = {Proceedings of the IEEE/CVF International Conference on Computer Vision (ICCV)},
    month     = {October},
    year      = {2025},
    pages     = {806-816}
}

@inproceedings{
li2025addressing,
title={Addressing Imbalanced Domain-Incremental Learning through Dual-Balance Collaborative Experts},
author={Lan Li and Da-Wei Zhou and Han-Jia Ye and De-Chuan Zhan},
booktitle={Forty-second International Conference on Machine Learning},
year={2025}
}

@article{Wang2026EKPC,
  author  = {Wang, Huaijie and Cheng, De and He, Lingfeng and Li, Yan and Li, Jie and Wang, Nannan and Gao, Xinbo},
  title   = {{EKPC}: Elastic Knowledge Preservation and Compensation for Class-Incremental Learning},
  journal = {International Journal of Computer Vision},
  year    = {2026},
  volume  = {134},
  number  = {5},
  pages   = {238},
  doi     = {10.1007/s11263-026-02820-4},
  url     = {},
  issn    = {1573-1405}
}

@article{Nguyen2026COBRA,
  author  = {Nguyen, Xuan-Bac and Serna-Aguilera, Manuel and Choudhary, Arabinda Kumar and Sinha, Pawan and Li, Xin and Luu, Khoa},
  title   = {{COBRA}: A Continual Learning Approach to Vision-Brain Understanding},
  journal = {International Journal of Computer Vision},
  year    = {2026},
  volume  = {134},
  number  = {1},
  pages   = {30},
  doi     = {10.1007/s11263-025-02617-x},
  url     = {},
  issn    = {1573-1405}
}

@inproceedings{
zhuang2022acil,
title={{ACIL}: Analytic Class-Incremental Learning with Absolute Memorization and Privacy Protection},
author={HUIPING ZHUANG and Zhenyu Weng and Hongxin Wei and RENCHUNZI XIE and Kar-Ann Toh and Zhiping Lin},
booktitle={Advances in Neural Information Processing Systems},
editor={Alice H. Oh and Alekh Agarwal and Danielle Belgrave and Kyunghyun Cho},
year={2022},
url={https://openreview.net/forum?id=Vc4QUfqr4do}
}

@ARTICLE{exemplar-free-trafficlight,
  author={Liu, MeiChen and Ma, Jun and Kang, Qi and Zhou, MengChu and Albeshri, Aiiad},
  journal={IEEE Transactions on Vehicular Technology}, 
  title={Exemplar-Free Class Incremental Learning for Traffic Sign Classification}, 
  year={2026},
  volume={75},
  number={3},
  pages={3556-3568},
  doi={10.1109/TVT.2025.3606740}}

@inproceedings{
dosovitskiy2021an,
title={An Image is Worth 16x16 Words: Transformers for Image Recognition at Scale},
author={Alexey Dosovitskiy and Lucas Beyer and Alexander Kolesnikov and Dirk Weissenborn and Xiaohua Zhai and Thomas Unterthiner and Mostafa Dehghani and Matthias Minderer and Georg Heigold and Sylvain Gelly and Jakob Uszkoreit and Neil Houlsby},
booktitle={International Conference on Learning Representations},
year={2021},
url={}
}

@article{MAGISTRI202482,
title = {Continual learning for adaptive social network identification},
journal = {Pattern Recognition Letters},
volume = {180},
pages = {82-89},
year = {2024},
issn = {0167-8655},
doi = {https://doi.org/10.1016/j.patrec.2024.02.020},
url = {},
author = {Simone Magistri and Daniele Baracchi and Dasara Shullani and Andrew D. Bagdanov and Alessandro Piva}
}

@Article{Shaheen2022,
author={Shaheen, Khadija
and Hanif, Muhammad Abdullah
and Hasan, Osman
and Shafique, Muhammad},
title={Continual Learning for Real-World Autonomous Systems: Algorithms, Challenges and Frameworks},
journal={Journal of Intelligent {\&} Robotic Systems},
year={2022},
month={Apr},
day={23},
volume={105},
number={1},
pages={9},
issn={1573-0409},
doi={10.1007/s10846-022-01603-6},
url={}
}

@Article{Lee2020,
author={Lee, Cecilia S.
and Lee, Aaron Y.},
title={Clinical applications of continual learning machine learning},
journal={The Lancet Digital Health},
year={2020},
month={Jun},
day={01},
publisher={Elsevier},
volume={2},
number={6},
pages={e279-e281},
issn={2589-7500},
doi={10.1016/S2589-7500(20)30102-3},
url={}
}

@ARTICLE{9771396,
  author={Liu, Weiwei and Nie, Xiangli and Zhang, Bo and Sun, Xian},
  journal={IEEE Transactions on Geoscience and Remote Sensing}, 
  title={Incremental Learning With Open-Set Recognition for Remote Sensing Image Scene Classification}, 
  year={2022},
  volume={60},
  number={},
  pages={1-16},
  doi={10.1109/TGRS.2022.3173995}}

@Article{French1999,
author={French, Robert M.},
title={Catastrophic forgetting in connectionist networks},
journal={Trends in Cognitive Sciences},
year={1999},
month={Apr},
day={01},
publisher={Elsevier},
volume={3},
number={4},
pages={128-135},
issn={1364-6613},
doi={10.1016/S1364-6613(99)01294-2},
url={}
}

@ARTICLE{10.3389/fpsyg.2013.00504,

AUTHOR={Mermillod, Martial  and Bugaiska, Aurélia  and BONIN, Patrick },

TITLE={The stability-plasticity dilemma: investigating the continuum from catastrophic forgetting to age-limited learning effects},

JOURNAL={Frontiers in Psychology},

VOLUME={4},

YEAR={2013},

URL={},

DOI={10.3389/fpsyg.2013.00504},

ISSN={1664-1078},

}

@inproceedings{
goswami2023fecam,
title={Fe{CAM}: Exploiting the Heterogeneity of Class Distributions in Exemplar-Free Continual Learning},
author={Dipam Goswami and Yuyang Liu and Bart{\l}omiej Twardowski and Joost van de Weijer},
booktitle={Thirty-seventh Conference on Neural Information Processing Systems},
year={2023},
url={}
}

@ARTICLE{10444954,
  author={Wang, Liyuan and Zhang, Xingxing and Su, Hang and Zhu, Jun},
  journal={IEEE Transactions on Pattern Analysis and Machine Intelligence}, 
  title={A Comprehensive Survey of Continual Learning: Theory, Method and Application}, 
  year={2024},
  volume={},
  number={},
  pages={1-20},
  doi={10.1109/TPAMI.2024.3367329}}

@ARTICLE{9349197,
  author={De Lange, Matthias and Aljundi, Rahaf and Masana, Marc and Parisot, Sarah and Jia, Xu and Leonardis, Aleš and Slabaugh, Gregory and Tuytelaars, Tinne},
  journal={IEEE Transactions on Pattern Analysis and Machine Intelligence}, 
  title={A Continual Learning Survey: Defying Forgetting in Classification Tasks}, 
  year={2022},
  volume={44},
  number={7},
  pages={3366-3385},
  doi={10.1109/TPAMI.2021.3057446}}

@ARTICLE{9891836,
  author={Boschini, Matteo and Bonicelli, Lorenzo and Buzzega, Pietro and Porrello, Angelo and Calderara, Simone},
  journal={IEEE Transactions on Pattern Analysis and Machine Intelligence}, 
  title={Class-Incremental Continual Learning Into the eXtended DER-Verse}, 
  year={2023},
  volume={45},
  number={5},
  pages={5497-5512},
  doi={10.1109/TPAMI.2022.3206549}}

@article{
verwimp2024continual,
title={Continual Learning: Applications and the Road Forward},
author={Eli Verwimp and Rahaf Aljundi and Shai Ben-David and Matthias Bethge and Andrea Cossu and Alexander Gepperth and Tyler L. Hayes and Eyke H{\"u}llermeier and Christopher Kanan and Dhireesha Kudithipudi and Christoph H. Lampert and Martin Mundt and Razvan Pascanu and Adrian Popescu and Andreas S. Tolias and Joost van de Weijer and Bing Liu and Vincenzo Lomonaco and Tinne Tuytelaars and Gido M van de Ven},
journal={Transactions on Machine Learning Research},
issn={2835-8856},
year={2024},
url={},
note={}
}

@Article{vandeVen2022,
author={van de Ven, Gido M.
and Tuytelaars, Tinne
and Tolias, Andreas S.},
title={Three types of incremental learning},
journal={Nature Machine Intelligence},
year={2022},
month={Dec},
day={01},
volume={4},
number={12},
pages={1185-1197},
issn={2522-5839},
doi={10.1038/s42256-022-00568-3},
url={}
}

@InProceedings{Wu_2019_CVPR,
author = {Wu, Yue and Chen, Yinpeng and Wang, Lijuan and Ye, Yuancheng and Liu, Zicheng and Guo, Yandong and Fu, Yun},
title = {Large Scale Incremental Learning},
booktitle = {Proceedings of the IEEE/CVF Conference on Computer Vision and Pattern Recognition (CVPR)},
month = {June},
year = {2019}
}

@article{HE2026114901,
title = {REAL: Representation enhanced analytic learning for exemplar-free class-incremental learning},
journal = {Knowledge-Based Systems},
volume = {332},
pages = {114901},
year = {2026},
issn = {0950-7051},
doi = {https://doi.org/10.1016/j.knosys.2025.114901},
url = {},
author = {Run He and Di Fang and Yizhu Chen and Kai Tong and Cen Chen and Yi Wang and Lap-pui Chau and Huiping Zhuang} 
}

@article{10.1145/3785663,
author = {Qazi, Mohammad Areeb and Hashmi, Anees Ur Rehman and Sanjeev, Santosh and Almakky, Ibrahim and Saeed, Numan and Gonz\'{a}lez, Camila and Yaqub, Mohammad},
title = {Continual Learning in Medical Imaging: A Survey and Practical Analysis},
year = {2026},
issue_date = {June 2026},
publisher = {Association for Computing Machinery},
address = {New York, NY, USA},
volume = {58},
number = {8},
issn = {0360-0300},
url = {},
doi = {10.1145/3785663},
journal = {ACM Comput. Surv.},
month = feb,
articleno = {197},
numpages = {25}
}

@inproceedings{
magistri2024elastic,
title={Elastic Feature Consolidation For Cold Start Exemplar-Free Incremental Learning},
author={Simone Magistri and Tomaso Trinci and Albin Soutif and Joost van de Weijer and Andrew D. Bagdanov},
booktitle={The Twelfth International Conference on Learning Representations},
year={2024},
url={}
}

@inproceedings{davari2022probing,
 title      = {Probing Representation Forgetting in Supervised and Unsupervised Continual Learning},
 author     = {Davari, MohammadReza and Asadi, Nader and Mudur, Sudhir and Aljundi, Rahaf and Belilovsky, Eugene},
 booktitle  = {Proceedings of the IEEE/CVF Conference on Computer Vision and Pattern Recognition (CVPR 2022)},
 year       = {2022},
 month      = {June},
 }

@article{
  gowda2023dual,
  title={Dual Cognitive Architecture: Incorporating Biases and Multi-Memory Systems for Lifelong Learning},
  author={Shruthi Gowda and Bahram Zonooz and Elahe Arani},
  journal={Transactions on Machine Learning Research},
  issn={2835-8856},
  year={2023},
  url={}
}

@inproceedings{NIPS2017_f8752278,
 author = {Lopez-Paz, David and Ranzato, Marc\textquotesingle Aurelio},
 booktitle = {Advances in Neural Information Processing Systems},
 editor = {I. Guyon and U. Von Luxburg and S. Bengio and H. Wallach and R. Fergus and S. Vishwanathan and R. Garnett},
 pages = {},
 publisher = {Curran Associates, Inc.},
 title = {Gradient Episodic Memory for Continual Learning},
 url = {},
 volume = {30},
 year = {2017}
}

@inproceedings{NEURIPS2019_15825aee,
 author = {Aljundi, Rahaf and Belilovsky, Eugene and Tuytelaars, Tinne and Charlin, Laurent and Caccia, Massimo and Lin, Min and Page-Caccia, Lucas},
 booktitle = {Advances in Neural Information Processing Systems},
 editor = {H. Wallach and H. Larochelle and A. Beygelzimer and F. d\textquotesingle Alch\'{e}-Buc and E. Fox and R. Garnett},
 pages = {},
 publisher = {Curran Associates, Inc.},
 title = {Online Continual Learning with Maximal Interfered Retrieval},
 url = {},
 volume = {32},
 year = {2019}
}

@inproceedings{NEURIPS2022_5ebbbac6,
 author = {Zhang, Yaqian and Pfahringer, Bernhard and Frank, Eibe and Bifet, Albert and Lim, Nick Jin Sean and Jia, Yunzhe},
 booktitle = {Advances in Neural Information Processing Systems},
 pages = {14771--14783},
 title = {A simple but strong baseline for online continual learning: Repeated Augmented Rehearsal},
 url = {},
 volume = {35},
 year = {2022}
}

@InProceedings{Aljundi_2019_CVPR,
author = {Aljundi, Rahaf and Kelchtermans, Klaas and Tuytelaars, Tinne},
title = {Task-Free Continual Learning},
booktitle = {Proceedings of the IEEE/CVF Conference on Computer Vision and Pattern Recognition (CVPR)},
month = {June},
year = {2019}
}

@article{naturalgradient,
  title={New insights and perspectives on the natural gradient method},
  author={Martens, James},
  journal={The Journal of Machine Learning Research},
  volume={21},
  number={1},
  pages={5776--5851},
  year={2020},
  publisher={JMLRORG}
}

@article{datamatrix,
  title={Model-centric data manifold: the data through the eyes of the model},
  author={Grementieri, Luca and Fioresi, Rita},
  journal={SIAM Journal on Imaging Sciences},
  volume={15},
  number={3},
  pages={1140--1156},
  year={2022},
  publisher={SIAM}
}

@article{knowledge_dist,
  title={Distilling the knowledge in a neural network},
  author={Hinton, Geoffrey and Vinyals, Oriol and Dean, Jeff},
  journal={arXiv preprint arXiv:1503.02531},
  year={2015}
}

@article{ewc,
  title={Overcoming catastrophic forgetting in neural networks},
  author={Kirkpatrick, James and Pascanu, Razvan and Rabinowitz, Neil and Veness, Joel and Desjardins, Guillaume and Rusu, Andrei A and Milan, Kieran and Quan, John and Ramalho, Tiago and Grabska-Barwinska, Agnieszka and others},
  journal={Proceedings of the national academy of sciences},
  volume={114},
  number={13},
  pages={3521--3526},
  year={2017},
  publisher={National Acad Sciences}
}

@inproceedings{mas,
  title={Memory aware synapses: Learning what (not) to forget},
  author={Aljundi, Rahaf and Babiloni, Francesca and Elhoseiny, Mohamed and Rohrbach, Marcus and Tuytelaars, Tinne},
  booktitle={Proceedings of the European conference on computer vision (ECCV)},
  pages={139--154},
  year={2018}
}

@inproceedings{rwalk,
  title={Riemannian walk for incremental learning: Understanding forgetting and intransigence},
  author={Chaudhry, Arslan and Dokania, Puneet K and Ajanthan, Thalaiyasingam and Torr, Philip HS},
  booktitle={Proceedings of the European conference on computer vision (ECCV)},
  pages={532--547},
  year={2018}
}

@article{huszar,
  title={Note on the quadratic penalties in elastic weight consolidation},
  author={Husz{\'a}r, Ferenc},
  journal={Proceedings of the National Academy of Sciences},
  volume={115},
  number={11},
  pages={E2496--E2497},
  year={2018},
  publisher={National Acad Sciences}
}

@InProceedings{pmlr-v70-zenke17a,
  title = 	 {Continual Learning Through Synaptic Intelligence},
  author =       {Friedemann Zenke and Ben Poole and Surya Ganguli},
  booktitle = 	 {Proceedings of the 34th International Conference on Machine Learning},
  pages = 	 {3987--3995},
  year = 	 {2017},
  editor = 	 {Precup, Doina and Teh, Yee Whye},
  volume = 	 {70},
  series = 	 {Proceedings of Machine Learning Research},
  month = 	 {06--11 Aug},
  publisher =    {PMLR},
  url = 	 {}
}

@InProceedings{pmlr-v151-benzing22a,
  title = 	 { Unifying Importance Based Regularisation Methods for Continual Learning },
  author =       {Benzing, Frederik},
  booktitle = 	 {Proceedings of The 25th International Conference on Artificial Intelligence and Statistics},
  pages = 	 {2372--2396},
  year = 	 {2022},
  editor = 	 {Camps-Valls, Gustau and Ruiz, Francisco J. R. and Valera, Isabel},
  volume = 	 {151},
  series = 	 {Proceedings of Machine Learning Research},
  month = 	 {28--30 Mar},
  publisher =    {PMLR},
  url = 	 {}
}

@inproceedings{
lange2023continual,
title={Continual evaluation for lifelong learning: Identifying the stability gap},
author={Matthias De Lange and Gido M van de Ven and Tinne Tuytelaars},
booktitle={The Eleventh International Conference on Learning Representations },
year={2023},
url={}
}

@article{WANG2025107053,
title = {Enhancing consistency and mitigating bias: A data replay approach for incremental learning},
journal = {Neural Networks},
volume = {184},
pages = {107053},
year = {2025},
issn = {0893-6080},
doi = {https://doi.org/10.1016/j.neunet.2024.107053},
url = {},
author = {Chenyang Wang and Junjun Jiang and Xingyu Hu and Xianming Liu and Xiangyang Ji}
}

@InProceedings{Belouadah_2019_ICCV,
author = {Belouadah, Eden and Popescu, Adrian},
title = {IL2M: Class Incremental Learning With Dual Memory},
booktitle = {Proceedings of the IEEE/CVF International Conference on Computer Vision (ICCV)},
month = {October},
year = {2019}
}

@inproceedings{rotate_net,
  title={Rotate your networks: Better weight consolidation and less catastrophic forgetting},
  author={Liu, Xialei and Masana, Marc and Herranz, Luis and Van de Weijer, Joost and Lopez, Antonio M and Bagdanov, Andrew D},
  booktitle={2018 24th International Conference on Pattern Recognition (ICPR)},
  pages={2262--2268},
  year={2018},
  organization={IEEE}
}

@article{kronecker_fisher,
  title={Online structured laplace approximations for overcoming catastrophic forgetting},
  author={Ritter, Hippolyt and Botev, Aleksandar and Barber, David},
  journal={Advances in Neural Information Processing Systems},
  volume={31},
  year={2018}
}

@article{lwf,
  title={Learning without forgetting},
  author={Li, Zhizhong and Hoiem, Derek},
  journal={IEEE transactions on pattern analysis and machine intelligence},
  volume={40},
  number={12},
  pages={2935--2947},
  year={2017},
  publisher={IEEE}
}

@article{der++,
  title={Dark experience for general continual learning: a strong, simple baseline},
  author={Buzzega, Pietro and Boschini, Matteo and Porrello, Angelo and Abati, Davide and Calderara, Simone},
  journal={Advances in neural information processing systems},
  volume={33},
  pages={15920--15930},
  year={2020}
}

@inproceedings{icarl,
  title={icarl: Incremental classifier and representation learning},
  author={Rebuffi, Sylvestre-Alvise and Kolesnikov, Alexander and Sperl, Georg and Lampert, Christoph H},
  booktitle={Proceedings of the IEEE conference on Computer Vision and Pattern Recognition},
  pages={2001--2010},
  year={2017}
}

@inproceedings{sdc,
  title={Semantic drift compensation for class-incremental learning},
  author={Yu, Lu and Twardowski, Bartlomiej and Liu, Xialei and Herranz, Luis and Wang, Kai and Cheng, Yongmei and Jui, Shangling and Weijer, Joost van de},
  booktitle={Proceedings of the IEEE/CVF conference on computer vision and pattern recognition},
  pages={6982--6991},
  year={2020}
}

@inproceedings{ucir,
  title={Learning a unified classifier incrementally via rebalancing},
  author={Hou, Saihui and Pan, Xinyu and Loy, Chen Change and Wang, Zilei and Lin, Dahua},
  booktitle={Proceedings of the IEEE/CVF conference on computer vision and pattern recognition},
  pages={831--839},
  year={2019}
}

@inproceedings{pass,
  title={Prototype augmentation and self-supervision for incremental learning},
  author={Zhu, Fei and Zhang, Xu-Yao and Wang, Chuang and Yin, Fei and Liu, Cheng-Lin},
  booktitle={Proceedings of the IEEE/CVF Conference on Computer Vision and Pattern Recognition},
  pages={5871--5880},
  year={2021}
}

@inproceedings{evanescent,
  title={Bring evanescent representations to life in lifelong class incremental learning},
  author={Toldo, Marco and Ozay, Mete},
  booktitle={Proceedings of the IEEE/CVF Conference on Computer Vision and Pattern Recognition},
  pages={16732--16741},
  year={2022}
}

@inproceedings{
ayub2021eec,
title={{\{}EEC{\}}: Learning to Encode and Regenerate Images for Continual Learning},
author={Ali Ayub and Alan Wagner},
booktitle={International Conference on Learning Representations},
year={2021},
url={}
}

@inproceedings{NIPS2017_0efbe980,
 author = {Shin, Hanul and Lee, Jung Kwon and Kim, Jaehong and Kim, Jiwon},
 booktitle = {Advances in Neural Information Processing Systems},
 editor = {I. Guyon and U. Von Luxburg and S. Bengio and H. Wallach and R. Fergus and S. Vishwanathan and R. Garnett},
 pages = {},
 publisher = {Curran Associates, Inc.},
 title = {Continual Learning with Deep Generative Replay},
 url = {},
 volume = {30},
 year = {2017}
}

@InProceedings{Xiang_2019_ICCV,
author = {Xiang, Ye and Fu, Ying and Ji, Pan and Huang, Hua},
title = {Incremental Learning Using Conditional Adversarial Networks},
booktitle = {Proceedings of the IEEE/CVF International Conference on Computer Vision (ICCV)},
month = {October},
year = {2019}
}

@article{amari,
  title={Natural gradient works efficiently in learning},
  author={Amari, Shun-Ichi},
  journal={Neural computation},
  volume={10},
  number={2},
  pages={251--276},
  year={1998},
  publisher={MIT Press}
}

@article{facil,
  title={Class-incremental learning: survey and performance evaluation on image classification},
  author={Masana, Marc and Liu, Xialei and Twardowski, Bart{\l}omiej and Menta, Mikel and Bagdanov, Andrew D and Van De Weijer, Joost},
  journal={IEEE Transactions on Pattern Analysis and Machine Intelligence},
  volume={45},
  number={5},
  pages={5513--5533},
  year={2022},
  publisher={IEEE}
}

@article{CIFAR100,
  title={Learning multiple layers of features from tiny images},
  author={Krizhevsky, Alex and Hinton, Geoffrey and others},
  year={2009},
  publisher={Toronto, ON, Canada},
  journal={Technical report}
}

@article{TinyImageNet,
  title={Tiny imagenet challenge},
  author={Wu, Jiayu and Zhang, Qixiang and Xu, Guoxi},
  journal={Technical report},
  year={2017},
  publisher={Stanford University}
}

@inproceedings{ImageNet,
  title={Imagenet: A large-scale hierarchical image database},
  author={Deng, Jia and Dong, Wei and Socher, Richard and Li, Li-Jia and Li, Kai and Fei-Fei, Li},
  booktitle={2009 IEEE conference on computer vision and pattern recognition},
  pages={248--255},
  year={2009},
  organization={Ieee}
}

@inproceedings{generative_features,
  title={Generative feature replay for class-incremental learning},
  author={Liu, Xialei and Wu, Chenshen and Menta, Mikel and Herranz, Luis and Raducanu, Bogdan and Bagdanov, Andrew D and Jui, Shangling and de Weijer, Joost van},
  booktitle={Proceedings of the IEEE/CVF Conference on Computer Vision and Pattern Recognition Workshops},
  pages={226--227},
  year={2020}
}

@article{resnet,
  title={Deep Residual Learning for Image Recognition},
  author={Kaiming He and X. Zhang and Shaoqing Ren and Jian Sun},
  journal={2016 IEEE Conference on Computer Vision and Pattern Recognition (CVPR)},
  year={2015},
  pages={770-778}
}

@incollection{mccloskey1989catastrophic,
  title={Catastrophic interference in connectionist networks: The sequential learning problem},
  author={McCloskey, Michael and Cohen, Neal J},
  booktitle={Psychology of learning and motivation},
  volume={24},
  pages={109--165},
  year={1989},
  publisher={Elsevier}
}

@incollection{Martens2012TrainingDA,
  title={Training deep and recurrent networks with hessian-free optimization},
  author={Martens, James and Sutskever, Ilya},
  booktitle={Neural Networks: Tricks of the Trade: Second Edition},
  pages={479--535},
  year={2012},
  publisher={Springer}
}

@inproceedings{adaptive_feature_consolidation,
  title={Class-incremental learning by knowledge distillation with adaptive feature consolidation},
  author={Kang, Minsoo and Park, Jaeyoo and Han, Bohyung},
  booktitle={Proceedings of the IEEE/CVF conference on computer vision and pattern recognition},
  pages={16071--16080},
  year={2022}
}

@article{adam,
  title={Adam: A method for stochastic optimization},
  author={Kingma, Diederik P and Ba, Jimmy},
  journal={arXiv preprint arXiv:1412.6980},
  year={2014}
}

@inproceedings{podnet,
  title={Podnet: Pooled outputs distillation for small-tasks incremental learning},
  author={Douillard, Arthur and Cord, Matthieu and Ollion, Charles and Robert, Thomas and Valle, Eduardo},
  booktitle={Computer Vision--ECCV 2020: 16th European Conference, Glasgow, UK, August 23--28, 2020, Proceedings, Part XX 16},
  pages={86--102},
  year={2020},
  organization={Springer}
}

@inproceedings{fetril,
  title={Fetril: Feature translation for exemplar-free class-incremental learning},
  author={Petit, Gr{\'e}goire and Popescu, Adrian and Schindler, Hugo and Picard, David and Delezoide, Bertrand},
  booktitle={Proceedings of the IEEE/CVF Winter Conference on Applications of Computer Vision},
  pages={3911--3920},
  year={2023}
}

@inproceedings{deesil,
  title={DeeSIL: Deep-Shallow Incremental Learning.},
  author={Belouadah, Eden and Popescu, Adrian},
  booktitle={Proceedings of the European Conference on Computer Vision (ECCV) Workshops},
  year={2018}
}

@article{il2a,
  title={Class-incremental learning via dual augmentation},
  author={Zhu, Fei and Cheng, Zhen and Zhang, Xu-yao and Liu, Cheng-lin},
  journal={Advances in Neural Information Processing Systems},
  volume={34},
  pages={14306--14318},
  year={2021}
}

@inproceedings{ssre,
  title={Self-sustaining representation expansion for non-exemplar class-incremental learning},
  author={Zhu, Kai and Zhai, Wei and Cao, Yang and Luo, Jiebo and Zha, Zheng-Jun},
  booktitle={Proceedings of the IEEE/CVF Conference on Computer Vision and Pattern Recognition},
  pages={9296--9305},
  year={2022}
}

@inproceedings{aanet,
  title={Adaptive aggregation networks for class-incremental learning},
  author={Liu, Yaoyao and Schiele, Bernt and Sun, Qianru},
  booktitle={Proceedings of the IEEE/CVF conference on Computer Vision and Pattern Recognition},
  pages={2544--2553},
  year={2021}
}

@inproceedings{der,
  title={Der: Dynamically expandable representation for class incremental learning},
  author={Yan, Shipeng and Xie, Jiangwei and He, Xuming},
  booktitle={Proceedings of the IEEE/CVF Conference on Computer Vision and Pattern Recognition},
  pages={3014--3023},
  year={2021}
}

@inproceedings{memo,
  title={A Model or 603 Exemplars: Towards Memory-Efficient Class-Incremental Learning},
  author={Zhou, Da-Wei and Wang, Qi-Wei and Ye, Han-Jia and Zhan, De-Chuan},
  booktitle={The Eleventh International Conference on Learning Representations},
  year={2022}
}

@inproceedings{soutifcormerais2023comprehensive,
  title={A comprehensive empirical evaluation on online continual learning},
  author={Soutif-Cormerais, Albin and Carta, Antonio and Cossu, Andrea and Hurtado, Julio and Lomonaco, Vincenzo and Van de Weijer, Joost and Hemati, Hamed},
  booktitle={Proceedings of the IEEE/CVF International Conference on Computer Vision},
  pages={3518--3528},
  year={2023}
}

@article{pycl,
    author = {Zhou, Da-Wei and Wang, Qi-Wei and Qi, Zhi-Hong and Ye, Han-Jia and Zhan, De-Chuan and Liu, Ziwei},
    title = {Deep Class-Incremental Learning: A Survey},
    journal = {arXiv preprint arXiv:2302.03648},
    year = {2023}
 }

@inproceedings{endtoend,
  title={End-to-end incremental learning},
  author={Castro, Francisco M and Mar{\'\i}n-Jim{\'e}nez, Manuel J and Guil, Nicol{\'a}s and Schmid, Cordelia and Alahari, Karteek},
  booktitle={Proceedings of the European conference on computer vision (ECCV)},
  pages={233--248},
  year={2018}
}

@inproceedings{smith2021always,
  title={Always be dreaming: A new approach for data-free class-incremental learning},
  author={Smith, James and Hsu, Yen-Chang and Balloch, Jonathan and Shen, Yilin and Jin, Hongxia and Kira, Zsolt},
  booktitle={Proceedings of the IEEE/CVF International Conference on Computer Vision},
  pages={9374--9384},
  year={2021}
}

@article{lfl,
  title={Less-forgetting learning in deep neural networks},
  author={Jung, Heechul and Ju, Jeongwoo and Jung, Minju and Kim, Junmo},
  journal={arXiv preprint arXiv:1607.00122},
  year={2016}
}

@inproceedings{gao2022r,
  title={R-dfcil: Relation-guided representation learning for data-free class incremental learning},
  author={Gao, Qiankun and Zhao, Chen and Ghanem, Bernard and Zhang, Jian},
  booktitle={European Conference on Computer Vision},
  pages={423--439},
  year={2022},
  organization={Springer}
}


\newpage
\section*{Appendix}

\appendix

\section{Scope and Summary Notation}
In these appendices, we provide the analytical formulation of the Empirical Feature Matrix (Section \ref{app:efm}), the algorithm for Elastic Feature Consolidation (Section \ref{app:alg_efc}), dataset and optimization details (Section \ref{sup:opt_details}), additional insights into the experiments (Section \ref{app:exp1} and Section \ref{app:exp2}), extensive validation of our regularization hyperparameters (Section \ref{app:validation_hyper}), further ablation studies (Section \ref{app:ablations}),  an analysis on the limitations of prototype rehearsal in EFC++ (Section \ref{app:cov_drift}) and finally, per-step plots for Warm Start and Cold Start scenarios for the small-scale class-IL experiments (Section \ref{sup:per-step-plots}).

For convenience, we summarize here the notation used in the paper:
\begin{itemize}
\item $\mathcal{M}_t$: the incremental model at task $t$.
\item $K$: the total number of tasks.
\item $\mathcal{C}_t$: the set of classes associated with task $t$.
\item $\mathcal{D}$: the incremental dataset containing samples $\mathcal{X}_t$ and labels $\mathcal{Y}_t$ for each task $t$.
\item $f(\cdot; \theta_t)$: The feature vector computed by the backbone at task $t$ with parameters $\theta_t$.
\item $n$: the dimensionality of the feature space.
\item $W_t$: the classifier at task $t$ with dimensions $\mathbb{R}^{n \times \sum_{j=1}^t \vert \mathcal{C}_j \vert}$.
\item $E_t$: empirical feature matrix computed at task $t$.
\item $\lambda_{\text{EFM}}$: hyperparameter associated to the Empirical Feature Matrix, Eq. 10 in the main paper.
\item $\eta$: damping hyperparameter associated to the Identity Matrix, Eq. 10 in the main paper.
\item $\textbf{X}\sim \mathcal{X}_t$: a batch of current task data.
\item $P_{1:t-1}$: prototypes accumulated up to the task $t-1$.
\item $\tilde{\mathbf{P}} \sim P_{1:t-1}$: a batch of prototypes with Gaussian perturbation.
\item $A^K_{\text{step}}$ and $A^K_{\text{inc}}$: the per-step incremental accuracy and the  average incremental accuracy, Eq. 21 in the main paper.
\item $F^K$ and $\text{PL}^K$: the average forgetting and the cumulative average accuracy on the last task, Eq. 23 in the main paper.

\end{itemize}

\section{The Empirical Feature Matrix}
\label{app:efm}
In this appendix we derive an analytic formulation of the local feature matrix defined in the main paper in Section 3.3. To simplify the notation, we recall its definition without explicitly specifying the task, as it is unnecessary for the derivation:

\begin{equation}
\label{eq:supp_localfm}
   E_{f(x)} = \underset{y\sim p(y)}{\mathbb{E}}\biggl[ \biggl( \frac{ \partial \log p(y) } {\partial f(x)} \biggl)  \biggl( \frac{\partial \log p(y)} {\partial f(x)} \biggl) ^\top \biggl].
\end{equation}
We begin by computing the Jacobian matrix with respect to the feature space of the output function $g:\mathbb{R}^m \rightarrow \mathbb{R}^m$ which is the log-likelihood of the model on logits $z$: $$g(z)=\log(\text{softmax}(z)) = [\log(\sigma_1(z)), \dots, \log(\sigma_m(z))]^{\top},$$
where
 $$\sigma_i(z) = \frac{e^{z_i}}{\sum_{j=1}^m e^{z_j}}.$$
The partial derivatives of $g$ with respect to each input $z_i$ are:
\begin{equation}
 \frac{\partial \log (\sigma_i(z)) } {\partial z_j}=
     \begin{cases}
         1-\sigma_i(z) & \text{if } i = j,\\
         \phantom{1}-\sigma_j(z) & \text{otherwise}. 
     \end{cases}
\end{equation}
Thus, the Jacobian matrix of $g$ is:
\begin{equation}
\begin{aligned}
     J(z) &= 
         \begin{bmatrix} 
             1-\sigma_1(z) & -\sigma_2(z) & \dots & -\sigma_m(z)\\
             -\sigma_1(z) & 1-\sigma_2(z)&  & \vdots \\
              \vdots &  & \ddots & \\
             -\sigma_1(z) &   \dots  &  & 1-\sigma_m(z)
         \end{bmatrix} \\
         \label{eqn:simple-jacobian}
     &= I_m - \mathds{1}_m \cdot \begin{bmatrix}
         \sigma_1(z) \\  \vdots \\ \sigma_m(z) \\
     \end{bmatrix}^{\top}  = I_m - P,
\end{aligned}
\end{equation}
where $I_m \in \mathbb{R}^{m \times m}$ is the identity matrix and $\mathds{1}_m$ represent the vector with all entries equal to $1$.

Recalling Eq.~\ref{eq:supp_localfm}, we are interested in computing the derivatives of the log-probability vector with respect to the feature vector:
\begin{equation}
\begin{aligned}
 \frac{ \partial \log p(y_1,\dots,y_m|f; W) } {\partial f} &= \frac{\partial(Wf)}{\partial f}  \frac{\partial(\log(\sigma(z)))} {\partial z} \\
 &= W J(z),
\end{aligned}
\end{equation}
where $z = Wf$ for $W$ the classifier weight matrix mapping features $f$ to logits $z$. Combining this with Eq.~\ref{eqn:simple-jacobian} we have:
\begin{align}
 \label{eqn:supp_fast-EFM}
   E_{f(x)} =  \underset{y\sim p(y)}{\mathbb{E}}\biggl[ W(I_m-P)_y (W(I_m-P)_y)^{\top} \biggl],
\end{align}
where $p(y)=p(y|f(x); W)$, $P$ is the matrix containing the probability vector associated with $f$ in each row, and $(I_m-P)_y$ is the column vector containing the $y_{th}$ row of the Jacobian matrix. Computing $E_t$ using Eq.~\ref{eqn:supp_fast-EFM} requires only a single forward pass of all data through the network, whereas a naive implementation requires an additional backward pass up to the feature embedding layer.

\section{Elastic Feature Consolidation algorithm}
\label{app:alg_efc}
For completeness we show in Algorithm~\ref{alg:efc_app} the pseudocode for Elastic Feature Consolidation, which uses the asymmetric cross-entropy loss instead of the prototype rebalancing phase as in EFC++. All the equations refer to the main paper.

 \label{alg:efc-pseudocode}
 \RestyleAlgo{ruled}
 \DontPrintSemicolon
 \begin{algorithm}
 \SetNoFillComment
 \label{alg:efc_app}
 \caption{Elastic Feature Consolidation}

        \KwData{ $\mathcal{M}_1$, $E_1$, $P_1$}
 \For{$t=2, \dots T$}{
   \tcc{Backbone and Classifier Train}
\For{\text{each optimization step}}{
        sample $\mathbf{X} \sim \mathcal{X}_t$, $\tilde{\mathbf{P}} \sim P_{1:t-1}$ \;
        
        compute  $\mathcal{L}^{\text{EFM}}_t$ \,\, (Eq. 10) 
        
         compute  $\mathcal{L}^{\text{PR-ACE}}_t$ \,\, (Eq. 13)

         $\mathcal{L}^{\text{EFC}}_t \gets \mathcal{L}^{\text{PR-ACE}}_t + \mathcal{L}^{\text{EFM}}_t$ \; 

         Update  $\theta_t \gets \theta_t - \alpha\frac{\partial \mathcal{L}_t^{\text{EFC}}}{\partial \theta_t}$

          Update  $W_t \gets W_t - \alpha\frac{\partial \mathcal{L}_t^{\text{EFC}}}{\partial W_t}$
     }
     \vspace{0.5em}
\tcc{Prototype Update}
     $ P_{1:t\text{-}1}  \gets P_{1:t\text{-}1} + \Delta(E_{t\text{-}1})$ \,\, (Eqs. 15, 16) \;
     
     $ P_{1:t}   \gets P_{1:t\text{-}1} \cup P_t$ \;
       \vspace{0.5em}
     \tcc{EFM Calculus}
     $E_t \gets \text{EFM}(\mathcal{X}_t,   \mathcal{M}_t )$ \,\, (Eq. 7) \;
 }
 \end{algorithm}

\section{Dataset, Implementation and Hyperparameter Settings}
\label{sup:opt_details}
In this section we expand the discussion of Section 5 of the main paper by providing additional information about the dataset and the implementation of the approaches we tested.

\subsection{Datasets}
\label{app:dataset}
We perform our experimental evaluation on four standard datasets. CIFAR-100~\cite{CIFAR100} consists of 60,000 images divided into 100 classes, with 600 images per class (500 for training and 100 for testing). ImageNet-1K is the original ImageNet dataset~\cite{ImageNet}, consisting of about 1.3 million images divided into 1,000 classes. Tiny-ImageNet~\cite{TinyImageNet} consists of 100,000 images divided into 200 classes, which are taken from ImageNet and resized to $64 \times 64$ pixels. ImageNet-Subset is a subset of the original ImageNet dataset that consists of 100 classes. The images are resized to $224 \times 224$ pixels. DN4IL~\cite{gowda2023dual} is a balanced subset of DomainNet and it consists of six domains, each with 100 classes, and a total of 85,000 images. The images are resized to $224 \times 224$ pixels. Each domain (i.e. \textit{clipart}, \textit{infograph}, \textit{painting}, \textit{quickdraw}, \textit{real}, and \textit{sketch}) is treated as a different task, resulting in a 6-step scenario used to evaluate the performance when large domain shifts occur.

\subsection{Implementation and Hyperparameters}
\label{supp:hyperparameters}
For all the methods, we use the standard ResNet-18~\cite{resnet} backbone trained from scratch. 

\minisection{First Task Optimization Details}. We use the same optimization settings for both the Warm Start and Cold Start scenarios.  We train the models on CIFAR-100 and on Tiny-ImageNet for 100 epochs with Adam~\cite{adam} using an initial learning rate of $1e^{\text{-}3}$ and fixed weight decay of $2e^{\text{-}4}$. The learning rate is reduced by a factor of 0.1 after 45 and 90 epochs (as done in \cite{pass,evanescent}). For ImageNet-Subset, we followed the implementation of PASS~\cite{pass}, fixing the number of epochs at 160, and used Stochastic Gradient Descent with an initial learning rate of 0.1, momentum of 0.9, and weight decay of $5e^{\text{-}4}$. The learning rate was reduced by a factor of 0.1 after 80, 120, and 150 epochs. We applied the same label and data augmentation (random crops and flips) for all the evaluated datasets. For the first task of each dataset, we use self-rotation as performed by~\cite{pass,evanescent}. For both ImageNet-1k and DN4IL, we apply the same hyperparameters and training protocol used for the ImageNet-Subset. The batch size is set to 64 for DN4IL and 256 for ImageNet-1k. For ImageNet-1k, self-rotation is deactivated for the first task.

\minisection{Incremental Steps}. Below we provide the hyperparameters and the optimization settings we used for the incremental steps of each state-of-the-art method we tested.
\begin{itemize}
    \item \textbf{EWC~\cite{ewc}}: We used the implementation of~\cite{facil}. Specifically, we configured the coefficient associated to the regularizer as $\lambda_{\text{E-FIM}} = 5000$ and the fusion of the old and new importance weights is done with $\alpha = 0.5$. For the incremental steps we fix the total number of epochs to $100$ and we use Adam optimizer with an initial learning rate of  $1e^{\text{-}3}$  and fixed weight decay of $2e^{\text{-}4}$. The learning rate is reduced by a factor of 0.1 after 45 and 90 epochs.
                            
    \item \textbf{LwF~\cite{lwf}}: We used the implementation of~\cite{facil}. In particular, we set the temperature parameter $T=2$ as proposed in the original work and the parameter associated to the regularizer $\lambda_{\text{LwF}}$ to $10$. For the incremental steps we fix the total number of epochs to $100$ and we use Adam optimizer with an initial learning rate of  $1e^{\text{-}3}$  and fixed weight decay of $2e^{\text{-}4}$. The learning rate is reduced by a factor of 0.1 after 45 and 90 epochs.
    
    \item \textbf{PASS~\cite{pass}}: We follow the implementation provided by the authors. It is an approach relying upon feature distillation and prototypes generation. Following the original paper we set $\lambda_{\text{FD}} = 10$ and $\lambda_{\text{pr}} = 10$.  In the original code, we found a temperature parameter, denoted as $T$, applied to the classification loss, which we set to $T=1$. As provided in the original paper, for the incremental steps we fix the total number of epochs to 100 and we use Adam optimizer with an initial learning rate of  $1e^{\text{-}3}$  and fixed weight decay of $2e^{\text{-}4}$. The learning rate is reduced by a factor of 0.1 after 45 and 90 epochs
    
    \item \textbf{SSRE~\cite{ssre}}: We follow the implementation provided by the authors. It is an approach relying upon feature distillation and prototypes generation. Following the original paper, we set $\lambda_{\text{FD}} = 10$ and $\lambda_{\text{pr}} = 10$. In the original code, we find a temperature parameter, denoted as $T$, applied to the classification loss, which we set to $T=1$. Following the code, for the incremental steps we fixed the total number of epochs to $60$ and used Adam with an initial learning rate of $2e^{\text{-}4}$ and fixed weight decay of  $5e^{\text{-}4}$. The learning rate is reduced by a factor of 0.1 after 45 epochs.
    
    \item \textbf{FeTrIL~\cite{fetril}}: We follow the official code provided by the authors. During the incremental steps, it uses a Linear SVM classifier working on the pseudo-features extracted from the frozen backbone after the first task. We set the SVM regularization $C=1$ and the tolerance to $0.0001$ as provided by the authors.

    \item \textbf{FeCAM~\cite{goswami2023fecam}}: We follow the implementation provided by the authors. During the steps, it uses Mahalanobis-based distance for classification on the features extracted from the frozen backbone after the first task and transformed via Tukey’s Ladder of Powers transformation. To improve and stabilize the Mahalanobis-based distance classification, it applies covariance normalization and covariance shrinkage. Following the original implementation, we set the hyperparameter of Tukey’s Ladder of Powers transformation, $\lambda$, to $0.5$ and the hyperparameters of the covariance shrinkage, $\gamma_1$ and $\gamma_2$, to $1$.

    \item \textbf{Deep Inversion Methods: }We trained \textbf{ABD}~\cite{smith2021always} and \textbf{R-DFCIL}~\cite{gao2022r} following their official implementations. The only difference from the original papers is that CIFAR-100 and Tiny-ImageNet were tested using a ResNet-32~\cite{resnet} backbone. However, for a fair comparison, we replaced it with ResNet-18. Regarding optimization details, we follow the same settings originally used by the authors for ImageNet-Subset, where they employed ResNet-18. Specifically, we train all benchmarks for 100 epochs for ABD and 120 epochs for R-DFCIL, using SGD with an initial learning rate of $0.1$ for CIFAR-100, Tiny-ImageNet, and DN4IL, and $0.005$ for ImageNet-Subset and ImageNet-1K, following the original learning rate schedule\footnote{We chose not to test ABD on DN4IL and ImageNet-1K because it generally performs worse than R-DFCIL on all other datasets.}. We apply a fixed weight decay of $1e^{\text{-}4}$ and a momentum of $0.9$. The batch size is set to $128$ for CIFAR-100 and Tiny-ImageNet, while for DN4IL, ImageNet-Subset, and ImageNet-1K, it is $64$. We use the same deep inversion network, running 10,000 inversion iterations for CIFAR-100 and Tiny-ImageNet, and 50,000 iterations for ImageNet-Subset and ImageNet-1K.  
\end{itemize} 

\section{Small-scale Class-IL Experiments}
\label{app:exp1}
\begin{figure}[t]
    \centering
    \includegraphics[width=\columnwidth]{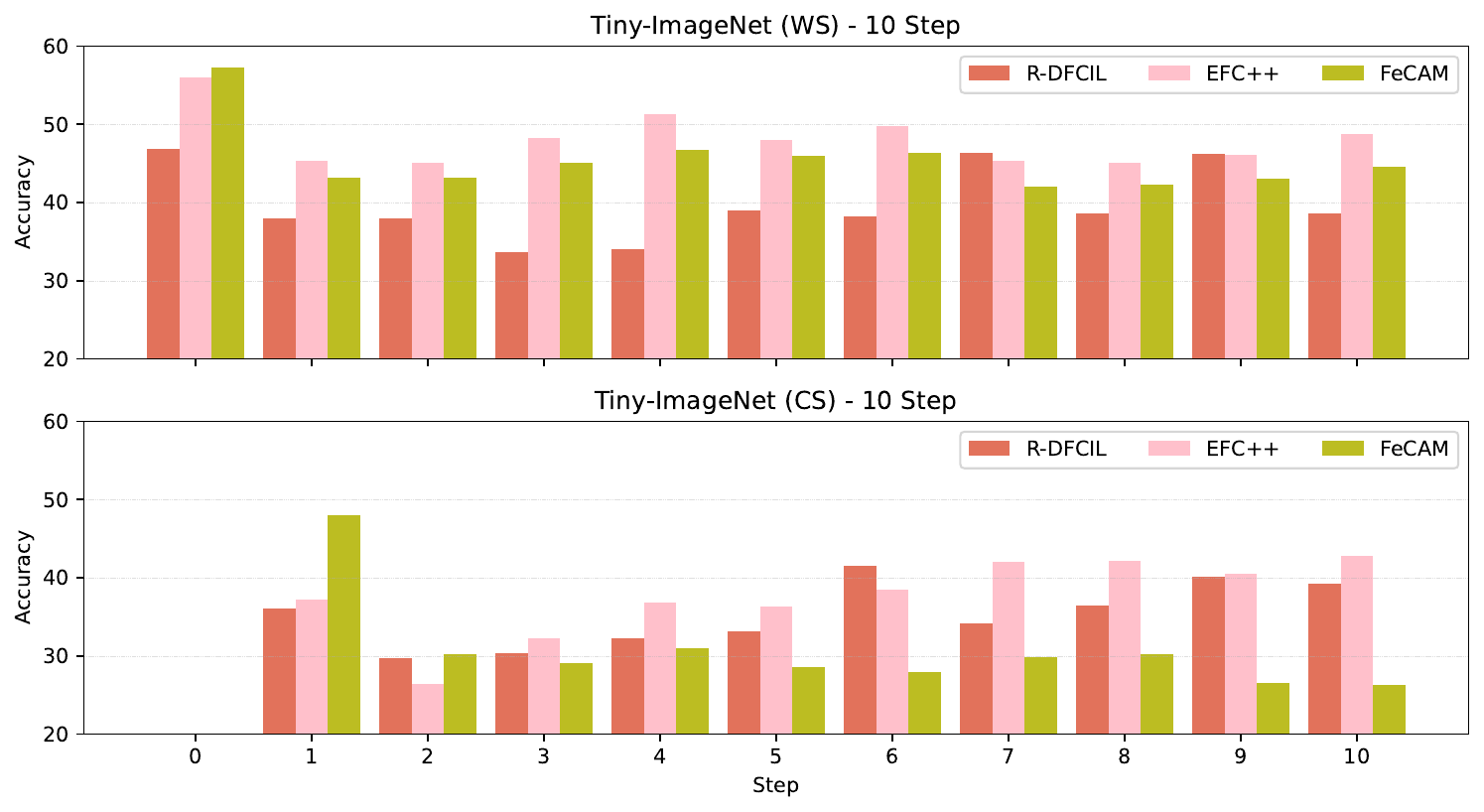}
    \includegraphics[width=\columnwidth]{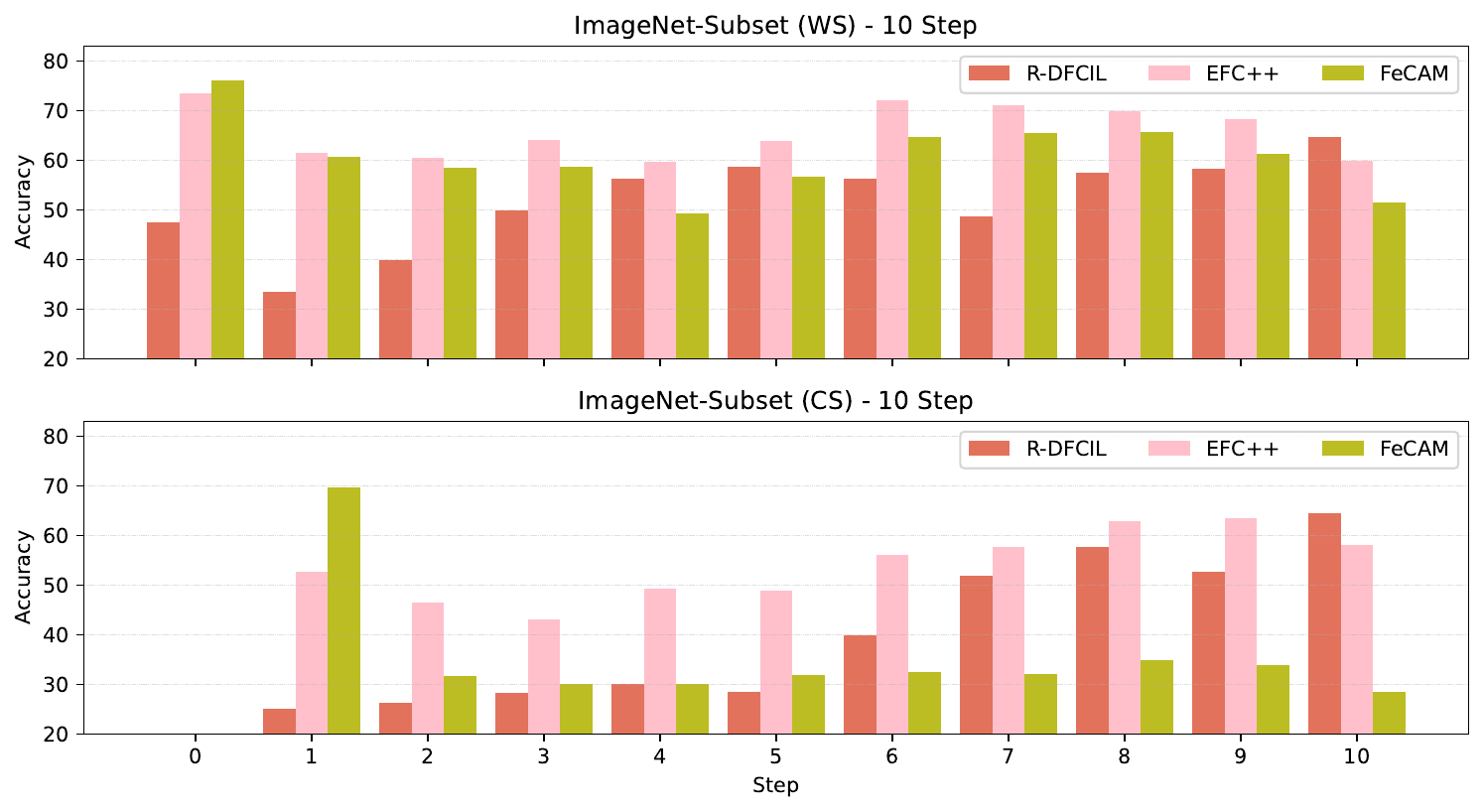}
    \caption{Accuracy on each task after the final training step on Tiny-ImageNet and ImageNet-Subset Warm Start and Cold Start. EFC++ achieves the best stability-plasticity trade-off compared to the nearest competitor in Warm Start (FeCAM) and Cold Start (R-DFCIL).}
    \label{fig:tiny_imgnet_bars}
\end{figure}

In Figure 6 of the main paper, we present the accuracy on each CIFAR-100 task after the final training step for both the Warm and Cold start scenarios. In Figure~\ref{fig:tiny_imgnet_bars}, we provide a similar plot for the remaining small-scale benchmarks. The results again confirm that the approaches exhibit significantly different performance in the Warm start versus Cold start settings, reflecting their respective stability-plasticity trade-offs.

\section{Large-Scale Class- and Domain-Incremental Learning} 
\label{app:exp2}
In this section, we evaluate EFC++ on large-scale incremental learning benchmarks, integrating the results from Section 6.2 of the main paper.  

\minisection{Experiments on ImageNet-1K}  
Table~\ref{tab:sota-imagenet1k} completes the results presented in Table 3 of the main paper for ImageNet-1K. As already discussed in Section 6.2, EFC++ achieves remarkable results in the Cold Start scenario compared to all other competitors, while in the Warm Start setting, it achieves performance comparable to FeCAM and EFC. This can be explained by observing that the generalization gained on the large first task in a large-scale dataset reduces the need for plasticity when learning new tasks, thereby aligning the performance across these approaches. It must be noted that the results reported in~\cite{fetril} are slightly better in the Warm Start scenario in terms of average incremental accuracy compared to those reported in Table~\ref{tab:sota-imagenet1k}. This gap is likely due to the performance on the initial task, which has a significant impact on average incremental accuracy. Finally, as already observed in Section 6.2, R-DFCIL struggles in large-scale scenarios regardless of the number of classes in the first task. This is because its generative inversion network must learn to generate high-resolution images for 1,000 distinct classes.

\begin{table}[t]
\centering
\caption{Large scale Class-incremental Experiments. We compare with the state-of-the-art on ImageNet-1K.}

\setlength{\tabcolsep}{3pt}
\resizebox{0.95\linewidth}{!}{%
\begin{tabular}{clcccccccc}
\toprule
 &  & \multicolumn{2}{c}{$\boldsymbol{A_{\mathbf{step}}^K}$} & \multicolumn{2}{c}{$\boldsymbol{A_{\mathbf{inc}}^K}$} \\
\multicolumn{1}{l}{}& \textbf{Method} & \textbf{10 Step} & \textbf{20 Step} & \textbf{10 Step} & \textbf{20 Step}  \\
\cmidrule(lr){2-2}
\cmidrule(lr){3-4}
\cmidrule(lr){5-6}
\cmidrule(lr){7-8}

\multirow{5}{*}{\rotatebox{90}{WS}} & FeTrIL & $55.26 \pm 0.16$ & $51.10 \pm 0.22$ & $63.42 \pm 0.08$ & $60.93 \pm 0.12$ \\
 & FeCAM & $58.73 \pm 0.09$ & $\underline{55.67} \pm 0.16$ & $65.83 \pm 0.14$ & $64.22 \pm 0.14$ \\
  & R-DFCIL & $35.47 \pm 0.19$ & $29.27 \pm 0.53$ & $46.42 \pm 0.18$ & $42.13 \pm 0.58$ \\
   & EFC & $\textbf{59.49} \pm 0.07$ & $\textbf{55.94} \pm 0.09$ & $\textbf{66.64} \pm 0.10$ & $\textbf{64.89} \pm 0.10$ \\
\cmidrule(lr){2-2}
\cmidrule(lr){3-4}
\cmidrule(lr){5-6}

 & EFC++ & $\underline{59.09} \pm 0.15$ & $54.82 \pm 0.17$ & $\underline{66.43} \pm 0.11$ & $\underline{64.45} \pm 0.20$ \\

\cmidrule[0.6pt](l){1-6}
 
\multirow{5}{*}{\rotatebox{90}{CS}} & FeTrIL & $34.28 \pm 0.20$ & $26.64 \pm 0.25$ & $48.93 \pm 0.15$ & $40.02 \pm 0.15$ \\
 & FeCAM & $36.16 \pm 0.10$ & $27.24 \pm 0.19$ & $50.38 \pm 0.14$ & $40.67 \pm 0.24$ \\
  & R-DFCIL & $29.36 \pm 0.17$ & $22.30 \pm 0.19$ & $40.86 \pm 0.37$ & $34.95 \pm 0.17$ \\
  & EFC & $\underline{42.62} \pm 0.08$ & $\underline{36.32} \pm 0.35$ & $\underline{56.52} \pm 0.10$ & $\underline{49.80} \pm 0.20$ \\ 
\cmidrule(lr){2-2}
\cmidrule(lr){3-4}
\cmidrule(lr){5-6}

 & EFC++ & $\textbf{44.72} \pm 0.11$ & $\textbf{37.46} \pm 0.03$ & $\textbf{58.39} \pm 0.16$ & $\textbf{52.34} \pm 0.11$ \\
\bottomrule
\end{tabular}}
\label{tab:sota-imagenet1k}
\end{table}

\begin{table}[t]
\centering
\caption{Domain- and Class-incremental Experiments. We compare with the state-of-the-art on DN4IL, a balanced subset of DomainNet for incremental learning~\cite{gowda2023dual}.}

\setlength{\tabcolsep}{4pt}
\resizebox{0.70\columnwidth}{!}{%
\begin{tabular}{cl c c}
\toprule
 & & \multicolumn{2}{c}{\textbf{Cold Start}} \\
 
\multicolumn{2}{l}{} & \multicolumn{1}{c}{  $\boldsymbol{A_{\mathbf{step}}^K}$ } & \multicolumn{1}{c}{$\boldsymbol{A_{\mathbf{inc}}^K}$} \\

  & \textbf{Method}    & \textbf{6 Step}  & \textbf{6 Step}   \\
\cmidrule(lr){2-2}
\cmidrule(lr){3-4}
\multirow{9}{*}{\rotatebox[origin=c]{90}{DN4IL}} 
      & EWC     & $\phantom{0}9.81 \pm 0.47$ & $25.55 \pm  0.23$ \\
      & LwF     &  $10.50 \pm 0.89$ &  $27.86 \pm 0.56$  \\
      & PASS    &  $24.44\pm 0.19$  &  $37.14 \pm 0.10$  \\
      & FeTrIL  & $30.95\pm0.17$ & $40.39\pm0.11$    \\
      & SSRE    & $32.05\pm 0.09$  & $41.25\pm0.18$    \\
      & FeCAM    & $ 36.42\pm 0.12$  & $ 42.93\pm0.11$    \\
      & R-DFCIL    & $ 22.31 \pm 0.21 $  & $ 32.18 \pm 0.29 $    \\
      & EFC    & $\underline{38.27}\pm 0.16$  &  $\underline{45.85} \pm 0.25$  \\
      \cmidrule(lr){2-2}
\cmidrule(lr){3-4}
      & EFC++  &  $\mathbf{39.32}\pm0.15$  & $\mathbf{46.20} \pm 0.21$ \\
\bottomrule
\end{tabular}}
 
\label{tab:domain}
\end{table}

\begin{figure}[t]
    \centering \includegraphics[width=0.9\columnwidth]{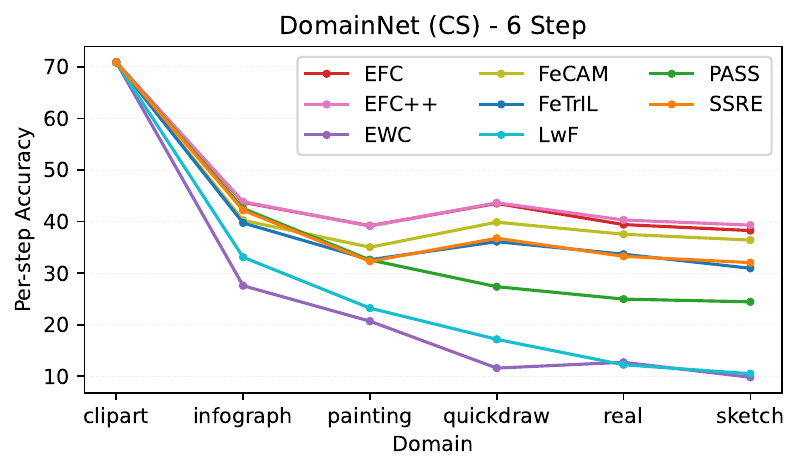}
    \caption{Per-step accuracy during the incremental learning on DN4IL. We compare EFC++ with the state-of-the-art in a setting in which each incremental learning step represents a different domain.}
    \label{fig:domain-net-perstep}
\end{figure}

\minisection{Experiments on DN4IL (DomainNet)}.
Table~\ref{tab:domain} extends Table 3 of the main paper for DN4IL by adding results for EwC, LwF, PASS, and SSRE, while also reporting the average incremental accuracy for all approaches. Among all the approaches, only FeCAM and SSRE achieve competitive performance, though it still falls short of the state-of-the-art. In Figure~\ref{fig:domain-net-perstep}, we show the per-step accuracy across all encountered domains.

\section{Validation on Regularization Hyperparameters}
\label{app:validation_hyper}
\begin{figure*}[t]
    
\includegraphics[width=\textwidth]{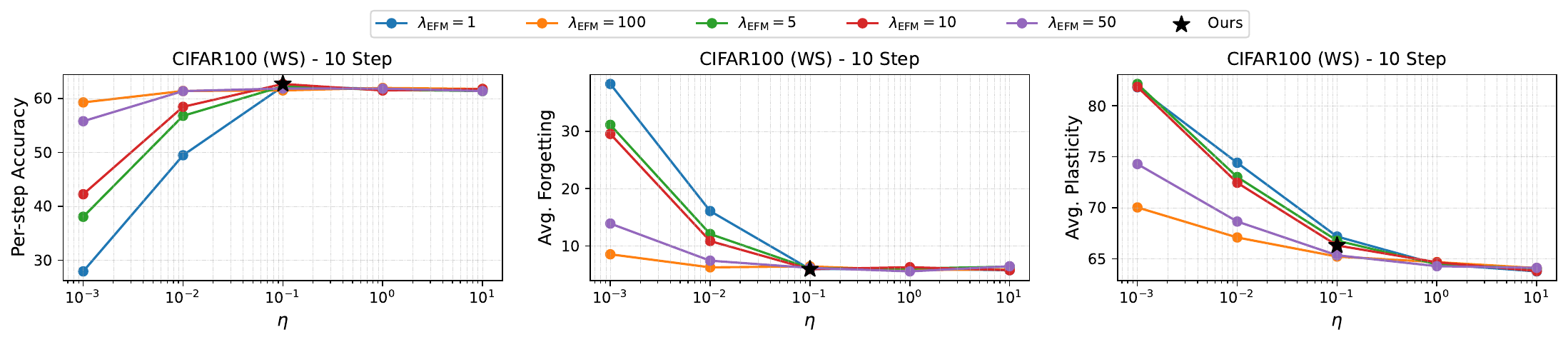}
\includegraphics[width=\textwidth]{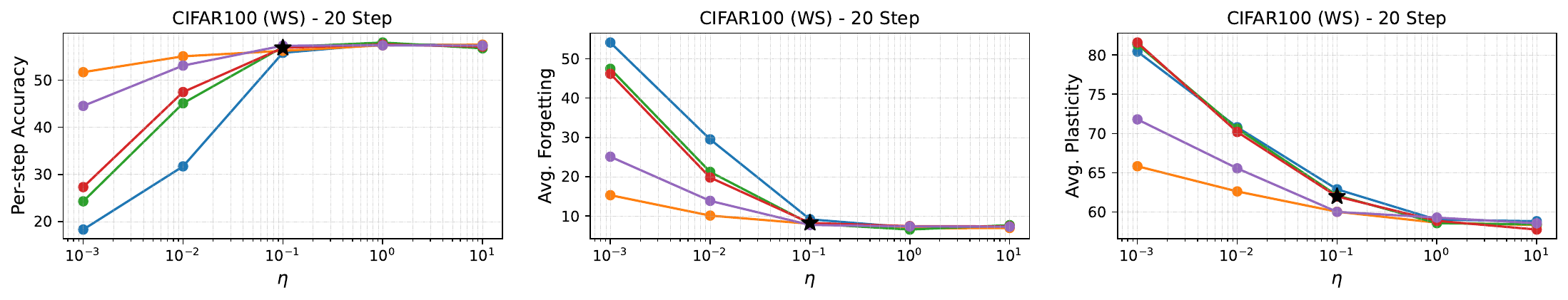}
\includegraphics[width=\textwidth]{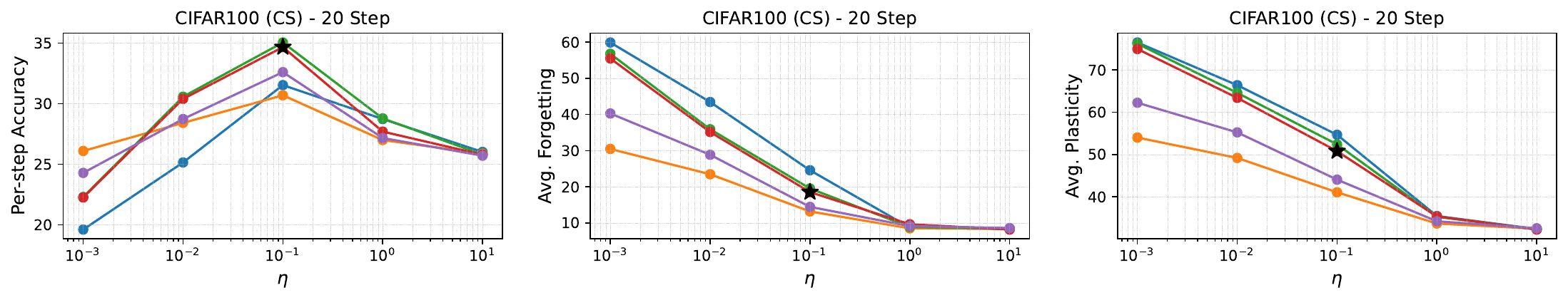}
\caption{A grid search testing 25 combinations of $\lambda_{\text{EFM}}$, with the $x$-axis in log scale for better visualization. The y-axes represent per-step accuracy, average forgetting, and average plasticity after the final step.}
\label{fig:abl_eta_lambda_app}
\end{figure*}

In this section, we present the results of the same analysis as Figure 12 in the main paper but for different scenarios on CIFAR-100. The grid search is performed over $\lambda_{\text{EFM}} \in \{1, 5, 10, 50, 100\}$ and damping $\eta \in \{10^{-3}, 10^{-2}, 10^{-1}, 1, 10\}$. Figure~\ref{fig:abl_eta_lambda_app} shows that the selected combination ($\lambda_{\text{EFM}} = 10$ and $\eta = 0.1$) exhibits a \textit{very good trade-off between stability and plasticity across all tested scenarios}. However, at the cost of an expensive search, we found that slightly better combinations exist for specific setting, e.g., changing $\lambda_{\text{EFM}}$ to 5 for a 20-step Cold Start or $\eta = 1$ for a 20-step Warm Start.  

\section{Ablation on EFM regularizer}
\label{app:ablations}
In Section 7.1 of the main paper, we discuss what happens when we substitute EFM with three established regularizers: the Fisher Information Matrix (E-FIM, Eq. 3 main), Feature Distillation (FD, Eq. 5 main), and Knowledge Distillation (KD). Here, we expand that discussion by providing more details on the choice of hyperparameters 
We recall that the \textit{only} modification to our training framework in this analysis concerns the regularization loss. The prototype update, which requires the EFM matrix, and the rebalancing of the classifier after training the backbone on the new classifier remain unchanged.  

For this analysis, we validate the hyperparameters for each regularizer: $\lambda_{\text{E-FIM}} \in \{100, 500, 5000, 10000, 100000\}$, $\lambda_{\text{KD}} \in \{1, 5, 10, 50, 100\}$, and $\lambda_{\text{FD}} \in \{0.1, 1, 5, 10, 50, 100\}$. Unlike EFM, where we have a theoretical explanation for the choice of hyperparameters (see Section 7.3 in the main paper), here we lack guidance on optimal values. Thus, in Table 5 of the main paper we report the results obtained with the hyperparameters that perform best in each setting. Looking at the final per-step accuracy, we found that the best trade-off across settings is achieved with $\lambda_{\text{E-FIM}} = 100,000$, $\lambda_{\text{KD}} = 50$, and $\lambda_{\text{FD}} = 1$.

\section{Covariance drift}
\label{app:cov_drift}
This section extends the discussion in Section 7.5 on the limitations of prototype rehearsal in EFC++ by illustrating the results obtained on Tiny-ImageNet and on ImageNet-Subset CS. In particular, Table~\ref{tab:cov_drift_tiny} and Table~\ref{tab:cov_drift_sub} compare different update strategies of class statistics: our approach (class means estimated via EFM and fixed covariances, highlighted in cyan) against an idealized upper bound using real mean and covariances.

\begin{table}[t]
\centering
\small
\caption{Impact of Mean and Covariance Drift on EFC++ tested on ImageNet-Subset CS.}
\label{tab:cov_drift_sub}
\resizebox{0.9\linewidth}{!}{%
\begin{tabular}{lllcc}
\toprule
\textbf{Step} & \textbf{Mean} & \textbf{Covariance} & \textbf{Accuracy} & \textbf{Delta} \\
\midrule
\multirow{5}{*}{10} 
    & Fixed     & Fixed     & $52.95 \pm 1.00$ & — \\
   
    & \cellcolor{cyan!20}EFM   & \cellcolor{cyan!20}Fixed & \cellcolor{cyan!20}$53.90 \pm 1.18$ & \cellcolor{cyan!20} +0.95 \\
    & EFM       & Real      & $52.44 \pm 1.21$ & $-$0.51 \\
    & Real      & Fixed     & $56.14 \pm 1.04$  & +3.19 \\

    & Real      & Real      & $55.69 \pm 1.13$ & +2.74 \\

\cmidrule{1-5}
\multirow{5}{*}{20} 
    & Fixed     & Fixed     & $39.10 \pm 0.86$ & — \\  
    & \cellcolor{cyan!20}EFM   & \cellcolor{cyan!20}Fixed     & \cellcolor{cyan!20}$40.86 \pm 1.65$ &\cellcolor{cyan!20} +1.76 \\
       & EFM   & Real      & $41.84 \pm \text{1.30}$  & +2.74  \\
    & Real      & Fixed     & $49.70 \pm \text{1.09}$ & +10.60 \\
 
    & Real      & Real      & $50.12 \pm \text{1.42}$  & +11.02 \\
\bottomrule
\end{tabular}}
\end{table}

\section{Per-Step Accuracy Plots}
\label{sup:per-step-plots}
\subsection{Warm Start Scenarios}
\label{sup:WS_per-step-plot}
\begin{figure*}
\includegraphics[width=0.99\textwidth]{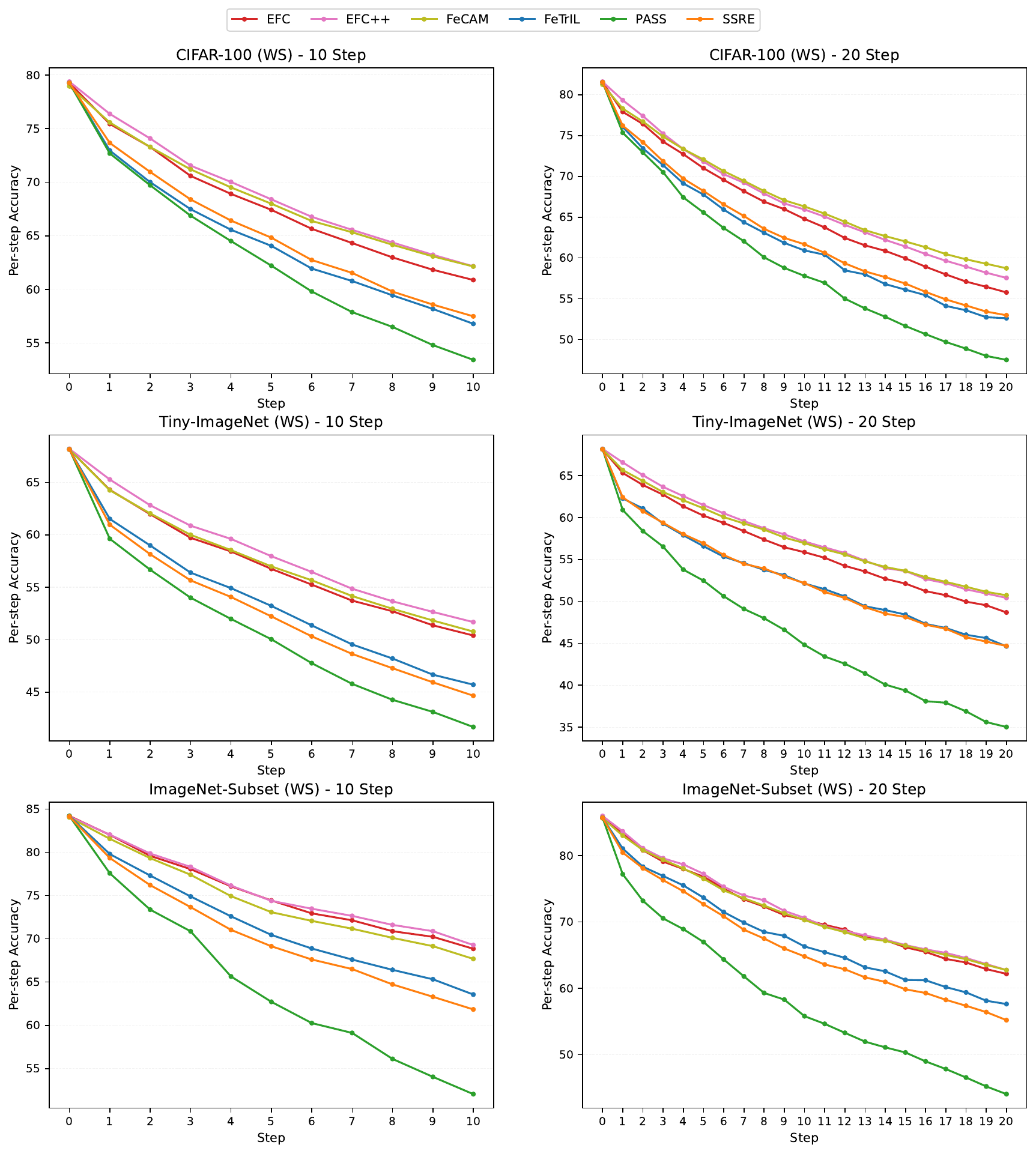}
    \caption{Warm Start (WS) per-step accuracy during the incremental learning. The plots compare recent EFCIL methods against EFC++ on three different datasets for different incremental step sequences. }
 \label{fig:per_step_WS_all}
\end{figure*}

In Figure~\ref{fig:per_step_WS_all} we compare the per-step incremental accuracy of EFC++ with recent approaches in Warm Start across all three datasets for various task sequences. All methods begin from the same starting point after training on the large first task. EFC++ obtains significantly better results compared
to FeCAM, in the 10-step scenario and achieves comparable results in the longest
scenario. Thanks to the Prototype Re-balancing phase, EFC++ achieves a better stability-plasticity tradeoff compared to EFC, consistently resulting in superior performance.

\subsection{Cold Start Scenarios}
\label{sup:CS_per-step-plot}
\begin{figure*}
\includegraphics[width=0.99\textwidth]{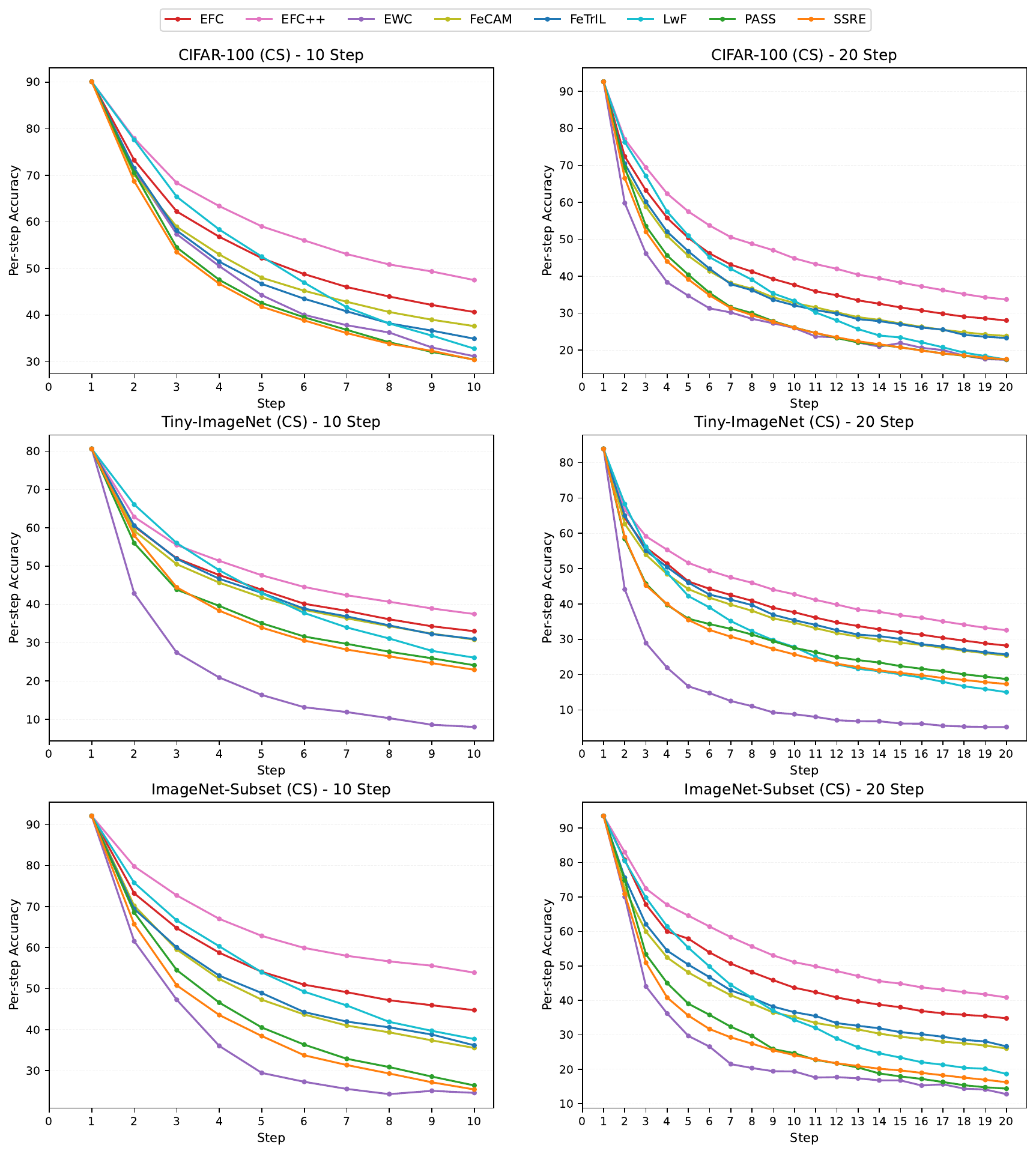}
    \caption{Cold Start (CS) per-step accuracy plots during incremental learning. The plots compare recent EFCIL methods with EFC++ on three different datasets for different incremental step sequences.}
    \label{fig:per_step_CS_all}
\end{figure*}

In Figure~\ref{fig:per_step_CS_all} we compare the per-step incremental accuracy of EFC++ with recent approaches in Cold Start across all three datasets for various task sequences. All methods begin from the same starting point after training on the small first task, but EFC++ (and EFC), which enhance plasticity thanks to the EFM regularizer, outperform state-of-the-art competitors. Thanks to the Prototype Re-balancing phase, EFC++ achieves a better stability-plasticity tradeoff compared to EFC, consistently resulting in superior performance.

\begin{table}[t]
\centering
\small
\caption{Impact of Mean and Covariance Drift on EFC++ tested on Tiny-ImageNet CS.}
\label{tab:cov_drift_tiny}
\resizebox{0.9\linewidth}{!}{%
\begin{tabular}{lllcc}
\toprule
\textbf{Step} & \textbf{Mean} & \textbf{Covariance} & \textbf{Accuracy} & \textbf{Delta} \\
\midrule
\multirow{5}{*}{10} 
    & Fixed     & Fixed     & $36.17 \pm 0.48$ & — \\
   
    & \cellcolor{cyan!20}EFM   & \cellcolor{cyan!20}Fixed & \cellcolor{cyan!20}$37.48 \pm 0.52$ & \cellcolor{cyan!20} +1.31 \\
    & EFM       & Real      & $37.47 \pm 0.27$ & +1.30 \\
    & Real      & Fixed     & $39.94 \pm 0.43$  & +3.77 \\

    & Real      & Real      & $39.95 \pm 0.55$ & +3.78 \\

\cmidrule{1-5}
\multirow{5}{*}{20} 
    & Fixed     & Fixed     & $29.04 \pm 0.65$ & — \\  
    & \cellcolor{cyan!20}EFM   & \cellcolor{cyan!20}Fixed     & \cellcolor{cyan!20}$32.56 \pm 0.44$ &\cellcolor{cyan!20} +3.52 \\
       & EFM   & Real      & $31.89 \pm 0.10$ & +2.85 \\
    & Real      & Fixed     & $36.30 \pm 0.31$ & +7.25 \\
 
    & Real      & Real      & $37.00 \pm 0.12$  & +7.96 \\
\bottomrule
\end{tabular}}
\end{table}


 \end{document}